\documentclass{article}

\usepackage{microtype}
\usepackage{graphicx}
\usepackage{subfigure}
\usepackage{booktabs} 
\usepackage{amsmath} 
\usepackage{amssymb}
\usepackage{enumitem}
\usepackage{alphalph}
\usepackage{etoolbox}
\usepackage{amsthm}  
\usepackage[dvipsnames]{xcolor}

\usepackage{minitoc}

\newtheorem{theorem}{Theorem}
\newtheorem{proposition}{Proposition}

\newtheorem{remark}{Remark}
\newtheorem{assumption}{Assumption}

\AtBeginDocument{%
  \AtBeginEnvironment{subequations}{%
  }
}

\allowdisplaybreaks

\newif\ifshowcomments
\showcommentsfalse

\ifshowcomments

\newcommand{\ben}[1]{\textcolor{OliveGreen}{[BA: #1]}}
\newcommand{\jp}[1]{\textcolor{Blue}{[JP: #1]}}

\else

\newcommand{\ben}[1]{}
\newcommand{\jp}[1]{}

\fi

\renewcommand{\cal}{\mathcal} 
\newcommand{\e}{{\varepsilon}}

\renewcommand{\P}{\mathbb{P}}
\newcommand{\E}{\mathbb{E}}
\newcommand{\C}{\mathbb{C}}

\newcommand{\R}{\mathbb{R}}

\newcommand{\deq}{\mathrel{\mathop:}=} 
\newcommand{\deqrev}{=\mathrel{\mathop:}} 
\newcommand{\ft}{\sigma_\textsc{t}}
\newcommand{\fs}{\sigma}
\newcommand{\bfx}{\mathbf{x}}
\newcommand{\tE}{T}
\newcommand{\Etrain}{E_\text{train}}
\newcommand{\Etest}{E_\text{test}}
\DeclareMathOperator{\tr}{tr}

\DeclareMathOperator{\diag}{diag}
\DeclareMathOperator{\erf}{erf}

\DeclareMathOperator{\relu}{ReLU}

\newcommand{\eq}[1]{\begin{equation}#1\end{equation}}

\newcommand{\al}[1]{\begin{align}#1\end{align}}
\newcommand{\als}[1]{\begin{align*}#1\end{align*}}

\newcommand{\pa}[1]{\left({#1}\right)}

\newcommand{\qa}[1]{\left[{#1}\right]}

\newcommand{\h}[1]{\{{#1}\}}

\newcommand{\ha}[1]{\left\{{#1}\right\}}

\newcommand{\abs}[1]{\lvert #1 \rvert}

\newcommand{\absa}[1]{\left\lvert #1 \right\rvert}

\newcommand{\norm}[1]{\lVert #1 \rVert}

\newcommand{\norma}[1]{\left\lVert #1 \right\rVert}

\usepackage{hyperref}

\usepackage[accepted]{icml2020}

\icmltitlerunning{The Neural Tangent Kernel in High Dimensions: Triple Descent and a Multi-Scale Theory of Generalization}

\begin{document}

\twocolumn[

\ifshowcomments
\icmltitle{\textcolor{red}{Comments are displayed!}The Neural Tangent Kernel in High Dimensions:\\Triple Descent and a Multi-Scale Theory of Generalization}
\else
\icmltitle{The Neural Tangent Kernel in High Dimensions:\\Triple Descent and a Multi-Scale Theory of Generalization}
\fi



\icmlsetsymbol{equal}{*}
\icmlsetsymbol{aires}{\textdagger}

\begin{icmlauthorlist}
\icmlauthor{Ben Adlam}{equal,aires,goo}
\icmlauthor{Jeffrey Pennington}{equal,goo}
\end{icmlauthorlist}

\icmlaffiliation{goo}{Google Brain}

\icmlcorrespondingauthor{Jeffrey Pennington}{jpennin@google.com}

\icmlkeywords{Machine Learning, ICML}

\vskip 0.3in
]



\printAffiliationsAndNotice{\icmlEqualContribution \textsuperscript{\textdagger}Work done as a member of the Google AI Residency Program.} 

\begin{abstract}

Modern deep learning models employ considerably more parameters than required to fit the training data. Whereas conventional statistical wisdom suggests such models should drastically overfit, in practice these models generalize remarkably well. An emerging paradigm for describing this unexpected behavior is in terms of a \emph{double descent} curve, in which increasing a model's capacity causes its test error to first decrease, then increase to a maximum near the interpolation threshold, and then decrease again in the overparameterized regime. Recent efforts to explain this phenomenon theoretically have focused on simple settings, such as linear regression or kernel regression with unstructured random features, which we argue are too coarse to reveal important nuances of actual neural networks. We provide a precise high-dimensional asymptotic analysis of generalization under kernel regression with the Neural Tangent Kernel, which characterizes the behavior of wide neural networks optimized with gradient descent. Our results reveal that the test error has non-monotonic behavior deep in the overparameterized regime and can even exhibit additional peaks and descents when the number of parameters scales quadratically with the dataset size.
\end{abstract}

\section{Introduction}
\label{sec_intro}

Machine learning models based on deep neural networks have achieved widespread success across a variety of domains, often playing integral roles in products and services people depend on. As users rely on these systems in increasingly important scenarios, it becomes paramount to establish a rigorous understanding for when the models might work, and, crucially, when they might not. Unfortunately, the current theoretical understanding of deep learning is modest at best, as large gaps persist between theory and observation and many basic questions remain unanswered.

\begin{figure}[t]
    \centering
    \includegraphics[width = 0.48\textwidth]{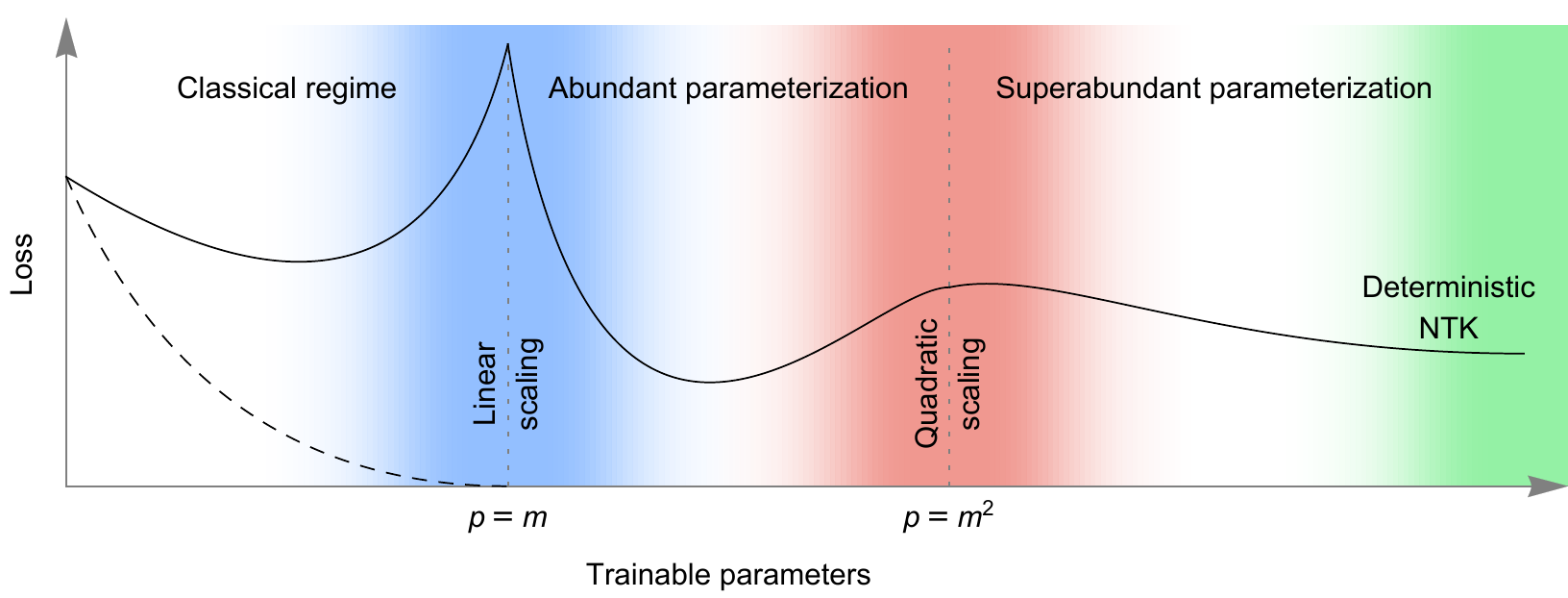}
    \vspace{-0.5cm}
    \caption{An illustration of multi-scale generalization phenomena for neural networks and related kernel methods. The classical U-shaped under- and over-fitting curve is shown on the far left. After a peak near the interpolation threshold, when the number of parameters $p$ equals the number of samples $m$, the test loss decreases again, a phenomenon known as \emph{double descent}. On the far right is the limit when $p\to\infty$, which is described by the Neural Tangent Kernel. In this work, we identify a new scale of interest in between these two regimes, namely when $p$ is quadratic in $m$, and show that it exhibits complex non-monotonic behavior, suggesting that double descent does not provide a complete picture. Putting these observations together we define three regimes separated by two transitional phases: (i) the \emph{classical regime} of underparameterization when $p<m$, (ii) the \emph{abundant parameterization} regime when $m<p<m^2$, and (iii) the \emph{superabundant parameterization} regime when $p>m^2$. The transitional phases between them are of particular interest as they produce non-monotonic behavior.}
    \vspace{-0.2cm}
    \label{fig:cartoon}
\end{figure}

One of the most conspicuous such gaps is the unexpectedly good generalization performance of large, heavily-overparameterized models. These models can be so expressive that they can perfectly fit the training data (even when the labels are replace by pure noise), but still manage to generalize well on real data~\cite{zhang2016understanding}. An emerging paradigm for describing this behavior is in terms of a double descent curve~\cite{belkin2019reconciling}, in which increasing a model's capacity causes its test error to first decrease, then increase to  a maximum near the interpolation threshold (where the number of parameters equals the number of samples), and then decrease again in the overparameterized regime. 

There are of course more elaborate measures of a model's capacity than a naive parameter count. Recent empirical and theoretical work studying the correlation of these capacity measures with generalization has found mixed results, with many measures having the opposite relationship with generalization that theory would predict~\cite{neyshabur2017exploring}. Other work has questioned whether it is possible in principle for uniform convergence results to explain the generalization performance of neural networks~\cite{nagarajan2019uniform}.

Our approach is quite different. We consider the algorithm's asymptotic performance on a specific data distribution, leveraging the large system size to get precise theoretical results. In particular, we examine the high-dimensional asymptotics of kernel ridge regression with respect to the Neural Tangent Kernel (NTK) \cite{jacot2018neural} and conclude that double descent does not always provide an accurate or complete picture of generalization performance. Instead, we identify complex non-monotonic behavior in the test error as the number of parameters varies across multiple scales and find that it can exhibit additional peaks and descents when the number of parameters scales quadratically with the dataset size.

Our theoretical analysis focuses on the NTK of a single-layer fully-connected model when the samples are drawn independently from a Gaussian distribution and the targets are generated by a wide teacher neural network. We provide an exact analytical characterization of the generalization error in the high-dimensional limit in which the number of samples $m$, the number of features $n_0$, and the number of hidden units $n_1$ tend to infinity with fixed ratios $\phi\deq n_0/m$ and $\psi \deq n_0/n_1$. By adjusting these ratios, we reveal the intricate ways in which the generalization error depends on the dataset size and the effective model capacity.

We investigate various limits of our results, including the behavior when the NTK degenerates into the kernel with respect to only the first-layer or only the second-layer weights. The latter corresponds to the standard setting of random feature ridge regression, which was recently analyzed in~\cite{mei2019generalization}. In this case, the total number of parameters $p$ is equal to the width $n_1$, i.e. $p = n_1 = (\phi/\psi)m$, so that $p$ is linear in the dataset size. In contrast, for the full kernel, the number of parameters is $p= (n_0+1)n_1 = (\phi^2/\psi)m^2 + (\phi/\psi) m$, \emph{i.e.} it is quadratic in the dataset size. By studying these two kernels, we derive insight into the generalization performance in the vicinities of linear and quadratic overparameterization, and by piecing these two perspectives together, we infer the existence of multi-scale phenomena, which sometimes can include triple descent. See Fig.~\ref{fig:cartoon} for an illustration and Fig.~\ref{fig:n1_ntk_triple_d} for empirical confirmation of this behavior.

\subsection{Our Contributions}
\label{subsec_contribution}
\begin{enumerate}[nolistsep]
    \item We derive exact high-dimensional asymptotic expressions for the test error of NTK ridge regression.
    \item We prove that the test error can exhibit non-monotonic behavior deep in the overparameterized regime.
    \item We investigate the origins of this non-monotonicity and attribute them to the kernel with respect to the second-layer weights.
    \item We provide empirical evidence that triple descent can indeed occur for finite-sized networks trained with gradient descent.
    \item We find exceptionally fast learning curves in the noiseless case, with $\Etest \sim m^{-2}$.
\end{enumerate}

\subsection{Related Work}
\label{subsec_related_work}
A recent line of work studying the behavior of interpolating models was initiated by the intriguing experimental results of~\cite{zhang2016understanding,belkin2018understand}, which showed that deep neural networks and kernel methods can generalize well even in the interpolation regime. A number of theoretical results have since established this behavior in certain settings, such as interpolating nearest neighbor schemes~\cite{belkin2018overfitting} and kernel regression~\cite{belkin2018does, liang2018just}.

These observations, coupled with classical notions of the bias-variance tradeoff, have given rise to the double descent paradigm for understanding how test error depends on model complexity. These ideas were first discussed in~\cite{belkin2019reconciling}, and empirical evidence was obtained in~\cite{advani2017high,geiger2019scaling} and recently in~\cite{nakkiran2019deep}. Precise theoretical predictions soon confirmed this picture for linear regression in various scenarios~\cite{belkin2019two, hastie2019surprises, mitra2019understanding}.

Linear models struggle to capture all of the phenomena relevant to double descent because the parameter count is tied to the number of features. Recent work found multiple descents in the test loss for minimum-norm interpolants in Reproducing Kernel Hilbert Spaces \cite{liang2020multiple}, but it similarly requires changing the data distribution to vary model capacity.
A precise analysis of a nonlinear system for a fixed data generating process is the most direct way to draw insight into double descent. 
A recent preprint~\cite{mei2019generalization} shares this view and adopts a similar analysis to ours, but focuses entirely on the standard case of unstructured random features. Such a setup can indeed model double descent, and certainly bears relevance to certain wide neural networks in which only the top-layer weights are optimized~\citep{neal1996priors,rahimi2008random,lee2017deep,matthews2018gaussian,lee2019wide}, but its connection to neural networks trained with gradient descent remains less clear.

Gradient-based training of wide neural networks initialized in the standard way was recently shown to correspond to kernel gradient descent with respect to the Neural Tangent Kernel~\cite{jacot2018neural}. This result has spawned renewed interest in kernel methods and their connection to deep learning; a woefully incomplete list of papers in this direction includes~\citet{lee2019wide,chizat2019lazy,du2018gradient,du2018provably,arora2019exact,xiao2019disentangling}.

To connect these research directions, our analysis requires tools and recent results from random matrix theory and free probability. A central challenge stems from the fact that many of the matrices in question have nonlinear dependencies between the elements, which arises from the nonlinear feature matrix $F=\fs(WX)$. This challenge was overcome in~\cite{pennington2017nonlinear}, which computed the spectrum of $F$, and in~\cite{pennington2018spectrum}, which examined the spectrum of the Fisher information matrix; see also~\cite{louart2018random}. We also utilize the results of~\cite{adlam2019random,peche2019note}, which established a linear signal plus noise model for $F$ that shares the same bulk statistics. This linearized model allows us to write the test error as the trace of a rational function of the underlying random matrices. The methods we use to compute such quantities rely on so-called \emph{linear pencils} that represent the rational function in terms of the inverse of a larger block matrix~\cite{helton2018applications}, and on operator-valued free probability for computing the trace of the latter~\cite{far2006spectra}.

\section{Preliminaries}
\label{sec_prelim}

In this section, we introduce our theoretical setting and some of the tools required to state our results. 

\subsection{Problem Setup and Notation}
\label{subsec_setup}
We consider the task of learning an unknown function from $m$ independent samples $(\bfx_i, y_i) \in \mathbb{R}^{n_0}\times \mathbb{R},\,i\le m$, where the datapoints are standard Gaussian, $\bfx_i \sim \cal{N}(0, I_{n_0})$, and the labels are generated by a wide\footnote{We assume the width $n_\textsc{t} \to \infty$, but the rate is not important.} single-hidden-layer neural network: 
\eq{
    y_i | \bfx_i, \Omega,\omega \sim \omega \ft(\Omega \bfx_i / \sqrt{n_0}) / \sqrt{n_\textsc{t}} + \varepsilon_i.
}
The teacher's activation function $\ft$ is applied coordinate-wise, and its parameters $\Omega \in \mathbb{R}^{n_\textsc{t}\times n_0}$ and $\omega \in \mathbb{R}^{1\times n_\textsc{t}}$ are matrices whose entries are independently sampled once for all data from $\cal{N}(0,1)$. We also allow for independent label noise, $\varepsilon_i \sim \cal{N}(0, \sigma_{\varepsilon}^2)$.

Let $\hat{y}(\bfx)$ denote the model's predictive function. We consider squared error, so the test loss is,
\eq{\label{eq_test_error}
    \E (y - \hat{y})^2 = \E_{\bfx,\varepsilon} (\omega \ft(\Omega \bfx/\sqrt{n_0})/\sqrt{n_\textsc{t}} + \varepsilon - \hat{y}(\bfx))^2,
}
where the expectation is over an iid test point $(\bfx, y)$ conditional on the training set, the teacher parameters, and any randomness in the learning algorithm producing $\hat{y}$, such as the random parameters defining the random features. Note that the test loss is a random variable; however, in the high-dimensional asymptotics we consider here, it concentrates about its mean.

\subsection{Neural Tangent Kernel Regression}
\label{subsec_kernel_regression}
We consider predictive functions $\hat{y}$ defined by approximate (\emph{i.e.} random feature) kernel ridge regression using the Neural Tangent Kernel (NTK) of a single-hidden-layer neural network. The NTK can be considered a kernel $K$ that is approximated by random features corresponding to the Jacobian $J$ of the network's output with respect to its parameters, \emph{i.e.} $K(\bfx_1,\bfx_2) = J(\bfx_1)J(\bfx_2)^\top$. As the width of the network becomes very large (compared to all other relevant scales in the system), the approximate NTK converges to a constant kernel determined by the network's initial parameters and describes the trajectory of the network's output under gradient descent. In particular,
\eq{\label{eq_ntk_t}
N_t(\bfx) = N_0(\bfx) + (Y - N_0(X))K^{-1}(I-e^{-\eta t K})K_\bfx\,,
}
where $N_t(\bfx)$ is the output of the network at time $t$, $K \deq K(\gamma) = K(X,X) + \gamma I_m$, $K_\bfx \deq K(X, \bfx)$, $\eta$ is the learning rate, and $\gamma$ is a ridge regularization constant\footnote{These overloaded definitions of $K$ can be distinguished by the number of arguments and should be clear from context.}. For this work, we are interested in the $t\to\infty$ limit of \eqref{eq_ntk_t}, which defines the predictive function,
\eq{\label{eq_yhat_ntk}
\hat{y}(\bfx) \deq N_\infty(\bfx) = N_0(\bfx) + (Y - N_0(X))K^{-1}K_\bfx\,.
}
We remark that if the width is not asymptotically larger than the dataset size, the validity of \eqref{eq_ntk_t} can break down and \eqref{eq_yhat_ntk} may not accurately describe the late-time predictions of the neural network. While this potential discrepancy is an interesting topic, we defer an in-depth analysis to future work (but see Fig.~\ref{fig:n1_ntk_triple_d}) for an empirical analysis of gradient descent). Instead, we regard \eqref{eq_yhat_ntk} as the definition of our predictive function and focus on kernel regression with the NTK. We believe this setup is interesting its own right; for example, recent work has demonstrated its effectiveness as a kernel method on complex image datasets \cite{li2019enhanced} and found it to be competitive with neural networks in small data regimes.

In this work, we restrict our study to the NTK of single-hidden-layer fully-connected networks. In particular, consider a network of with width $n_1$ and pointwise activation function $\fs$, defined by,
\eq{\label{eq_N0}
N_0(\bfx) = W_2\fs(W_1\bfx/\sqrt{n_0})/\sqrt{n_1}\,,
}
for initial weight matrices $W_1 \in \mathbb{R}^{n_1\times n_0}$ and $W_2\in\mathbb{R}^{1\times n_1}$ with iid entries $[W_1]_{ij} \sim \mathcal{N}(0,1)$\footnote{Any non-zero $\sigma_{W_1}^2$ can be absorbed into a redefinition of $\fs$.} and $[W_2]_{i} \sim \mathcal{N}(0,\sigma_{W_2}^2)$.

We collect our assumptions on the activation functions below, in Assumption~\ref{as_activation}. Their main purpose is to ensure that certain moments and derivatives exist almost surely, but for simplicity we state somewhat stronger conditions than are actually required for our analysis. To simplify the already cumbersome algebraic manipulations, we assume that $\sigma$ has zero Gaussian mean. We emphasize that this condition is not essential and our techniques easily generalize to all commonly used activation functions.

\begin{assumption}\label{as_activation}
    The activation functions $\sigma,\sigma_\textsc{t}:\R\to\R$ are assumed to be differentiable almost everywhere. We assume $\absa{\sigma(x)},\absa{\sigma'(x)},\absa{\sigma_\textsc{t}(x)}=\mathcal{O}\pa{\exp(Cx)}$ for some positive constant $C$, which implies all the Gaussian moments of $\sigma$, $\sigma'$, and $\sigma_\textsc{t}$ exist, and we assume $\E\sigma(Z)=0$ for $Z\sim\mathcal{N}(0,1)$. 
\end{assumption}

The Jacobian of \eqref{eq_N0} with respect to the parameters naturally decomposes into the Jacobian with respect to $W_1$ and $W_2$, \emph{i.e.} $J(\bfx) = [\partial N_0(\bfx)/\partial W_1, \partial N_0(\bfx)/\partial W_2] = [J_1(\bfx), J_2(\bfx)]$. Therefore the kernel $K$ also decomposes this way, and we can write.
\begin{align}
K(\bfx_1, \bfx_2) = & \; J_1(\bfx_1)J_1(\bfx_2)^\top +  J_2(\bfx_1)J_2(\bfx_2)^\top\\
\deqrev & \;  K_1(\bfx_1, \bfx_2) + K_2(\bfx_1, \bfx_2)
\label{eqn_K_K1K2}
\end{align}
A simple calculation yields the per-layer constituent kernels,
\begin{align}
\label{eq_K1}
K_1(X, X) & = \frac{X^\top X}{n_0} \odot \frac{\pa{F'}^\top \diag(W_2)^2 F'}{n_1}\\
\label{eq_K2}
K_2(X, X) &= \frac{1}{n_1}F^\top F\,,
\end{align}
where we have introduced the abbreviations $F=\fs(W_1X/\sqrt{n_0})$ and $F'=\fs'(W_1X/\sqrt{n_0})$. Notice that when $\sigma_{W_2}^2 \to 0$, $K = K_2$, \emph{i.e.} the NTK degenerates into the standard random features kernel. However, the predictive function \eqref{eq_yhat_ntk} contains an offset $N_0(\bfx)$ which would typically be set to zero in standard random feature kernel regression because it simply increases the variance of test predictions. Removing this variance component has an analogous operation in neural network training: either the function value at initialization can be subtracted throughout training, or a symmetrization trick can be used in which two copies of the NN are initialized identically, and their normalized difference $N \equiv \pa{N^{(a)}-N^{(b)}}/{\sqrt{2}}$ is trained with gradient descent. Either method preserves the kernel $K$ while enforcing $N_0 \equiv 0$. We call this setup \emph{centering}, and present results with and without it.

Finally, we note that ridge regularization in the kernel perspective corresponds to using L2 regularization of the neural network's weights toward their initial values.

\section{Three Regimes of Parameterization}

In this section, we outline an argument based on the structure of the NTK as to why one should expect the test error to exhibit non-trivial phenomena at two different scales of overparameterization. From the expressions for the test error \eqref{eq_test_error} and the predictive function \eqref{eq_yhat_ntk}, it is evident that the behavior of the test error is determined by the spectral properties of the NTK. Although the fine details of the relationship can only be revealed by the explicit calculation, we can nevertheless make some basic high-level observations based on the coarser structure of the kernel.

The number of trainable parameters $p$ relative to the dataset size $m$ controls the amount of parameterization or complexity of a model. In our setting of a single-hidden-layer fully-connected neural network, $p=n_1(n_0+1)$, and for a fixed dataset, we can adjust the ratio $p/m$ by varying the hidden-layer width $n_1$.

\begin{figure}[t]
    \centering
    \includegraphics[width = 0.48\textwidth]{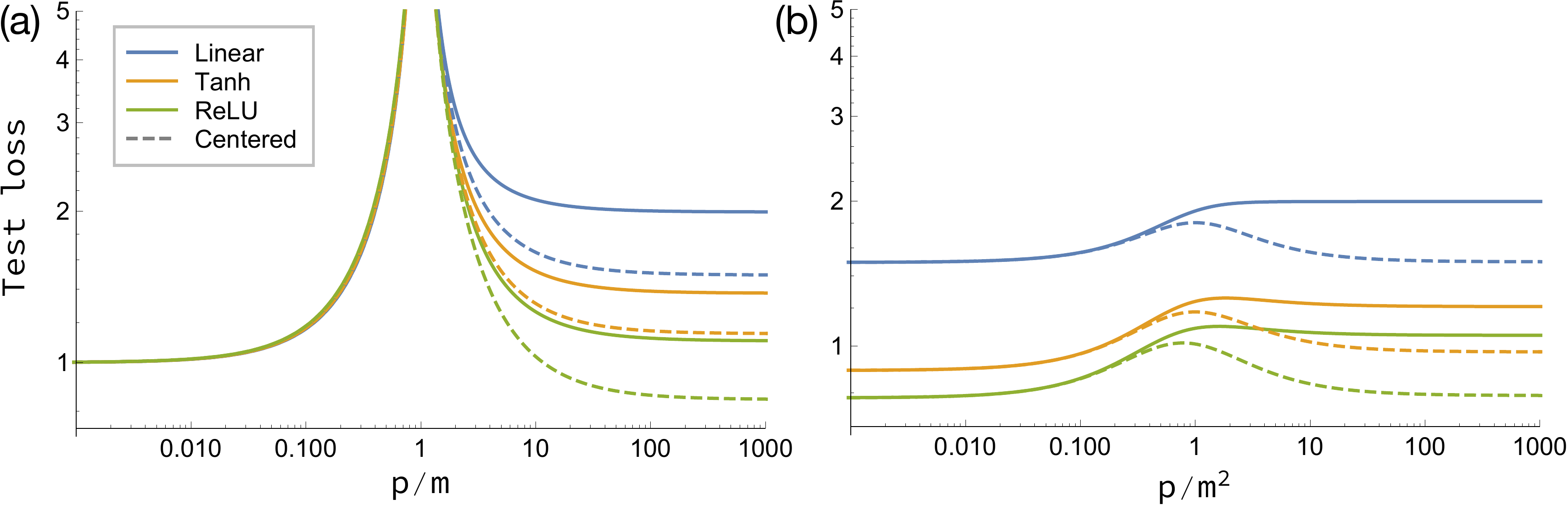}
    \vspace{-0.6cm}
    \caption{Theoretical results for the test error with and without centering for different activation functions with $\phi = 2$, $\gamma = 10^{-3}$, and $\text{SNR} = 1$ for (a) the second-layer kernel $K_2$ and (b) the full NTK $K$ as the number of parameters $p$ is varied by changing the network width. Nonmonotonic behavior is clearly visible at the linear scaling transition ($p=m$) and the quadratic scaling transition ($p=m^2$). Here, ReLU denotes the zero-mean function $\sigma(x) = \text{max}(x,0) - 1/\sqrt{2\pi}$.}
    \label{fig:K2vsp}
\end{figure}

The simplest way to see that there should be two scales comes from examining the two terms in the kernel separately. Because $K_1=J_1J_1^T$ and $J_1\in \mathbb{R}^{m\times n_0 n_1}$, the first-layer kernel has rank at most $\min\h{n_0 n_1, m}$, which suggests nontrivial transitional behavior when $p = \Theta(m)$. Similarly, the rank of $K_2$ is at most $\min\h{n_1, m}$, which suggests a second interesting scale when $n_1 = \Theta(m)$, or equivalently, when $p = \Theta(m^2)$ if $n_0 = \Theta(n_1)$. Our explicit calculations confirm that interesting phenomena indeed occur at these scales, as can be seen in Fig.~\ref{fig:K2vsp}.

\label{sec_results1}

These two scales partition the degree of parameterization into three regimes. We consider the \emph{classical} regime to be when $p\lesssim m$ because classical generalization theory tends to hold and the U-shaped test error curve is observed.  The transition around $p=\Theta(m)$ manifests as a sharp rise in the test loss near the interpolation threshold, followed by a quick descent as $p$ increases further, as can be seen in Fig.~\ref{fig:K2vsp}(a). We call this the \emph{linear scaling} transition. After this, we enter a regime we call \emph{abundant parameterization} when $m\lesssim p \lesssim m^2$. In this regime, the test error tends to decrease until $p$ nears the vicinity of $m^2$, where it can sometimes increase again, producing a second U-shaped curve. When $p=\Theta(m^2)$, another transition is observed, which we call the \emph{quadratic scaling} transition, which can be seen in Fig.~\ref{fig:K2vsp}(b). On the other side of this transition, $p\gtrsim m^2$, a regime we call \emph{superabundant parameterization}. See Fig~\ref{fig:cartoon} for an illustration of this general picture.

While the classical regime has been long studied, and the superabundant regime has generated considerable recent interest due to the NTK, our main aim in delineating the above regimes is to highlight the existence of the intermediate scale containing complex phenomenology
. For this reason, we focus our theoretical analysis on the novel scaling regime in which $p = \Theta(m^2)$. In particular, as mentioned in Sec.~\ref{sec_intro}, we consider the high-dimensional asymptotics in which $n_0, n_1, m \to \infty$ with $\phi \deq n_0/m$ and $\psi \deq n_0/n_1$ held constant.

\section{Overview of Techniques}
\label{sec_overview}
In this section, we provide a high-level overview of the analytical tools and mathematical results we use to compute the generalization error. To begin with, let us first describe the main technical challenges in computing explicit asymptotic limits of~\eqref{eq_test_error}.

The first challenge, which is evident upon inspecting \eqref{eq_K1}, is that the kernel contains a Hadamard product of random matrices, for which concrete results in the random matrix literature are few and far between. We address this problem in Sec.~\ref{sec_hadamard}.

The second challenge, which is apparent by inspecting \eqref{eq_K2}, is that the kernel depends on random matrices with nonlinear dependencies between the entries. We describe how to circumvent this difficulty in Sec.~\ref{sec_gaussian}.

Finally, by expanding the square in \eqref{eq_test_error} and substituting \eqref{eq_yhat_ntk}, we find terms that are constant, linear, and quadratic in $K^{-1}$. Some of the random matrices that appear inside the matrix inverses (\emph{e.g.} $X$, and $W_1$) also appear outside of them as multiplicative factors, a situation that prevents the straightforward application of many standard proof techniques in random matrix theory. We describe how to overcome this challenge in Sec.~\ref{sec_ovfp}.

\subsection{Simplification of First-Layer Kernel}
\label{sec_hadamard}
A straightforward central limiting argument shows that in the asymptotic limit the entries of $W_1X/\sqrt{n_0}$ are marginally Gaussian with mean zero and unit variance. As such, the first and second moments of the entries in the matrix $F' = \fs'(W_1X/\sqrt{n_0})$ are equal to
\eq{
\sqrt{\zeta} \deq \mathbb{E}_{z\sim \mathcal{N}(0,1)} \fs'(z)\,,\quad \eta' \deq \mathbb{E}_{z\sim \mathcal{N}(0,1)} \fs'(z)^2\,.
}
It follows that we can split $K_1$ into two terms,
\eq{\label{eq_k1_split}
    \frac{X^\top X}{n_0} \odot \frac{\pa{\bar{F}'}^\top \diag(W_2)^2 \bar{F}'}{n_1} + \sigma_{W_2}^2 \zeta \frac{X^\top X}{n_0},
}
where $\bar{F}'$ is a centered version of $F'$. Focusing on the first term, because $n_0 n_1 = \phi^2/\psi m^2$, the random fluctuations in the off-diagonal elements are $\mathcal{O}(1/m)$, which are too small to contribute to the spectrum or moments of an $m\times m$ matrix whose diagonal entries are order one. In fact, the diagonal entries are simply proportional to the variance of the entries of $F'$, namely $(\eta'-\zeta)$. Putting this together, we can eliminate the Hadamard product entirely and write,
\eq{
\label{eq_k1_split_simp}
K_1 \cong \sigma_{W_2}^2 (\eta'-\zeta) I_m + \frac{\sigma_{W_2}^2 \zeta}{n_0}X^\top X\, ,
}
where the $\cong$ notation means the two matrices share the same bulk statistics asymptotically. We make this argument precise in Sec.~\ref{sec_lin_ntk}.

\subsection{Linearization 1: Gaussian Equivalents}
\label{sec_gaussian}
The test error \eqref{eq_test_error} involves large random matrices with nonlinear dependencies, which are not immediately amenable to standard methods of analysis in random matrix theory. The main culprit is the random feature matrix $F=\fs(W_1X/\sqrt{n_0})$, but $f \deq \fs(W_1\bfx/\sqrt{n_0})$, $Y = \omega \ft(\Omega X / \sqrt{n_0}) / \sqrt{n_\textsc{t}} + \mathcal{E}$, and $y \deq \omega \ft(\Omega \bfx / \sqrt{n_0}) / \sqrt{n_\textsc{t}}$ all suffer from the same issue.

The solution is to replace each of these matrices with an equivalent matrix without nonlinear dependencies, but chosen to maintain the same first- and second-order moments for all of the terms that appear in the test error \eqref{eq_test_error}. This approach was described for $F$ in~\cite{adlam2019random} (see also~\cite{peche2019note}). The upshot is that the test error is asymptotically invariant to the following substitutions,
\begin{align}
F \to F^{\text{lin}} &\deq \sqrt{\frac{\zeta}{n_0}} W_1 X + \sqrt{\eta - \zeta} \Theta_F \label{eq_Flin}\\
Y \to Y^{\text{lin}} &\deq \sqrt{\frac{\zeta_\textsc{t}}{n_\textsc{t} n_0}} \omega \Omega X + \sqrt{\frac{\eta_\textsc{t} - \zeta_\textsc{t}}{n_\textsc{t}}} \omega \Theta_Y + \mathcal{E} \label{eq_Ylin} \\
f \to f^{\text{lin}} &\deq \sqrt{\frac{\zeta}{n_0}} W_1 \bfx + \sqrt{\eta - \zeta} \theta_f \label{eq_flin}\\
y \to y^{\text{lin}} &
\deq \sqrt{\frac{\zeta_\textsc{t}}{n_\textsc{t}n_0}} \omega\Omega \bfx + \sqrt{\frac{\eta_\textsc{t} - \zeta_\textsc{t}}{n_\textsc{t}}} \omega\theta_y \label{eq_ylin}\,.
\end{align}
The new objects $\Theta_F$, $\Theta_Y$, $\theta_f$, and $\theta_y$ are matrices of the appropriate shapes with iid standard Gaussian entries. The constants $\eta, \zeta, \eta_\textsc{t}$, and $\zeta_\textsc{t}$ are chosen so that the mixed moments up to second order are the same for the original and linearized versions. In particular,
\begin{align}
    \zeta &\deq [\mathbb{E}_{z\sim \mathcal{N}(0,1)} \fs'(z)]^2\,,\; \;\;\,\eta \deq \mathbb{E}_{z\sim \mathcal{N}(0,1)} \fs(z)^2\,,\\ 
    \zeta_\textsc{t} &\deq [\mathbb{E}_{z\sim \mathcal{N}(0,1)} \ft'(z)]^2\,,\;\; \eta_\textsc{t} \deq \mathbb{E}_{z\sim \mathcal{N}(0,1)} \ft(z)^2\,.
\end{align}
The statement that the test error only depends on $Y^{\text{lin}}$ is consistent with the observations made in~\cite{ghorbani2019linearized,mei2019generalization} that in the high-dimensional regime where $n_0 = \Theta(m)$, only linear functions of the data can be learned. Indeed, $Y^{\text{lin}}$ is equivalent to a linear teacher plus noise with signal-to-noise ratio given by,
\eq{\label{eq_SNR}
\text{SNR} = \frac{\zeta_\textsc{t}}{\eta_\textsc{t} -\zeta_\textsc{t} + \sigma_\varepsilon^2}\,.
}
We often make this equivalence to a linear teacher explicit by setting $\ft(x) = x$, which implies $\eta_\textsc{t} = \zeta_\textsc{t} = 1$. Doing so also removes the noise from the test label, but since this noise merely contributes an additive shift to the test loss, removing it does not change any of our conclusions.

\subsection{Linearization 2: Linear Pencil}
\label{sec_ovfp}
Next we turn our attention to the actual computation of the asymptotic test loss. Expanding the test error~\eqref{eq_test_error} we have\footnote{For simplicity, we discuss the centered setting with $N_0 = 0$, which captures all of the technical complexities.},
\begin{align}
\Etest & \deq \mathbb{E}_{(\bfx,y)} (y - \hat{y}(\bfx))^2 \\
& = \mathbb{E}_{(\bfx,\varepsilon)}\Big[\tr(y^\top y) -2\tr(K_\bfx^\top K^{-1}Y^\top y)\nonumber\\
&\qquad\quad\quad + \tr(K_\bfx^\top K^{-1}Y^\top Y K^{-1} K_\bfx)\Big]\label{eq_Etest3}\,.
\end{align}
The simplification~\eqref{eq_k1_split_simp} gives,
\begin{align}
    K &= \sigma_{W_2}^2 \qa{(\eta' - \zeta)I_m + \frac{\zeta X^\top\! X}{n_0}} + \frac{F^\top \!F}{n_1} + \gamma I_m \label{eq_K_simp}\\
    K_\bfx &= \frac{\sigma_{W_2}^2 \zeta}{n_0}X^\top \bfx + \frac{1}{n_1} F^\top f\,,
\end{align}
which, when applied to \eqref{eq_Etest3} together with the substitutions \eqref{eq_Flin}-\eqref{eq_ylin}, expresses the test error directly in terms of the iid Gaussian random matrices $W_1,X,\Theta_F,\Omega,\Theta_Y,\mathcal{E},\theta_f,\theta_y$ and $\bfx$. The expectations over $\bfx$ and $\mathcal{E}$ are trivial because these variables do not appear inside the matrix inverse $K^{-1}$. Moreover, asymptotically the traces concentrate around their means with respect to $\Omega,\Theta_Y,\theta_f$ and $\theta_y$, which we can also compute easily for the same reason. Therefore, the test error can be written as,
\eq{\label{eq_Etest_sum}
\Etest = a_0 + \sum_i b_i \tr(B_i K^{-1}) + \sum_i c_i \tr(C_i K^{-1} D_i K^{-1})\,
}
where $B_i, C_i, D_i$ are monomials in $\{W_1, X, \Theta_F\}$ and their transposes, and $a_0, b_i, c_i \in \mathbb{R}$.

Eqn.~\eqref{eq_Etest_sum} is a rational function of the noncommutative random variables $W_1, X,$ and $\Theta_F$. A useful result from noncommutative algebra guarantees that such a rational function can be \emph{linearized} in the sense that it can be expressed in terms of the inverse of a matrix whose entries are linear in the noncommutative variables. This representation is often called a linear pencil, and is not unique; see \emph{e.g.}~\cite{helton2018applications} for details.

To illustrate this concept, consider the simple case of $K^{-1}$. After applying the substitutions~\eqref{eq_Flin}-\eqref{eq_ylin} to \eqref{eq_K_simp}, a linear pencil is given by
\eq{
\left[\begin{smallmatrix}
[\gamma + \sigma_{W_2}^2 (\eta' - \zeta)]I & \frac{\sigma_{W_2}^2 \zeta}{n_0}X^\top & \frac{\sqrt{\eta - \zeta}}{n_0} \Theta_F^\top & \frac{\sqrt{\zeta}}{\sqrt{n_0}n_1}X^\top&\\
-X & I & 0 & 0&\\
-\sqrt{\eta - \zeta}\Theta_F & -\frac{\sqrt{\zeta}}{\sqrt{n_0}}W_1 & I & 0&\\
0 & 0 & -W_1^\top & I&
\end{smallmatrix}\right]^{-1}_{11}\,,\nonumber
}
which can be checked by an explicit computation of the block matrix inverse. After obtaining a linear pencil for each of the terms in~\eqref{eq_Etest_sum}, the only task that remains is computing the trace. Since each linear pencil is a block matrix whose blocks are iid Gaussian random matrices, its trace can be evaluated using the techniques described in~\cite{far2006spectra} or through the general formalism of operator-valued free probability. We refer the reader to the book~\cite{mingo2017free} for more details on these topics.

\section{Asymptotic Training and Test Error}
The calculations described in the previous section are presented in the Supplementary Materials. Here we present the main results.
\label{sec:asymptotic}
\begin{proposition}
\label{prop:t1t2}
As $n_0,n_1,m\to\infty$ with $\phi = n_0/m$ and $\psi = n_0/n_1$ fixed, the traces $\tau_1(z) \deq \frac{1}{m}\E\tr(K(z)^{-1})$ and $\tau_2(z) \deq \frac{1}{m}\E \tr(\frac{1}{n_0}X^\top X K(z)^{-1})$ are given by the unique solutions to the coupled polynomial equations,
\begin{equation}
\begin{split}
\label{eq:prop1}
  &\phi\left(\zeta  \tau_2 \tau_1+ \phi(\tau_2 -\tau_1) \right)+\zeta  \tau_1 \tau_2
  \psi  \left(z \tau_1-1\right)\\
  &\quad = -\zeta  \tau_1 \tau_2 \sigma _{W_2}^2
  \left(\zeta  \left(\tau_2-\tau_1\right) \psi +\tau_1 \psi  \eta '+\phi \right)\\
  &\zeta \tau_1^2 \tau_2 \left(\eta'-\eta\right) \sigma _{W_2}^2+\zeta  \tau_1 \tau_2 \left(z \tau_1 -1\right)\\
  &\quad = \left(\tau_2-\tau_1\right) \phi  \left(\zeta  \left(\tau_2-\tau_1\right)+\eta  \tau_1\right)\,,
\end{split}
\end{equation}
such that $\tau_1, \tau_2\in \mathbb{C}^+$ for $z \in \mathbb{C}^+$.
\end{proposition}
\begin{theorem}
\label{thm:main}
Let $\gamma = \text{Re}(z)$ and let $\tau_1$ and $\tau_2$ be defined as in Proposition~\ref{prop:t1t2} with $\text{Im}(z) \to 0^+$. Then the asymptotic training error $\Etrain = \frac{1}{m}\mathbb{E}\lVert Y - \hat{y}(X)\rVert_F^2$ is given by,
\begin{equation}
\begin{split}
    \Etrain &= -\gamma^2(\sigma_\varepsilon^2 \tau_1' + \tau_2') + \nu \sigma_{W_2}^2\gamma^2 (\tau_1 + \gamma \tau_1')\\
    &\quad +\nu \sigma_{W_2}^4\gamma^2\left((\eta'-\zeta)\tau_1'+\zeta \tau_2'\right)\,,
\end{split}
\end{equation}
and the asymptotic test error $\Etest = \mathbb{E} (y - \hat{y}(\bfx))^2$ is given by
\begin{equation}
\label{eq:Etest_Etrain}
    \Etest = (\gamma \tau_1)^{-2}\Etrain - \sigma_{\varepsilon}^2\,.
\end{equation}
\end{theorem}

\begin{remark}
The subtraction of $\sigma_{\varepsilon}^2$ in eqn.~(\ref{eq:Etest_Etrain}) is because we have assumed that there is no label noise on the test points. Had we included the same label noise on both the training and test distributions, that term would be absent. 
\end{remark}
\begin{remark}
When $\nu = 0$, the quantity $(\gamma \tau_1)^{-2}\Etrain$ on the right hand side of eqn.~(\ref{eq:Etest_Etrain}) is precisely the generalized cross-validation (GCV) metric of~\cite{golub1979generalized}. Theorem~\ref{thm:main} shows that the GCV gives the exact asymptotic test error for the problem studied here.
\end{remark}

\section{Test Error in Limiting Cases}
\label{sec:limit}
While the explicit formulas in preceding section provide an exact characterization of the asymptotic training and test loss, they do not readily admit clear interpretations. On the other hand, eqn.~(\ref{eq:prop1}) and therefore the expressions for $\Etest$ simplify considerably under several natural limits, which we examine in this section.
\begin{figure*}[th!]
    \centering
    \includegraphics[width = \textwidth]{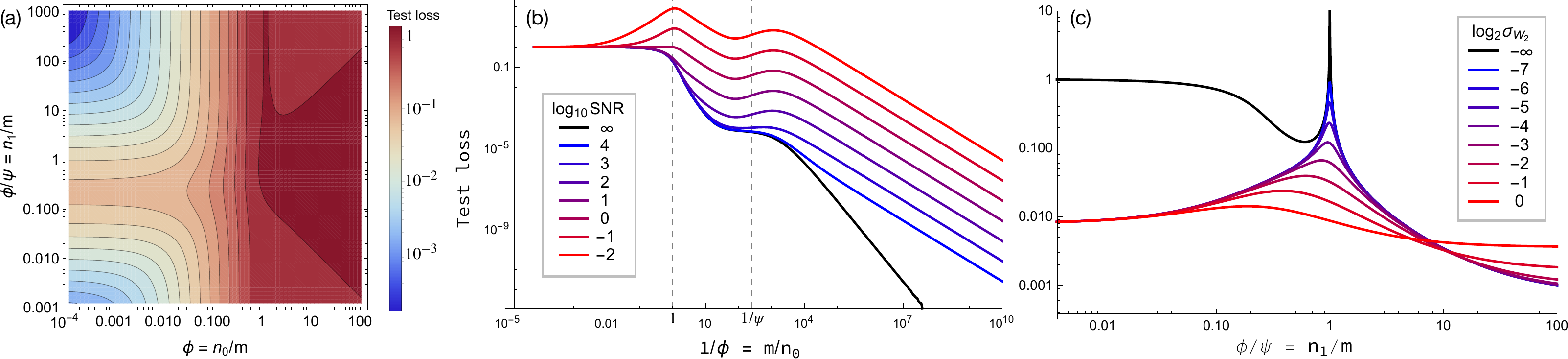}
    \vspace{-0.5cm}
    \caption{Test error for NTK regression with $\fs = \tanh$ under various scenarios. (a) Contour plot of the error as a function of $\phi = n_0/m$ and $\phi/\psi = n_1/m$ for $\gamma = 0$ and $\text{SNR}=1$. The non-monotonic behavior is evident not just in the width $n_1$, but also in the number of features $n_0$. (b) Learning curves for the NTK for different signal-to-noise ratios. With no noise (black curve), the error decreases quadratically in the dataset size $m$, otherwise it decreases linearly. Dashed lines indicate $m=n_0$ and $m=n_1$, where humps emerge for low SNR. (c) Test error as a function of width for various values of $\sigma_{W_2}$, which controls the relative contribution of $K_1$ and $K_2$. As $\sigma_{W_2}$ decreases (red to blue), the kernel becomes more like $K_2$ and the small hump at the quadratic transition increases in size until it resembles the large spike at the linear transition, suggesting that $K_2$ is responsible for the non-monotonicity in the overparameterized regime.}
    \label{fig:snr_and_c1}
\end{figure*}
\subsection{Large Width Limit}
Here we examine the test error in the superabundant regime in which the width $n_1$ is larger than any constant times the dataset size $m$, which can be obtained by letting $\psi \to 0$ and $\psi/\phi \to 0$. In this setting we find,
\begin{align}
    \Etest|_{\psi = 0} & = \frac{1}{2\phi \chi_0}\left(\chi_0(\phi-1) + \xi \phi(1+\phi) + \rho(1-3\phi)\right) \nonumber\\
    &\quad+\frac{\nu \sigma_{W_2}^2}{2 \phi\chi_0}\left((\eta \phi+\zeta)(\rho + \xi \phi) -4 \zeta \rho\phi  \right)\nonumber\\
    &\quad+\frac{\nu \sigma_{W_2}^2}{2 \phi} (\eta \phi-\zeta) + \frac{\phi \xi+ \rho -\chi_0}{2\chi_0 \text{SNR}}\label{eq_E_psi0}\,,
\end{align}
where $\nu=0$ with centering and $\nu=1$ without it and $\rho \deq \zeta(1+\sigma_{W_2}^2)$, $\,\xi \deq \gamma + \eta + \sigma_{W_2}^2 \eta'$, and 
\eq{
\chi_0 \deq \sqrt{(\rho + \xi \phi)^2 - 4 \phi \rho^2}\,.
}
The learning curve is remarkably steep with centering. To see this, we expand the result as $m\to\infty$, \emph{i.e.} as $\phi \to 0$,
\eq{ \label{eq_Etest_psi0_smallphi}
\Etest|_{\psi = 0} = 
\begin{cases}
\frac{\phi}{\text{SNR}} + \mathcal{O}(\phi^2) & \text{SNR}<\infty \\ (1-\frac{\xi}{\rho})^2 \phi^2 + \mathcal{O}(\phi^3) & \text{SNR} = \infty
\end{cases}\,.
}
Interestingly, we see that when the network is super abundantly parameterized, we obtain very fast learning curves: for finite SNR, $\Etest \sim m^{-1}$, and in the noiseless case $\Etest \sim m^{-2}$. See Fig~\ref{fig:snr_and_c1}(b).

\subsection{Small Width Limit}
Here we consider the limit in which the width $n_1$ is smaller than any constant times the dataset size $m$ or the number of features $n_0$, which can be obtained by letting $\psi\to\infty$ with $\phi$ held constant. In this setting we find,
\begin{align}
\Etest|_{\psi\to\infty} & = \frac{1}{2\phi \chi_1}\left(\chi_1 (\phi-1) + \xi_1 \phi(1+\phi) + \zeta(1-3\phi)\right) \nonumber\\
    & \quad + \frac{1}{2 \chi_1 \text{SNR}}\left(\phi \xi_1+ \zeta -\chi_1\right)\label{eq:test_error_small_width}\,,
\end{align}
where $\,\xi_1 \deq \eta' + \gamma/\sigma_{W_2}^2$, and 
\eq{
\chi_1 \deq \sqrt{(\zeta + \xi_1 \phi)^2 - 4 \phi \zeta^2}\,.
}
The small width limit characterizes one boundary of the abundant parameterization regime and as such provides an upper bound on the test loss in that regime. Therefore, a sufficient condition for the global minimum to occur at intermediate widths is $\Etest|_{\psi\to\infty} < \Etest|_{\psi = 0}$. By comparing eqn.~(\ref{eq_E_psi0}) to eqn.~(\ref{eq:test_error_small_width}), precise though unenlightening constraints on the parameters can be derived for satisfying this condition. One such configuration is illustrated in Fig.~\ref{fig:n1_ntk_triple_d}(b).
\subsection{Large Dataset Limit}
Here we consider the limit in which the dataset $m$ is larger than any constant times the width $n_1$, which can be obtained by letting $\phi \to 0$ with $\phi/\psi \to 0$. In this setting we find,
\begin{align}
    \Etest|_{\phi \to 0} =
\begin{cases}
\frac{1 + \psi}{\text{SNR}}(\frac{\phi}{\psi}) + \mathcal{O}(\frac{\phi}{\psi})^2 & \text{SNR} < \infty \\
\frac{\tau^2 (\nu \zeta^2 \sigma_{W_2}^4 + \kappa)}{(\eta-\zeta)\zeta^2 \sigma_{W_2}^4}(\frac{\phi}{\psi})^2 + \mathcal{O}(\frac{\phi}{\psi})^3 & \text{SNR} = \infty\nonumber 
\end{cases}\,,
\end{align}
where $\nu = 0$ with centering and $\nu = 1$ with without it and,
\eq{
\tau \deq \gamma + \sigma_{W_2}^2(\eta'-\zeta)\,,\quad \kappa \deq \zeta \psi + (\eta - \zeta)\psi^2\,.
}
Here again we observe very steep learning curves, similar to the large width limit above. 

\subsection{Ridgeless Limit: First-Layer Kernel}
Here we examine the ridgeless limit $\gamma\to 0$ of the first-layer kernel $K_1$. We find that the result can be obtained through a degeneration of \eqref{eq_E_psi0},
\begin{align}
    \Etest^{K_1}&|_{\gamma = 0}  = \lim_{\sigma_{W_2}\to\infty} \Etest|_{\psi = 0} \\
    & = \frac{1}{2\phi \bar{\chi}}\left(\bar{\chi}(\phi-1) + \eta' \phi(1+\phi) + \zeta(1-3\phi)\right) \nonumber\\
    & \quad + \frac{1}{2\bar{\chi} \text{SNR}}\left(\phi \eta'+ \zeta -\bar{\chi}\right)\,,
\end{align}
where, $\bar{\chi} \deq \sqrt{(\zeta + \eta' \phi)^2 - 4 \phi \zeta^2}$ and we have specialized to the centered case $\nu=0$. The expansion as $m\to\infty$ also looks similar to \eqref{eq_Etest_psi0_smallphi} and can be obtained from that equation by substituting $\xi/\rho \to \eta'/\zeta$.

\subsection{Ridgeless Limit: Second-Layer Kernel}
Here we examine the ridgeless limit $\gamma \to 0$ when the kernel is due to the second-layer weights only, \emph{i.e.} $K_2$. This limit can be obtained by letting $\sigma_{W_2}\to 0$. In this setting, the result can be expressed as,
\begin{align}
    \Etest^{K_2}|_{\gamma = 0} &= \frac{\phi}{\text{SNR}}\frac{1}{|\phi - \psi|}+\frac{2 \omega \zeta - \beta}{2 \zeta |\phi-\psi|}\;+ \nonumber \\ 
    &\quad\quad \delta_{\phi>\psi}\left(\frac{\beta - 2\chi}{2 \chi \text{SNR}} -\frac{\beta (\eta - \zeta)}{2 \zeta \chi}\right)\label{eq:K2test}\,,
\end{align}
where $\omega \deq \max\h{\phi,\psi}$, $\beta \deq \zeta + \omega \eta - \chi$, and 
\eq{
\chi = \sqrt{(\zeta + 4 \omega \eta)^2 - 4 \omega \zeta^2}\,,
}
and we have again specialized to the centered case $\nu = 0$. This expression is in agreement with the result presented in~\cite{mei2019generalization}.

\begin{figure*}[th!]
    \centering
    \includegraphics[width = \textwidth]{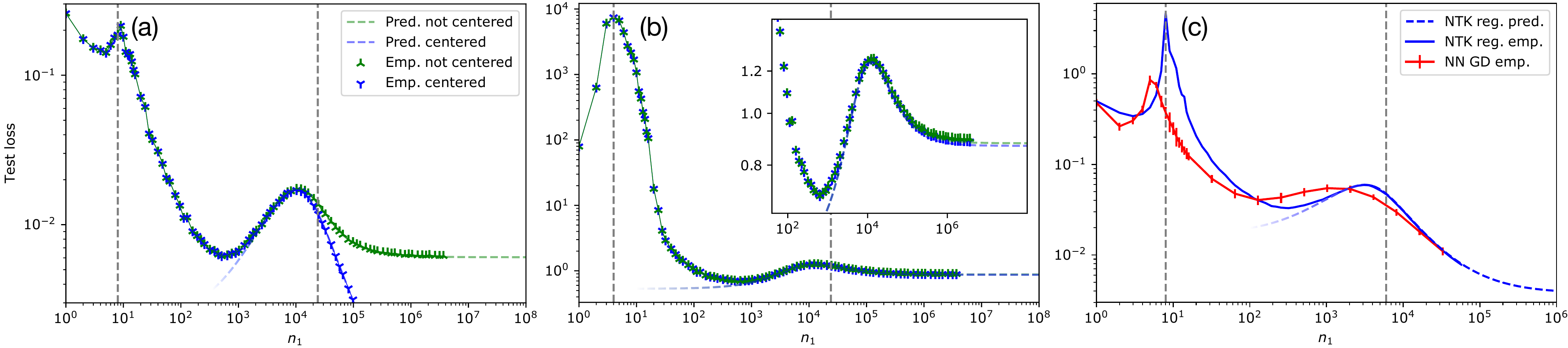}
    \vspace{-0.3cm}
    \caption{Empirical validation of multi-scale phenomena, triple descent, and the linear and quadratic scaling transitions for kernel regression (a,b,c) and gradient descent (c). All cases show a peak near the linear parameterization transition (first dashed vertical line), as well as a bump near the quadratic transition (second dashed vertical line). Theoretical predictions (dashed blue) agree with kernel regression in their regime of validity (quadratic parameterization). While the global minimum is often at $n_1=\infty$, it need not be as illustrated in (b). The NTK does not perfectly describe gradient dynamics in high dimensions, so the deviations between the red (GD) and blue (kernel regression) curves in (c) are expected. (a) Mean of five trials with $m=24000$, $n_0=3000$, $\sigma_{W_2}^2=1/8$, $\sigma_\e^2=0$, $\gamma=10^{-6}$, and $\fs=\erf$. (b) Mean of five trials with $m=24000$, $n_0 = 6000$, $\sigma_{W_2}^2=1/8$, $\sigma_\e^2=4$, and $\fs = c(\erf(6(x+1)+\erf(6(x-1))$ with $c$ chosen so $\zeta = 1/4$. (c) Mean and standard deviation of 20 trials with $m=6000$, $n_0=750$, $\sigma_{W_2}^2=1/8$, $\sigma_\e^2=0$, and $\fs=\relu$.}
    \label{fig:n1_ntk_triple_d}
\end{figure*}
When the system is far in the regime of abundant parameterization, namely $p=n_1 \gg m$ (or $\psi/\phi \to 0$), we can examine the large dataset behavior by first sending $\psi \to 0$ and then expanding as $\phi \to 0$. The result is described by \eqref{eq_Etest_psi0_smallphi} by substituting $\xi/\rho \to \eta/\zeta$.

\section{Quadratic Overparameterization}
\label{sec:pheno}

In this section, we investigate the implications of our theoretical results about the generalization performance of NTK regression in the quadratic scaling limit $n_0,n_1,m\to\infty$ with $\phi=n_0/m$ and $\psi = n_0/n_1$ held constant. Our high-level observation is that there is complex non-monotonic behavior in this regime as these ratios are varied, and that this behavior can depend on the signal-to-noise ratio and the initial parameter variance $\sigma_{W_2}^2$ in intricate ways. We highlight a few examples in Fig.~\ref{fig:snr_and_c1}.

In Fig.~\ref{fig:snr_and_c1}(a), we plot the test error as a function of $\phi$ and $\phi/\psi$, which reveals the behavior of jointly varying the number of features $n_0$ and the number of hidden units $n_1$. As expected from Fig.~\ref{fig:K2vsp}(b), for fixed $\phi$ the test error has a hump near $n_1 = m$. Perhaps unexpectedly, for large $n_1$, the test loss exhibits non-monotonic dependence on $n_0$, with a spike near $n_0 = m$. Notice that for small $n_1$, this non-monotonicity disappears. It is clear that the test error depends in a complex way on both variables, underscoring the richness of the quadratically-overparameterized regime.

Fig.~\ref{fig:snr_and_c1}(b) shows learning curves for fixed $\psi$ and various values of the $\text{SNR}$. For small enough $\text{SNR}$, there are visible bumps in the vicinity of $m=n_0$ and $m=n_1$ that reveal the existence of regimes in which more training data actually hurts test performance. Note that $n_0 = \Theta(n_1)$ so these two humps are separated by a constant factor, so the presence of two humps in this figure is not evidence of multi-scale behavior, though it surely reflects the complex behavior at the quadratic scale.

It is natural to wonder about the origins of this complex behavior. Can it be attributed to a particular component of the kernel $K$? We investigate this question in Fig.~\ref{fig:snr_and_c1}(c), which shows how the test error changes as the relative contributions of the per-layer kernels $K_1$ and $K_2$ are varied. By decreasing $\sigma_{W_2}$, the contribution of $K_1$ decreases and the kernel becomes more like $K_2$, and the small hump at the quadratic transition increases in size until it resembles the large spike at the linear transition (\emph{c.f.} Fig.~\ref{fig:K2vsp}), suggesting that $K_2$ is in fact responsible for the non-monotonicity in the quadratically-overparameterized regime.

\section{Empirical Validation}
\label{sec_empirical}
Our theoretical results establish the existence of nontrivial behavior of the test loss at $p=m$ for the second-layer kernel $K_2$ and at $p=m^2$ for the full kernel $K$. While these results are strongly suggestive of multi-scale behavior, they do not prove this behavior exists for a single kernel, nor do they guarantee it will be revealed for finite-size systems, let alone for models trained with gradient descent. Here we provide positive empirical evidence on all counts.

Fig.~\ref{fig:n1_ntk_triple_d} demonstrates multi-scale phenomena, triple descent, and the linear and quadratic scaling transitions for random feature NTK regression and gradient descent for finite-dimensional systems. The simulations all show a peak near the linear parameterization transition, as well as a bump near the quadratic transition. The asymptotic theoretical predictions agree well with kernel regression in their regime of validity, which is when $n_1$ is near $m$. While we found that the global minimum of the test error is often at $p=\infty$, there are some configurations for which the optimal $p$ lies between $m$ and $m^2$, as illustrated in Fig.~\ref{fig:n1_ntk_triple_d}(b).

Fig.~\ref{fig:n1_ntk_triple_d}(a) clearly shows triple descent for NTK regression and a marked difference in loss with and without centering, suggesting that this source of variance may often dominate the error for large $n_1$.

Fig.~\ref{fig:n1_ntk_triple_d}(c) confirms the existence of triple descent for a single-layer neural network trained with gradient descent. The noticeable difference between kernel regression and the actual neural network is to be expected because the NTK can change during the course of training when the width is not significantly larger than the dataset size. Indeed, the deviation diminishes for large $n_1$. In any case, the qualitative behavior is similar across all scales, providing support for the validity of our framework beyond pure kernel methods.

\section{Conclusion}
In this work, we provided a precise description of the high-dimensional asymptotic generalization performance of kernel regression with the Neural Tangent Kernel of a single-hidden-layer neural network. Our results revealed that the test error has complex non-monotonic behavior deep in the overparameterized regime, indicating that double descent does not always provide an accurate or complete picture of generalization performance. Instead, we argued that the test error may exhibit additional peaks and descents as the number of parameters varies across multiple scales, and we provided empirical evidence of this behavior for kernel ridge regression and for neural networks trained with gradient descent. We conjecture that similar multi-scale phenomena may exist for broader classes of architectures and datasets, but we leave that investigation for future work.
\bibliography{main}
\bibliographystyle{icml2020}

\newpage
\onecolumn

\begin{center}
\textbf{{\large Supplemental Material} \\
}
\end{center}
\setcounter{equation}{0}
\renewcommand{\theequation}{S\arabic{equation}}
\renewcommand{\bibnumfmt}[1]{[S#1]}
\setcounter{table}{0}
\renewcommand{\thetable}{S\arabic{table}}
\setcounter{figure}{0}
\renewcommand{\thefigure}{S\arabic{figure}}
\setcounter{remark}{0}
\renewcommand{\theremark}{S\arabic{remark}}
\setcounter{section}{0}
\renewcommand{\thesection}{S\arabic{section}}
\setcounter{theorem}{0}
\renewcommand{\thetheorem}{S\arabic{theorem}}
\makeatletter

\newcommand{\ttau}{\tilde{\tau}}

\renewcommand{\addcontentsline}{\oldaddcontentsline}
\setcounter{tocdepth}{2}
\tableofcontents

\section{Simplification of the first-layer kernel}
\label{sec_lin_ntk}
In this section, we get explicit control in spectral norm of the difference between the empirical (i.e. finite-size) NTK and the version in eqn.~(\ref{eq_K_simp}) that arises through the simplification of the first-layer kernel $K_1$ in eqn.~(\ref{eq_k1_split_simp}). We will use the notation $A_{i:}=(A_{i1},\ldots, A_{in})$, where $A_{:i}$ is defined similarly. Recall from eqns.~(\ref{eq_K1}) and~(\ref{eq_K2}) that the empirical NTK is given by 
\eq{
    K \deq \frac{X^\top X}{n_0} \odot \frac{(F')^\top \diag(W_2)^2F'}{n_1} + \frac{F^\top F}{n_1} + \gamma I
}
and from eqn.~(\ref{eq_K_simp}) the simplified kernel is given by
\eq{
    K_\text{simp}\deq \zeta\frac{X^\top X}{n_0} +(\eta'-\zeta)I + \frac{F^\top F}{n_1}  + \gamma I,
}
Also, define
\eq{\label{eq_R_def}
    R\deq \frac{\zeta'}{n_0}\mathbf{1}\mathbf{1}^\top,
}
where $\zeta'\deq \qa{\E_{z\sim\mathcal{N}(0,1)} \sigma''(z)}^2$. In this section, we show for any $\e, \delta>0$
\eq{\label{eq_spectral}
    \P\ha{ \norm{K-K_\text{simp}-R} > n_0^{2\e-1/4} } < \delta
}
for sufficiently large $n_0$. 

Let $\E$ be expectation over $W_1$ and $W_2$ conditional on $X$. We note that with high-probability for any $\e>0$ that $(X^\top X/n_0)_{ab} = \delta_{ab} + \cal{O}\pa{n_0^{\e-1/2}}$ for all $a$ and $b$, and that $\norm{X^\top X/n_0}\leq n^\e$, since $X$ is i.i.d. Gaussian. The use of $\cal{O}$ hides uniform constants.

Define 
\begin{equation}
    \Delta_k \deq \frac{X^\top X}{n_0} \odot \pa{(W_2)_k^2 \pa{F_{k:}'}^\top F_{k:}' - \bar{M} }\quad\text{and}\quad
    \Delta \deq \frac{1}{n_1}\sum_{k=1}^{n_1} \Delta_k,
\end{equation}
where $\bar{M}\deq \E \pa{F_{k:}'}^\top F_{k:}'$ (which does not depend on $k$), so $\E \Delta_k=0$. Then
\eq{\label{eq_k_diff}
   K - K_\text{simp} - R = \Delta +  \qa{\frac{X^\top X}{n_0} \odot \pa{M - \bar{M} } - R } + \pa{\frac{X^\top X}{n_0} \odot M - K_\text{simp} },
}
where $M \deq \zeta \mathbf{1}\mathbf{1}^\top + (\eta'-\zeta) I$. Elementary arguments given in Sec.~\ref{sec:bounding_remaining} show that, in operator norm, the two rightmost terms in eqn.~\eqref{eq_k_diff} are bounded by $\cal{O}(n_0^{3\e-1/2})$. In Sec.~\ref{sec:bounding_delta}, we bound $\Delta$ by using the fact that, conditional on $X$, $\Delta$ is a sum of independent random matrices to apply the matrix Bernstein inequality \cite{tropp2015introduction}. 

\subsection{Bounding $\Delta$}
\label{sec:bounding_delta}
We start with a supremum bound on $\norm{\Delta_k}$. For any vector $\mathbf{v} = \sum_k v_k \mathbf{e}_k $, we have
\eq{\label{eq_sup_norm}
    \norm{\Delta_k \mathbf{v} } \leq\sum_k \abs{v_k}\norm{\Delta_k \mathbf{e}_k} \leq n_1^{-1} \sup_{a,b} \abs{(W_2)_k^2 F'_{ka}F'_{kb} - \bar{M}} \cdot \norm{X^\top X/n_0} \sqrt{m}, 
}
by the Cauchy-Schwarz inequality. Note that by assumption on $X$, eqn.~\eqref{eq_sup_norm} is $\cal{O}\pa{n_0^{2\e}m^{1/2}/n_1} = \cal{O}(n_0^{3\e -1/2})$.

Now we bound the variance term. Consider the $(a,b)$ entry of $ \E \Delta_k^2$:
\als{
    &\frac{1}{n_1^2} \sum_{l=1}^{m} (X^\top X/n_0)_{al}(X^\top X/n_0)_{lb} \E \qa{ \pa{(W_2)_k^2 F'_{ka} F'_{kl} -\bar{M}_{al} }\pa{(W_2)_k^2 F'_{kl} F'_{kb} -\bar{M}_{lb}} } \\
    &= \frac{1}{n_1^2} \sum_{l=1}^{m} (X^\top X/n_0)_{al}(X^\top X/n_0)_{lb} \pa{3\E\qa{ F'_{ka}F'_{kl}F_{kl}' F_{kb}' }- \E\qa{ F'_{kl}F'_{kb}} \E \qa{F'_{ka} F'_{kl}} },
}
which we note is the same for all $k$. We now calculate these 2- and 4-point expectations to leading order.

Since the entries of $WX$ are multivariate Gaussian conditional on $X$, we find
\eq{
    \E F'_{ka} F'_{kb} = \E f'(Z_a) f'(Z_b),
}
where 
\al{ 
    (Z_a,Z_b)&\sim \cal{N}\pa{ \mathbf{0}, \frac{1}{n_0} \begin{pmatrix} X_{:a}^\top X_{:a} & X_{:a}^\top X_{:b} \\ X_{:a}^\top X_{:b} & X_{:b}^\top X_{:b} \end{pmatrix} } \\
    &\equiv \cal{N}\pa{ \mathbf{0},  \begin{pmatrix} 1+ \cal{O}\pa{n_0^{\e-1/2}} &  \cal{O}\pa{n_0^{\e-1/2}} \\ \cal{O}\pa{n_0^{\e-1/2}} & 1 + \cal{O}\pa{n_0^{\e-1/2}} \end{pmatrix} }.
}
Taylor expanding in the covariance term, one can show that, for all $a$,
\eq{\label{eq_f_var}
     \E f'(Z_a)^2 = \eta' +  \tilde{R} \pa{\frac{X_{:a}^\top X_{:a}}{n_0}-1}  + \cal{O}(n_0^{2\e-1}),
}
where $\tilde{R}\deq \E\qa{ f''(Z)^2 + f'(Z)f'''(Z)}$,
and for all $a\neq b$,
\eq{\label{eq_f_covar}
     \E f'(Z_a) f'(Z_b) = \zeta + \frac{\xi\xi''}{2}\pa{\frac{X_{:a}^\top X_{:a}}{n_0}+\frac{X_{:b}^\top X_{:b}}{n_0}-2} + \zeta'\frac{X_{:a}^\top X_{:b}}{n_0} + \cal{O}(n_0^{2\e-1}),
}
where $\xi\deq \E f'(Z)$ and $\xi''\deq \E f''(Z)$. Using the same argument, we find \jp{I guess this is $l\neq a$? Also, I would use $l\to c$, i.e. keep the early letters for the second index}
\eq{
    \E \pa{F'_{ka}}^2\pa{F'_{kl}}^2 = \pa{\eta'}^2 + \cal{O}(n_0^{2\e-1/2});
}
\eq{
    \E \pa{F'_{ka}}^4 = C_4 + \cal{O}(n_0^{2\e-1/2});
}
for $l, a,b$ distinct,
\eq{
    \E F'_{ka}F_{kb}' \pa{F'_{kl}}^2 = \zeta \eta'+\mathcal{O}\pa{n_0^{2\e-1/2}};
}
for $l\neq a$,
\eq{
    \E F'_{ka} \pa{F'_{kl}}^3 = C_3 + \cal{O}(n_0^{2\e-1/2}),
}
for some constants $C_3$ and $C_4$.

Thus, we may write
\eq{
   \sum_{k} \E \Delta_k^2 = \frac{1}{n_1} (X^\top X/n_0)^2 \odot M_2 + E,
}
where
\eq{
    M_2 \deq (3\zeta\eta' - \zeta^2) \mathbf{1}\mathbf{1}^\top + 3\eta'(\eta' - \zeta) I
}
and
\eq{
    E_{ab}\deq\frac{1}{n_1}\sum_l (X^\top X/n_0)_{al}(X^\top X/n_0)_{lb} \e_{abl}
}
for some $\e_{abl} = \cal{O}(n_0^{2\e-1/2})$. We find $\norm{\frac{1}{n_1} (X^\top X/n_0)^2 \odot M_2} = \cal{O}(n_0^{\e}/n_1)$ and
\al{
    \norm{E} &\leq \norm{E}_\textsc{f} \\
    &= \pa{ \sum_{a,b} \abs{E_{ab}}^2 }^{1/2} \\
    & =\pa{  \frac{1}{n_1^2} \sum_{a,b} \pa{ \sum_l (X^\top X/n_0)_{al}^2 (X^\top X/n_0)_{lb}^2 \sum_l \e_{abl}^2}   }^{1/2} \\
    & =\sqrt{ \cal{O}(n_0^{6\e - 1} n_1^{-2} m^2\pa{ m^2 n_0^{-2} + mn_0^{-1} + mn_0^{-2} +1 }}\\
    & = \cal{O}(n_0^{3\e-1/2})
}
using the Cauchy-Schwarz inequality and that assumption that all dimensions are on the same order.

Thus finally applying the matrix Bernstein inequality with $t=C n_0^{4\e-1/4}$ for some sufficiently large constant $C$, we find for any $\delta>0$
\eq{\label{eq_spectral_bound}
    \P\ha{ \norm{\Delta} > C n_0^{4\e-1/4} } < \delta
}
for sufficiently large $n_0$. Moreover, eqn.~\eqref{eq_spectral_bound} holds with $X$ random as it is independent of $W_1$ and $W_2$, and our assumptions on $X$ hold for any $\delta'>0$ for sufficiently large $n_0$.

\subsection{Bounding remaining terms}
\label{sec:bounding_remaining}
Using eqns.~\eqref{eq_f_var} and \eqref{eq_f_covar}, we have
\begin{equation}
    \begin{split}
     \frac{X^\top X}{n_0} \odot \pa{\bar{M} - M } - R &= \tilde{R} \pa{\frac{X^\top X}{n_0} - I} \odot I +  \frac{\xi\xi''}{2} \frac{X^\top X}{n_0} \odot \pa{ \mathbf{e} \mathbf{1}^\top + \mathbf{1}\mathbf{e}^\top }\odot \pa{\mathbf{1}\mathbf{1}^\top - I } \\
    &\quad +   \zeta' \frac{X^\top X}{n_0}\odot \frac{X^\top X}{n_0} \odot \pa{\mathbf{1}\mathbf{1}^\top - I } - R + E,
    \end{split}
\end{equation}
where $E$'s diagonal entries are $\cal{O}(n_0^{2\e-1})$ and off-diagonal entries are 
$\cal{O}(n_0^{3\e-3/2})$. Taking the terms one by one, we first bound
\eq{\label{eq_final_bound2}
    \norma{ \tilde{R} \pa{\frac{X^\top X}{n_0} - I} \odot I } = \sup_a \absa{ \tilde{R} \pa{\frac{X_{:a}^\top X_{:a}}{n_0} - 1}  } = \cal{O}(n^{\e-1/2}) 
}
Next, we bound
\eq{\label{eq_quartic_X}
    \norma{ \frac{\xi\xi''}{2} \frac{X^\top X}{n_0} \odot \pa{ \mathbf{e} \mathbf{1}^\top + \mathbf{1}\mathbf{e}^\top } \odot \pa{\mathbf{1}\mathbf{1}^\top - I } } \leq \cal{O}(n_0^{\e-1/2}).
}
Eqn.~\eqref{eq_quartic_X} can be demonstrated by taking the 4th power of the trace as in \cite{el2010spectrum}. This is expected, since the entries are mean zero and have variance order $\cal{O}(n_0^{-1})$. Proving the spectral bound is a straightforward calculation using the independence of the entries of $X$, but we avoid details here. The final term can also be bounded in this way, yielding,
\eq{
    \norma{\zeta' \frac{X^\top X}{n_0}\odot \frac{X^\top X}{n_0} \odot \pa{\mathbf{1}\mathbf{1}^\top - I } - R} = \cal{O}(n_0^{\e-1/2}).
}
The inclusion of the matrix $R$ is necessary, due to the nonzero mean of the entries. See \cite{el2010spectrum} for an example of this calculation.

Similarly using the assumptions on $X$, we can bound the remaining diagonal matrix of eqn.~\eqref{eq_k_diff} as follows
\al{
    \norma{\pa{\frac{X^\top X}{n_0} \odot M - K_\text{simp} }} &= (\eta'-\zeta)\norma{\diag(X^\top X/n_0) - I} \nonumber\\
    &= (\eta'-\zeta)\sup_a \absa{ \frac{1}{n_0} \sum_k X_{ka}^2 - 1 } \nonumber\\
    & = \cal{O}(n^{\e-1/2}).\label{eq_final_bound}
}
Summing our bounds on $\Delta$ and eqns.~\eqref{eq_final_bound2}-\eqref{eq_final_bound} completes the proof of eqn.~\eqref{eq_spectral}.

\section{Gaussian equivalents}

In this section we discuss the key arguments for existence of Gaussian equivalents and the linearizations of Sec.~\ref{sec_gaussian}. As all the main elements of this argument have been established elsewhere, here we just provide the main intuitions and refer to prior work for the details.

Many of the statistics of random matrices are universal, that is, their limiting behavior as the matrix gets larger is insensitive to the detailed properties of their entries' distributions. Considerable work has gone into demonstrating universality for an increasingly large class of random matrices and a growing number of detailed statistics. In our case, the test loss is a global measurement of several random matrices. This perspective gives some intuition for why we are able to replace many of the intractable terms in the expressions we analyze with tractable terms, which only need to match quite superficial properties of the distributions to ensure the limiting test loss is the same.

In Secs.~\ref{sec_exact_asymptotics_train} and \ref{sec_exact_asymptotics}, we use this replacement strategy in two distinct situations. The first is for terms of the form
\eq{\label{eq_con_1}
    \tr(AB) = \sum_{ij}A_{ij}B_{ji},
}
for deterministic $A$ and random $B$. Under assumptions on $A$ and $B$, standard concentration inequalities can be used to describe the limiting behavior of sums like eqn.~\eqref{eq_con_1}. In our setting, one finds that this behavior only depends on the the low-order moments of $B$. By matching these low-order moments with Gaussian random variables, we can replace $B$ with a Gaussian random matrix with the same limiting behavior. Note, often $A$ is not actually deterministic, we are simply conditioning on it and only considering the randomness in $B$. The approach is suitable for determining the average behavior of eqn.~\eqref{eq_con_1} when we have control over the (weak) correlations in the entries of $A$ and $B$. Linearizing the matrices $A$ and $B$ in this setting is just a convenient bookkeeping device for performing these computations.

When one of the matrices in eqn.~\eqref{eq_con_1} is inverted, the situation is more complex, and indeed this is the case for the kernel matrix $K$ in expressions for the training and test loss. To apply the linear pencil algorithm, we have to replace the NTK in all expressions with a linearized version (see eqn.~\eqref{eq_K_simp}), which is a rational expression of the  i.i.d. Gaussian matrices, $X$, $W_1$, \emph{etc.} In Sec.~\ref{sec_lin_ntk}, we bounded the difference between the first-layer kernel and its linearization, thus removing the Hadamard product structure. It remains to linearize the second-layer kernel, \emph{i.e.} linearize $F$. This has been discussed in previous works, see \cite{mei2019generalization, adlam2019random,peche2019note,benigni2019eigenvalue}. 

It should be expected that a linearized version of $F$ will lead to the same asymptotic statistics due to some very general results on the limiting behavior of expressions of the form,
\eq{\label{eq_con_2}
    \tr\pa{A\frac{1}{B-zI}},
}
where $A$ is symmetric and $z\in\C^{+}$. The resolvent matrix $(B-z)^{-1}$ is intimately related to the spectral properties of $B$. Recently, isotropic results for quite general $A$ have been developed for matrices with correlated entries, which show that under certain assumptions the limiting behavior of eqn.~\eqref{eq_con_2} depends only on the low-order moments of $B$. Specifically, the limiting behavior of eqn.~\eqref{eq_con_2} is described by the matrix Dyson equation in many cases. For a summary of these results and related topics see e.g.~\cite{erdos2019matrix}. While we do not explicitly show the correlation structure of $K$ meets the conditions known to suffice for the matrix Dyson equation, the assumptions in Sec.~\ref{sec_prelim} imply that the correlations between entries of $K$ are weak, which is the essential ingredient.

Finding Gaussian equivalents for $A$ and $B$ in expressions like eqns.~\eqref{eq_con_1} and \eqref{eq_con_2} is relatively simple in our case. We encounter terms for which the matrix $B$ depends on some other random matrix $C$ through a coordinate-wise nonlinear function $f(C)$. For such cases, Taylor expanding the function $f$ is the key tool to finding these equivalents (see e.g.~\cite{adlam2019random} for more details on this type of approach).

\section{Exact asymptotics for the training loss}
\label{sec_exact_asymptotics_train}
\subsection{Decomposition of terms}
The model's predictions on the training set, $\hat{y}(X)$, take a simple form,
\begin{align}
\hat{y}(X) &= N_0(X) + (Y - N_0(X))K^{-1}K(X,X)\\
&= Y - \gamma (Y-N_0(X))K^{-1}\,.
\end{align}
The expected training loss can be written as,
\begin{align}
\Etrain &= \frac{1}{m}\mathbb{E}_{}\tr\big((Y - \hat{y}(X))(Y - \hat{y}(X))^\top\big)\\
&= \frac{\gamma^2}{m}\mathbb{E}_{}\tr\big((Y-N_0(X))^\top (Y-N_0(X))K^{-2}\big)\\
&= \tE_1 + \nu \tE_2
\end{align}
where $\nu=0$ with centering and $\nu=1$ without it and,
\begin{align}
T_1 &= \frac{\gamma^2}{m}\mathbb{E}_{}\tr(Y^\top Y K^{-2})\\
\label{eq_def_T2} T_2 &= \frac{\gamma^2}{m}\E\tr(N_0(X)^\top N_0(X) K^{-2})\,.
\end{align}
Note we can suppress the terms linear in $N_0$ since they vanish in expectation owing to the linear dependence on the mean-zero random variable $\omega$. Here $K = K(X,X) + \gamma I_m$ is the linearized NTK and is given by,
\eq{\label{eq_K}
    K = \sigma_{W_2}^2 \big[(\eta' - \zeta)I_m + \frac{\zeta X^\top\! X}{n_0}\big] + \frac{F^\top \!F}{n_1} + \gamma I_m\,.
}
This substitution can be justified using the result of Sec.~\ref{sec_lin_ntk}:
\al{
    &\absa{\frac{\gamma^2}{m}\mathbb{E}_{} \tr\pa{Y^\top Y K_\text{simp}^{-2}}) - \frac{\gamma^2}{m}\mathbb{E}_{}\tr(Y^\top Y K^{-2}) } = \absa{ \frac{\gamma^2}{m}\mathbb{E}_{}  Y \pa{K_\text{simp}^{-2} - K^{-2}} Y^\top} \\
    &\quad \leq \frac{\gamma^2\zeta'}{mn_0}\E \abs{ Y\mathbf{1}^\top}_2 + \frac{\gamma^2}{m} \E \norm{Y}^2_2 \norm{R+K_\text{simp}^{-2} - K^{-2}} = o(1).
}
Eqn.~\eqref{eq_def_T2} is similar.

Note that taking the expectation over $W_2$ in eqn.~\eqref{eq_def_T2} and eqn.~\eqref{eq_K} yields
\eq{
    \E_{W_2} N_0(X)^\top N_0(X) = \sigma_{W_2}^2 K -\sigma_{W_2}^2 \big[\sigma_{W_2}^2\big(\eta' - \zeta\big) + \gamma \big]I_m -\sigma_{W_2}^4 \frac{\zeta X^\top\! X}{n_0}\,,
}
since $\E_{W_2} N_0(X)^\top N_0(X) = \sigma_{W_2}^2/n_1 F^T F$.

Next we recall the substitution~\eqref{eq_Ylin},
\eq{\label{eq_sub_Y}
Y \to Y^{\text{lin}} = \frac{1}{\sqrt{n_0 n_\textsc{t}}} \omega\Omega X + \mathcal{E}\,,
}
which can be used to calculate the expectation over $\omega$ and $\Omega$ to leading order (\emph{i.e.} with remainder terms $o(1)$) using the approach of eqn.~\eqref{eq_con_1}. Concretely,
\eq{
    \frac{\gamma^2}{m} \E_{\omega,\Omega,\mathcal{E}} \tr(Y^\top Y K^{-2}) = \frac{\gamma^2}{m} \E_{\omega,\Omega,\mathcal{E}} \tr(\pa{Y^{\text{lin}}}^\top Y^{\text{lin}} K^{-2})  +o(1)= \frac{\gamma^2}{m} \tr\qa{\pa{ \frac{1}{n_0}X^\top X + \sigma_{\varepsilon}^2 I_m } K^{-2}}+o(1).
}

Putting these pieces together, we can write for $\tau_1 = \tau_1(\gamma)$ and $\tau_2 = \tau_2(\gamma)$,
\begin{align}
    T_1 &= -\gamma^2(\sigma_\varepsilon^2 \tau_1' + \tau_2')\\
    T_2 &= \sigma_{W_2}^2\gamma^2\left(\tau_1 + (\sigma_{W_2}^2(\eta'-\zeta) + \gamma)\tau_1'+\sigma_{W_2}^2\zeta \tau_2'\right)\label{eq:T2}\,,
\end{align}
where,
\begin{equation}
\label{eqn:tau}
    \tau_1 = \frac{1}{m}\E\tr(K^{-1})\,,\quad\text{and}\quad\tau_2 = \frac{1}{m}\E\tr(\frac{1}{n_0}X^\top XK^{-1})\,.
\end{equation}
Self-consistent equations for $\tau_1$ and $\tau_2$ can be computed using the resolvent method, as was done in~\cite{adlam2019random} for the case of $\sigma_{W_2} = 0$. In order to pave the way for the analysis of the test error, we instead demonstrate how to compute these traces using operator-valued free probability.

\begin{remark}
In the remainder of this section, and in Sec.~\ref{sec_exact_asymptotics}, we assume at times that $\sigma$ is non-linear (so that $\eta'>\zeta$ and $\eta > \zeta$) and/or $\gamma > 0$ in order that certain denominator factors are non-zero.  The linear and/or ridgeless cases can be obtained by limits of our general results, or through special cases of the pertinent intermediate formulas.
\end{remark}

\subsection{Linear pencils}
\label{sec:pencil1}
To begin, we construct linear pencils for $\tau_1$ and $\tau_2$. Using the linearization eqn.~(\ref{eq_Flin}), a straightforward block-matrix inversion confirms that
\begin{equation}
    \tau_1 = \E\tr([Q_T^{-1}]_{1,1})\,\quad\text{and}\quad \E\tau_2 = \tr([Q_T^{-1}]_{2,4})\,, 
\end{equation}
where,
\begin{equation}
    Q_T = \left(
\begin{array}{cccc}
 I_m \left(\gamma +\sigma_{W_2}^2 \left(\eta '-\zeta \right)\right) & \frac{\zeta  X^\top \sigma_{W_2}^2}{n_0} & \frac{\sqrt{\eta -\zeta } \Theta_F^\top}{n_1} & \frac{\sqrt{\zeta } X^\top}{\sqrt{n_0} n_1} \\
 -X & I_{n_0} & 0 & 0 \\
 -\sqrt{\eta -\zeta } \Theta_F & -\frac{\sqrt{\zeta } W_1}{\sqrt{n_0}} & I_{n_1} & 0 \\
 0 & 0 & \frac{\sqrt{\zeta } \psi  W_1^\top}{\sqrt{n_0} \phi } & -\frac{\sqrt{\zeta } \psi  I_{n_0}}{\sqrt{n_0} \phi } \\
\end{array}
\right)\,.
\end{equation}
The matrix $Q_T$ is not self-adjoint, but a self-adjoint representation can be obtained from it by doubling the dimensionality. In particular, letting
\eq{
\bar{Q}_T = \begin{pmatrix}
0 & Q_T^\top \\
Q_T & 0
\end{pmatrix}\,,
}
we have,
\eq{
\tau_1 = \E\tr([\bar{Q}_T^{-1}]_{1,5})\,,\quad\text{and}\quad \E\tr([\bar{Q}_T^{-1}]_{2,8})\,.
}
Observe that $\bar{Q}_{T}$ is a self-adjoint matrix whose blocks are either constants or proportional to one of $\{X,X^\top,W_1,W_1^\top,\Theta_F,\Theta_F^\top\}$; let us denote the constant terms as $Z$. As such, we can directly utilize the results of~\cite{far2006spectra,mingo2017free} to compute the necessary traces.

\subsection{Operator-valued Stieltjes transform}
The traces can be extracted from the operator-valued Stieltjes transform $G:M_d(\mathbb{C})^+\to M_d(\mathbb{C})^+$, which is a solution of the equation,
\eq{\label{eqn_Geqn_train}
ZG = I_d + \eta(G) G\,,
}
where $d$ is the number of blocks, $\eta: M_d(\mathbb{C})\to M_d(\mathbb{C})$ defined by
\eq{\label{eqn_Deqn}
[\eta(D)]_{ij} = \sum_{kl} \sigma(i,k;l,j) \alpha_k D_{kl} \,,
}
where $\alpha_k$ is dimensionality of the $k$th block and $\sigma(i,k;l,k)$ denotes the covariance between the entries of the blocks $ij$ block of $\bar{Q}$ and entries of the $kl$ block of $\bar{Q}$. Eqn.~\eqref{eqn_Geqn_train} may admit many solutions, but there is a unique solution such that $\text{Im}G \succ 0$ for $\text{Im} Z\succ 0$.

The constants $Z$, the entries of $\sigma$, and therefore the equations~\eqref{eqn_Deqn} are manifest by inspection of the block matrix representation for $\bar{Q}_T$. Although the matrix representation of the equations is too large to reproduce here, we can nevertheless extract the equations satisfied by each entry of $G$.

The equations satisfied by the operator-valued Stieltjes transform $G$ of $\bar{Q}_T$ induce the following structure on $G$,
\eq{
G = \begin{pmatrix} 0 & G_{12} \\ G_{12}^\top & 0 \end{pmatrix}\,,
}\,
where,
\eq{
G_{12} = \left(
\begin{array}{cccc}
 \tau_1 & 0 & 0 & 0 \\
 0 & g_3 & 0 & \tau_2 \\
 0 & 0 & g_4 & 0 \\
 0 & g_6 & 0 & g_5 \\
\end{array}
\right)
}
and the independent entry-wise component functions $g_i$, $\tau_1$ and $\tau_2$ satisfy the following system of polynomial equations,
\begin{align}
0 &= \sqrt{\zeta } g_6 \psi -\zeta  g_3 g_4 \sqrt{n_0}\\
0 &= \sqrt{\zeta } \psi  \big(\tau _2-g_3 \tau _1\big)\\
0 &= \sqrt{\zeta } \psi  \big(g_5-g_6 \tau _1\big)+\sqrt{n_0} \phi\\
0 &= -\zeta  g_4 g_5-g_6 \big(\zeta  \tau _1 \sigma _{W_2}^2+\phi \big)\\
0 &= \sqrt{\zeta } g_5 \psi +\sqrt{n_0} \big(\phi -\zeta  g_4 \tau _2\big)\\
0 &= \phi -g_4 \big(\tau _1 \psi  (\eta -\zeta )+\zeta  \tau _2 \psi +\phi \big)\\
0 &= -\zeta  g_4 \tau _2-g_3 \big(\zeta  \tau _1 \sigma _{W_2}^2+\phi \big)+\phi\\
0 &= -\sqrt{\zeta } g_5 \tau _1 \psi -\sqrt{n_0} \tau _2 \big(\zeta  \tau _1 \sigma _{W_2}^2+\phi \big)\\
0 &= \sqrt{n_0} \big(\phi -g_3 \big(\zeta  \tau _1 \sigma _{W_2}^2+\phi \big)\big)-\sqrt{\zeta } g_6 \tau _1 \psi\\
0 &= \sqrt{n_0} \big(1-\tau _1 \big(\gamma +g_4 (\eta -\zeta )+\sigma _{W_2}^2 \big(\eta '+\zeta  \big(g_3-1\big)\big)\big)\big)-\sqrt{\zeta } g_6 \tau _1 \psi
\,.
\end{align}
It is straightforward algebra to eliminate $g_3,g_4,g_5$ and $g_6$ from the above equations. A simple set of equations for $\tau_1$ and $\tau_2$ follows,
\begin{align}
 0&=\phi\left(\zeta  \tau_2 \tau_1+ \phi(\tau_2 -\tau_1) \right)+\zeta  \tau_1 \tau_2
  \psi  \left(\gamma \tau_1-1\right) +\zeta  \tau_1 \tau_2 \sigma _{W_2}^2
  \left(\zeta  \left(\tau_2-\tau_1\right) \psi +\tau_1 \psi  \eta '+\phi \right)\label{eq:tau1}\\
  0&=\zeta \tau_1^2 \tau_2 \left(\eta'-\eta\right) \sigma _{W_2}^2+\zeta  \tau_1 \tau_2 \left(\gamma \tau_1 -1\right) - \left(\tau_2-\tau_1\right) \phi  \left(\zeta  \left(\tau_2-\tau_1\right)+\eta  \tau_1\right)\label{eq:tau2}\,.
\end{align}
Although these equations admit multiple solutions, the general results of~\cite{far2006spectra,mingo2017free} guarantee that the correct root is given by the unique solutions $\tau_1,\tau_2: \mathbb{C}^+\to\mathbb{C}^+$ which are analytic in the upper half-plane.

It will prove useful to obtain expressions for $\tau_1'(\gamma)$ and $\tau_2'(\gamma)$. By differentiating eqns.~(\ref{eq:tau1}) and~(\ref{eq:tau2}) with respect to $\gamma$, we find
\begin{align}
\tau_1' &= -\frac{\zeta ^2 \tau _2^2 \big(\psi  \tilde{\tau }_1^2-\phi ^2\big)}{\psi  \tilde{\tau }_1^2 \big(\zeta ^2 \big(\tilde{\tau }_2+1\big){}^2+\phi  \big(\zeta  \tilde{\tau }_2+\eta \big) \big(\zeta  \tilde{\tau }_2 \big(2 \tilde{\tau }_2+3\big)+\eta \big)\big)+\zeta ^2 \phi ^2 \big(\tilde{\tau }_2+1\big){}^2 \big(\phi  \tilde{\tau }_2^2-1\big)}\label{eq:dtau1}\\
\tau_2'&=-\frac{\zeta  \tau _2^2 \big(\psi  \tilde{\tau }_1^2 (\zeta -\eta )-\zeta  \phi ^2 \big(\tilde{\tau }_2+1\big){}^2\big)}{\psi  \tilde{\tau }_1^2 \big(\zeta ^2 \big(\tilde{\tau }_2+1\big){}^2+\phi  \big(\zeta  \tilde{\tau }_2+\eta \big) \big(\zeta  \tilde{\tau }_2 \big(2 \tilde{\tau }_2+3\big)+\eta \big)\big)+\zeta ^2 \phi ^2 \big(\tilde{\tau }_2+1\big){}^2 \big(\phi  \tilde{\tau }_2^2-1\big)}\label{eq:dtau2}\,,
\end{align}
where we have introduced some auxiliary variables to ease the presentation,
\begin{equation}
\label{eq:ttau}
    \ttau_1 = \sigma_{W_2}^2 \zeta \tau_2 + \phi \ttau_2\,\quad\text{and}\quad \ttau_2 = -1 + \tau_2/\tau_1\,.
\end{equation}
\section{Exact asymptotics for the test loss}
\label{sec_exact_asymptotics}
\subsection{Decomposition of terms}
As described in Sec.~\ref{sec_ovfp}, the test loss can be written as,
\eq{
\label{eq:Etest}
\Etest = \mathbb{E}(y - \hat{y}(\bfx))^2  = E_1 + E_2 + E_3
}
with
\begin{align}
E_1 &= \mathbb{E}\tr(y(\bfx)y(\bfx)^\top) + \mathbb{E}\tr(N_0(\bfx)N_0(\bfx)^\top)\\
E_2 &= -2\mathbb{E}\tr(K_\bfx^\top K^{-1}Y^\top y(\bfx)) - 2\mathbb{E}\tr(K_\bfx^\top K^{-1}N_0(X)^\top N_0(\bfx))\\
E_3 &= \mathbb{E}\tr(K_\bfx^\top K^{-1}Y^\top Y K^{-1} K_\bfx) +  \mathbb{E}\tr(K_\bfx^\top K^{-1}N_0(X)^\top N_0(X) K^{-1} K_\bfx) \,.
\end{align}
As in Sec.~\ref{sec_exact_asymptotics_train}, we suppress the terms linear in $\omega$ as they vanish in expectation. The Neural Tangent Kernels ${K = K(X,X) + \gamma I}$ and $K_\bfx = K(X,\bfx)$ are given by,
\eq{\label{eq_K_and_Kx}
    K = \sigma_{W_2}^2 \big[(\eta' - \zeta)I_m + \frac{\zeta X^\top\! X}{n_0}\big] + \frac{F^\top \!F}{n_1} + \gamma I_m\,,\qquad\text{and}\qquad K_\bfx = \frac{\sigma_{W_2}^2 \zeta}{n_0}X^\top \bfx + \frac{1}{n_1} F^\top f\,,
}
where the substitution for the linearized NTK is justified as in Sec.~\ref{sec_exact_asymptotics_train} using the spectral norm bound of Sec.~\ref{sec_lin_ntk}.

Using the cyclicity and linearity of the trace, the expectation over $\bfx$ requires the computation of
\eq{
\mathbb{E}_{\bfx}K_\bfx K_\bfx^\top \,,\qquad \mathbb{E}_{\bfx} y(\bfx)K_\bfx^\top\,,\qquad \mathbb{E}_{\bfx} y(\bfx) y(\bfx)^\top\,,\qquad \mathbb{E}_{\bfx} N_0(\bfx)K_\bfx^\top\,,  \qquad\text{and}\quad \mathbb{E}_{\bfx} N_0(\bfx) N_0(\bfx)^\top\,.
}
As described in Sec.~\ref{sec_gaussian}, without loss of generality we can consider the case of a linear teacher, so that $\eta_\textsc{t} = \zeta_\textsc{t} = 1$ and ~\eqref{eq_ylin} and~\eqref{eq_flin} become
\eq{\label{eq_sub_y_f}
y \to y^{\text{lin}} = \frac{\sqrt{\zeta_\textsc{t}}}{\sqrt{n_0 n_\textsc{t}}} \omega\Omega \bfx + \sqrt{\eta_\textsc{t} - \zeta_\textsc{t}} \frac{1}{\sqrt{n_\textsc{t}}}\omega\theta_y = \frac{
1}{\sqrt{n_0 n_\textsc{t}}}\omega\Omega \bfx\,\qquad\text{and}\qquad f \to f^{\text{lin}} = \frac{\sqrt{\zeta}}{\sqrt{n_0}} W_1 \bfx + \sqrt{\eta - \zeta} \theta_f\, .
}
Using these substitutions, the expectations over $\bfx$ are now trivial and we readily find,
\begin{align}
   \mathbb{E}_\bfx K_\bfx K_\bfx^\top & = \frac{\sigma_{W_2}^4 \zeta^2}{n_0^2} X^\top X + \frac{\sigma_{W_2}^2 \zeta^{3/2}}{n_0^{3/2} n_1}(X^\top W_1^T F + F^\top W_1 X) + \frac{1}{n_1^2}F^\top\big(\frac{\zeta}{n_0} W_1 W_1^\top + (\eta - \zeta) I_{n_1}\big)F\\
\mathbb{E}_{\bfx} y(\bfx)K_\bfx^\top & =   \frac{\sigma_{W_2}^2 \zeta}{n_0^{3/2} \sqrt{n_\textsc{t}}} \omega\Omega X + \frac{\sqrt{\zeta}}{n_0n_1 \sqrt{n_\textsc{t}} } \omega\Omega  W_1^\top F\\
\mathbb{E}_{\bfx} y(\bfx)y(\bfx)^\top &= \frac{1}{n_0 n_\textsc{t}} \omega\Omega \Omega^\top\omega^\top\\
\mathbb{E}_{\bfx} N_0(\bfx)K_\bfx^\top & = \frac{\sigma_{W_2}^2 \zeta^{3/2}}{n_0^{3/2}\sqrt{n_1}} W_2W_1X + \frac{1}{n_1^{3/2}}W_2\big(\frac{\zeta}{n_0}W_1W_1^\top+(\eta-\zeta)I_{n_1}\big)F\\ 
\mathbb{E}_{\bfx} \tr(N_0(\bfx) N_0(\bfx)^\top) & = \sigma_{W_2}^2\eta\,.
\end{align}
One may interpret the substitutions in eqn.~\eqref{eq_sub_y_f} as a tool to calculate the expectations above to leading order as it leads to terms like eqn.~\eqref{eq_con_1}. Next we recall the substitution~\eqref{eq_sub_Y},
\eq{
Y \to Y^{\text{lin}} = \frac{1}{\sqrt{n_0 n_\textsc{t}}} \omega\Omega X + \mathcal{E}\,.
}
As above, we consider the leading order behavior with respect to the random variables $\omega$, $\Omega$, and $W_2$ using eqn.~\eqref{eq_con_1} to find
\begin{align}
    \E_{\omega,\Omega,\mathcal{E}}\qa{Y^\top Y} &= \frac{1}{n_0}X^\top X + \sigma_{\varepsilon}^2 I_m\\
    \E_{\omega,\Omega,\mathcal{E},W_2}\qa{Y^\top \mathbb{E}_{\bfx} y(\bfx)K_\bfx^\top} &= \frac{\sigma_{W_2}^2 \zeta}{n_0^2} X^\top X + \frac{\sqrt{\zeta}}{n_0^{3/2}n_1} X^\top  W_1^\top F\\
    \E_{W_2}\qa{N_0(X)^\top N_0(X)} &= \frac{\sigma_{W_2}^2}{n_1}F^\top F\\
    \E_{W_2}\qa{N_0(X)^\top \mathbb{E}_{\bfx} N_0(\bfx)K_\bfx^\top} &= \frac{\sigma_{W_2}^4 \zeta^{3/2}}{n_0^{3/2}n_1} F^\top W_1X + \frac{\sigma_{W_2}^2}{n_1^2}F^\top\big(\frac{\zeta}{n_0}W_1W_1^\top+(\eta-\zeta)I_{n_1}\big)F\,.
\end{align}
Using~\eqref{eq_Flin},
\eq{\label{eq_sub_F}
F \to F^{\text{lin}} = \frac{\sqrt{\zeta}}{\sqrt{n_0}} W_1 X + \sqrt{\eta - \zeta} \Theta_F\,,
}
we can write,
\eq{
\frac{\sqrt{\zeta}}{\sqrt{n_0}} F^\top W_1 X + \frac{\sqrt{\zeta}}{\sqrt{n_0}} X^\top W_1^\top F = F^\top F + \frac{\zeta}{n_0}X^\top W_1^\top W_1 X - (\eta-\zeta)\Theta_F^\top\Theta_F\,.
}
Putting these pieces together, we have
\begin{align}
    E_1 & = 1 + \nu \sigma_{W_2}^2 \eta\label{eqn:E1}\\
    E_2 &= E_{21} + \nu E_{22}\label{eqn:E2}\\
    E_3 &= E_{31} + E_{32} + \nu E_{33}\label{eqn:E3}\,,
\end{align}
where $\nu = 0$ with centering and $\nu = 1$ without it,
\begin{align}
E_{21} & = -\E\tr\bigg(2\frac{\sigma_{W_2}^2 \zeta}{n_0^2}X K^{-1} X^\top + \frac{1}{n_0 n_1} F K^{-1} F^\top + \frac{\zeta}{n_0^2 n_1} W_1 X K^{-1}X^\top W_1^\top - \frac{\eta-\zeta}{n_0 n_1}\Theta_F K^{-1}\Theta_F^\top\bigg)\label{eq_E2}\\
E_{22} & = -\frac{2 \sigma_{W_2}^2}{n_1}\E\tr\bigg( \frac{\sigma_{W_2}^2 \zeta^{3/2}}{n_0^{3/2}} K^{-1}F^\top W_1 X +\frac{\zeta}{n_0 n_1} K^{-1}F^\top W_1W_1^\top F + \frac{\eta-\zeta}{n_1}K^{-1}F^\top F \bigg)\\
E_{31} &= \sigma_{\varepsilon}^2 \E\tr\left(K^{-1} \Sigma_3 K^{-1}\right)\\
E_{32} &= \frac{1}{n_0} \E\tr\left(X K^{-1} \Sigma_3 K^{-1}X^\top\right)\\
E_{33} &= \frac{\sigma_{W_2}^2}{n_1} \E\tr\left(F K^{-1} \Sigma_3 K^{-1}F^\top\right)\,,
\end{align}
and,
\eq{
\Sigma_3 = \frac{\sigma_{W_2}^4 \zeta^2}{n_0^2} X^\top X + \big(\frac{\sigma_{W_2}^2 \zeta}{n_0 n_1} + \frac{\eta-\zeta}{n_1^2}\big)F^\top F + \frac{\zeta}{n_0n_1^2}F^\top W_1 W_1^\top F + \frac{\sigma_{W_2}^2 \zeta^2}{n_0^2 n_1} X^\top W_1^\top W_1 X - \frac{\sigma_{W_2}^2 \zeta(\eta - \zeta)}{n_0 n_1}\Theta_F^\top \Theta_F\,.
}
\subsection{Linear pencils}
\label{sec:pencil2}
Repeated application of the Schur complement formula for block matrix inversion establishes the following representations for $E_{21}, E_{22}, E_{31}, E_{32}, E_{33}.$
\subsubsection{$E_{21}$}
A linear pencil for $E_{21}$ follows from the representation,
\eq{
E_{21} = \E\tr(U_{21}^T Q_{21}^{-1} V_{21})\,,
}
where,
\begin{align}
    U_{21}^T  &= \left(
\begin{array}{cccccccccccccc}
 0 & -\frac{2 \zeta  I_{n_0} \sigma_{W_2}^2}{n_0} & 0 & 0 & 0 & \frac{(\eta -\zeta ) I_{n_1}}{n_0} & 0 & 0 & 0 & 0 & 0 & -\frac{I_{n_1}}{n_0} & 0 & 0 
\end{array}
\right)\\
V_{21}^T & = 
\left(
\begin{array}{cccccccccccccc}
 0 & 0 & 0 & -\frac{\sqrt{n_0} n_1 I_{n_0}}{\sqrt{\zeta }} & 0 & 0 & 0 & 0 & 0 & I_{n_1} & 0 & 0 & 0 & 0 \\
\end{array}
\right)
\end{align}
and,
\eq{
Q_{21} =
\begin{pmatrix}
Q_{21}^{11} & 0 & 0 \\
0 & Q_{21}^{22} & Q_{21}^{23} \\
0 & 0 & Q_{21}^{33} \\ 
\end{pmatrix}
}
with,
\begin{align}
Q_{21}^{11} &=
\begin{pmatrix}
 I_m \left(\gamma +\sigma_{W_2}^2 \left(\eta '-\zeta \right)\right) & \frac{\zeta  X^\top \sigma_{W_2}^2}{n_0} & \frac{\sqrt{\eta -\zeta } \Theta_F^\top}{n_1} & \frac{\sqrt{\zeta } X^\top}{\sqrt{n_0} n_1}  \\
 -X & I_{n_0} & 0 & 0  \\
 -\sqrt{\eta -\zeta } \Theta_F & -\frac{\sqrt{\zeta } W_1}{\sqrt{n_0}} & I_{n_1} & 0\\
 0 & 0 & -W_1^\top & I_{n_0}
\end{pmatrix}\\
Q_{21}^{22} &=
\begin{pmatrix}
 I_m \left(\gamma +\sigma_{W_2}^2 \left(\eta '-\zeta \right)\right) & 0 & \frac{\zeta  X^\top \sigma_{W_2}^2}{n_0} & \frac{\sqrt{\eta -\zeta } \Theta_F^\top}{n_1} & \frac{\sqrt{\zeta } X^\top}{\sqrt{n_0} n_1}\\
-\Theta_F & I_{n_1} & -\frac{\sqrt{\zeta } W_1}{\sqrt{n_0} \sqrt{\eta -\zeta }} & 0 & 0\\
-X & 0 & I_{n_0} & 0 & 0 \\
-\sqrt{\eta -\zeta } \Theta_F & 0 & -\frac{\sqrt{\zeta } W_1}{\sqrt{n_0}} & I_{n_1}\\
0 & 0 & 0 & -W_1^\top & I_{n_0}\\
\end{pmatrix}\\
Q_{21}^{23} &=
\begin{pmatrix}
-\Theta_F^\top & 0 & 0 & 0 & 0 \\
 0 & 0 & 0 & \frac{\sqrt{\zeta } W_1}{\sqrt{n_0} (\eta -\zeta )} & 0 \\
 0 & 0 & 0 & 0 & 0 \\
 0 & 0 & 0 & 0 & 0 \\
 0 & 0 & 0 & 0 & 0 \\
 I_{n_1} & 0 & 0 & 0 & 0 \\
\end{pmatrix}\\
Q_{21}^{33} &=
\begin{pmatrix}
-\sqrt{\eta -\zeta } \Theta_F^\top & I_m \left(\gamma +\sigma_{W_2}^2 \left(\eta '-\zeta \right)\right) & \frac{\sqrt{\eta -\zeta } \Theta_F^\top}{n_1} & \frac{\zeta  X^\top \sigma_{W_2}^2}{n_0} & \frac{\sqrt{\zeta } X^\top}{\sqrt{n_0} n_1} \\
 0 & -\sqrt{\eta -\zeta } \Theta_F & I_{n_1} & -\frac{\sqrt{\zeta } W_1}{\sqrt{n_0}} & 0 \\
 0 & -X & 0 & I_{n_0} & 0 \\
 n_1 W_1^\top & 0 & -W_1^\top & 0 & I_{n_0}
\end{pmatrix}\,.
\end{align}

\subsubsection{$E_{22}$}
A linear pencil for $E_{22}$ follows from the representation,
\eq{
E_{22} = \E\tr(U_{22}^T Q_{22}^{-1} V_{22})\,,
}
where,
\begin{align}
    U_{22}^T  &= 
\begin{pmatrix}
0 & -\frac{2 \sqrt{\zeta } I_{n_1} \sigma_{W_2}^2 \left(n_0 (\eta -\zeta )+\zeta  n_1 \sigma_{W_2}^2\right)}{n_0^{3/2} n_1} & 0 & \frac{2 (\zeta -\eta ) I_{n_1} \sigma_{W_2}^2}{n_1} & 0 & 0 & 0
\end{pmatrix}\\
V_{22}^T & = 
\begin{pmatrix}
 0 & 0 & 0 & 0 & 0 & - n_1 I_{n_1} & 0
\end{pmatrix}
\end{align}
and,
\eq{
Q_{22} =
\left(
\begin{smallmatrix}
I_{n_0} & 0 & -X & 0 & 0 & 0 & 0 \\
 -W_1 & I_{n_1} & 0 & 0 & -\frac{\sqrt{n_0} W_1}{\sqrt{\zeta } n_1 \sigma_{W_2}^2} & 0 & 0 \\
 \frac{\zeta  X^\top \sigma_{W_2}^2}{n_0} & 0 & I_m \left(\gamma +\sigma_{W_2}^2 \left(\eta '-\zeta \right)\right) & 0 & 0 & \frac{\sqrt{\eta -\zeta } \Theta_F^\top}{n_1} & \frac{\sqrt{\zeta } X^\top}{\sqrt{n_0} n_1} \\
 0 & 0 & -\sqrt{\eta -\zeta } \Theta_F & I_{n_1} & \frac{W_1}{n_1 \sigma_{W_2}^2} & 0 & 0 \\
 0 & -\frac{\sqrt{\zeta } W_1^\top}{\sqrt{n_0}} & 0 & -W_1^\top & I_{n_0} & 0 & 0 \\
 -\frac{\sqrt{\zeta } W_1}{\sqrt{n_0}} & 0 & -\sqrt{\eta -\zeta } \Theta_F & 0 & 0 & I_{n_1} & 0 \\
 0 & 0 & 0 & 0 & 0 & -W_1^\top & I_{n_0} 
\end{smallmatrix}
\right)\,.
}

\subsubsection{$E_{31}$}
A linear pencil for $E_{31}$ follows from the representation,
\eq{
E_{31} = \E\tr(U_{31}^T Q_{31}^{-1} V_{31})\,,
}
where,
\eq{
    U_{31}^T  = 
\begin{pmatrix}
m \sigma _ {\varepsilon }^2 I_m & 0 & 0 & 0 & 0 & 0 & 0 & \
0
\end{pmatrix}\,,\quad
V_{31}^T = 
\begin{pmatrix}
 0 & 0 & 0 & 0 & 0 & I_m & 0 & 0
\end{pmatrix}
}
and, for $\beta = \left(n_0 (\zeta -\eta )-\zeta  n_1 \sigma_{W_2}^2\right)$,
\eq{
Q_{31} =
\left(
\begin{smallmatrix}
I_m \left(\gamma +\sigma_{W_2}^2 \left(\eta '-\zeta \right)\right) & \frac{\zeta  X^\top \sigma_{W_2}^2}{n_0} & \frac{\sqrt{\eta -\zeta } \Theta_F^\top}{n_1} & \frac{\sqrt{\zeta } X^\top}{\sqrt{n_0} n_1} & -\frac{\zeta ^2 X^\top \sigma_{W_2}^4}{n_0^2} & 0 & \frac{\sqrt{\eta -\zeta } \Theta_F^\top \beta}{n_0 n_1^2} & \frac{\sqrt{\zeta } X^\top \beta}{n_0^{3/2} n_1^2} \\
 -X & I_{n_0} & 0 & 0 & 0 & 0 & 0 & 0 \\
 -\sqrt{\eta -\zeta } \Theta_F & -\frac{\sqrt{\zeta } W_1}{\sqrt{n_0}} & I_{n_1} & 0 & 0 & -\frac{\zeta  \sqrt{\eta -\zeta } \Theta_F \sigma_{W_2}^2}{n_0} & 0 & \frac{\zeta  W_1}{n_0 n_1} \\
 0 & 0 & -W_1^\top & I_{n_0} & 0 & 0 & \frac{\zeta  W_1^\top \sigma_{W_2}^2}{n_0} & 0 \\
 0 & 0 & 0 & 0 & I_{n_0} & -X & 0 & 0 \\
 0 & 0 & 0 & 0 & \frac{\zeta  X^\top \sigma_{W_2}^2}{n_0} & I_m \left(\gamma +\sigma_{W_2}^2 \left(\eta '-\zeta \right)\right) & \frac{\sqrt{\eta -\zeta } \Theta_F^\top}{n_1} & \frac{\sqrt{\zeta } X^\top}{\sqrt{n_0} n_1} \\
 0 & 0 & 0 & 0 & -\frac{\sqrt{\zeta } W_1}{\sqrt{n_0}} & -\sqrt{\eta -\zeta } \Theta_F & I_{n_1} & 0 \\
 0 & 0 & 0 & 0 & 0 & 0 & -W_1^\top & I_{n_0} 
\end{smallmatrix}
\right)\,.
}
\subsubsection{$E_{32}$}
A linear pencil for $E_{32}$ follows from the representation,
\eq{
E_{32} = \E\tr(U_{32}^T Q_{32}^{-1} V_{32})\,,
}
where,
\eq{
    U_{32}^T  = 
\begin{pmatrix}
0 & I_{n_0} & 0 & 0 & 0 & 0 & 0 & 0 & 0
\end{pmatrix}\,,\quad
V_{32}^T = 
\begin{pmatrix}
 0 & 0 & 0 & 0 & 0 & 0 & 0 & 0 & -\frac{\sqrt{n_0} n_1 I_{n_0}}{\sqrt{\zeta }}
\end{pmatrix}
}
and, for $\beta = \left(n_0 (\zeta -\eta )-\zeta  n_1 \sigma_{W_2}^2\right)$
\eq{
Q_{32} =
\left(
\begin{smallmatrix}
I_m \left(\gamma +\sigma_{W_2}^2 \left(\eta '-\zeta \right)\right) & 0 & \frac{\zeta  X^\top \sigma_{W_2}^2}{n_0} & \frac{\sqrt{\eta -\zeta } \Theta_F^\top}{n_1} & \frac{\sqrt{\zeta } X^\top}{\sqrt{n_0} n_1} & -\frac{\zeta ^2 X^\top \sigma_{W_2}^4}{n_0^2} & 0 & \frac{\sqrt{\eta -\zeta } \Theta_F^\top \beta}{n_0 n_1^2} & 0 \\
 -X & I_{n_0} & 0 & 0 & 0 & 0 & 0 & 0 & 0 \\
 -X & 0 & I_{n_0} & 0 & 0 & 0 & 0 & \frac{\sqrt{\zeta } W_1^\top}{\sqrt{n_0} n_1} & 0 \\
 -\sqrt{\eta -\zeta } \Theta_F & 0 & -\frac{\sqrt{\zeta } W_1}{\sqrt{n_0}} & I_{n_1} & 0 & 0 & -\frac{\zeta  \sqrt{\eta -\zeta } \Theta_F \sigma_{W_2}^2}{n_0} & 0 & 0 \\
 0 & 0 & 0 & -W_1^\top & I_{n_0} & 0 & 0 & W_1^\top \left(\frac{\eta -\zeta }{n_1}+\frac{\zeta  \sigma_{W_2}^2}{n_0}\right) & 0 \\
 0 & 0 & 0 & 0 & 0 & I_{n_0} & -X & 0 & 0 \\
 0 & 0 & 0 & 0 & 0 & \frac{\zeta  X^\top \sigma_{W_2}^2}{n_0} & I_m \left(\gamma +\sigma_{W_2}^2 \left(\eta '-\zeta \right)\right) & \frac{\sqrt{\eta -\zeta } \Theta_F^\top}{n_1} & \frac{\sqrt{\zeta } X^\top}{\sqrt{n_0} n_1} \\
 0 & 0 & 0 & 0 & 0 & -\frac{\sqrt{\zeta } W_1}{\sqrt{n_0}} & -\sqrt{\eta -\zeta } \Theta_F & I_{n_1} & 0 \\
 0 & 0 & 0 & 0 & 0 & 0 & 0 & -W_1^\top & I_{n_0}
\end{smallmatrix}
\right)\,.
}

\subsubsection{$E_{33}$}
A linear pencil for $E_{33}$ follows from the representation,
\eq{
E_{33} =\E \tr(U_{33}^T Q_{33}^{-1} V_{33})\,,
}
where,
\begin{align}
    U_{33}^T  & = 
\left(
\begin{array}{ccccccccccc}
 0 & I_{n_1} \sigma_{W_2}^2 & 0 & 0 & 0 & 0 & 0 & 0 & 0 & 0 & 0 \\
\end{array}
\right)\\
V_{33}^T &= 
\left(
\begin{array}{ccccccccccc}
 0 & 0 & 0 & 0 & 0 & 0 & 0 & 0 & 0 & -n_1 I_{n_1} & 0 \\
\end{array}
\right)
\end{align}
and, for $\beta = \left(n_0 (\zeta -\eta )-\zeta  n_1 \sigma_{W_2}^2\right)$,
\eq{
Q_{33} =
\left(
\begin{smallmatrix}
I_m \left(\gamma +\sigma_{W_2}^2 \left(\eta '-\zeta \right)\right) & 0 & 0 & \frac{\zeta  X^\top \sigma_{W_2}^2}{n_0} & \frac{\sqrt{\eta -\zeta } \Theta_F^\top}{n_1} & \frac{\sqrt{\zeta } X^\top}{\sqrt{n_0} n_1} & -\frac{\zeta ^2 X^\top \sigma_{W_2}^4}{n_0^2} & 0 & \frac{\sqrt{\eta -\zeta } \Theta_F^\top \beta}{n_0 n_1^2} & 0 & 0 \\
 -\sqrt{\eta -\zeta } \Theta_F & I_{n_1} & -\frac{\sqrt{\zeta } W_1}{\sqrt{n_0}} & 0 & 0 & 0 & 0 & 0 & 0 & 0 & 0 \\
 -X & 0 & I_{n_0} & 0 & 0 & 0 & 0 & 0 & 0 & 0 & 0 \\
 -X & 0 & 0 & I_{n_0} & 0 & 0 & 0 & 0 & \frac{\sqrt{\zeta } W_1^\top}{\sqrt{n_0} n_1} & 0 & 0 \\
 -\sqrt{\eta -\zeta } \Theta_F & 0 & 0 & -\frac{\sqrt{\zeta } W_1}{\sqrt{n_0}} & I_{n_1} & 0 & 0 & -\frac{\zeta  \sqrt{\eta -\zeta } \Theta_F \sigma_{W_2}^2}{n_0} & 0 & 0 & 0 \\
 0 & 0 & 0 & 0 & -W_1^\top & I_{n_0} & 0 & 0 & W_1^\top \left(\frac{\eta -\zeta }{n_1}+\frac{\zeta  \sigma_{W_2}^2}{n_0}\right) & 0 & 0 \\
 0 & 0 & 0 & 0 & 0 & 0 & I_{n_0} & -X & 0 & 0 & 0 \\
 0 & 0 & 0 & 0 & 0 & 0 & \frac{\zeta  X^\top \sigma_{W_2}^2}{n_0} & I_m \left(\gamma +\sigma_{W_2}^2 \left(\eta '-\zeta \right)\right) & 0 & \frac{\sqrt{\eta -\zeta } \Theta_F^\top}{n_1} & \frac{\sqrt{\zeta } X^\top}{\sqrt{n_0} n_1} \\
 0 & 0 & 0 & 0 & 0 & 0 & -\frac{\sqrt{\zeta } W_1}{\sqrt{n_0}} & -\sqrt{\eta -\zeta } \Theta_F & I_{n_1} & 0 & 0 \\
 0 & 0 & 0 & 0 & 0 & 0 & -\frac{\sqrt{\zeta } W_1}{\sqrt{n_0}} & -\sqrt{\eta -\zeta } \Theta_F & 0 & I_{n_1} & 0 \\
 0 & 0 & 0 & 0 & 0 & 0 & 0 & 0 & 0 & -W_1^\top & I_{n_0} 
\end{smallmatrix}
\right)\,.
}
\subsection{Operator-valued Stieltjes transform}
Even though the individual error terms $E_{21}, E_{22}, E_{31}, E_{32}, E_{33}$ can be written as the trace of self-adjoint matrices, the individual $Q$ matrices are not themselves self-adjoint. However, by enlarging the dimensionality by a factor of two, equivalent self-adjoint representations can easily be constructed. To do so, we simply utilize the identity,
\eq{
U^T Q V = \bar{U}^\top \bar{Q} \bar{V} \equiv
\begin{pmatrix}
\frac{1}{2} U^\top & V^\top
\end{pmatrix}
\begin{pmatrix}
0 & Q^\top \\
Q & 0
\end{pmatrix}
\begin{pmatrix}
\frac{1}{2} U\\
V
\end{pmatrix}\,.
}
Observe that $\bar{Q}_{21},\bar{Q}_{22},\bar{Q}_{31}, \bar{Q}_{32}$ and $\bar{Q}_{33}$ are all self-adjoint block matrices whose blocks are either constants or proportional to one of $\{X,X^\top,W_1,W_1^\top,\Theta_F,\Theta_F^\top\}$; let us denote the constant terms as $Z$. As such, we can directly utilize the results of~\cite{far2006spectra,mingo2017free} to compute the error terms in question.

For each linear pencil, the corresponding error term can be extracted from the operator-valued Stieltjes transform $G:M_d(\mathbb{C})^+\to M_d(\mathbb{C})^+$, which is a solution of the equation,
\eq{\label{eqn_Geqn}
ZG = I_d + \eta(G) G\,,
}
where $d$ is the number of blocks, $\eta: M_d(\mathbb{C})\to M_d(\mathbb{C})$ defined by
\eq{\label{eqn_Deqn2}
[\eta(D)]_{ij} = \sum_{kl} \sigma(i,k;l,j) \alpha_k D_{kl} \,,
}
where $\alpha_k$ is dimensionality of the $k$th block and $\sigma(i,k;l,k)$ denotes the covariance between the entries of the $ij$ block of $\bar{Q}$ and entries of the $kl$ block of $\bar{Q}$. Eqn.~\eqref{eqn_Geqn} may admit many solutions, but there is a unique solution such that $\text{Im}G \succ 0$ for $\text{Im} Z\succ 0$.

The constants $Z$, the entries of $\sigma$, and therefore the equations~\eqref{eqn_Deqn2} are manifest by inspection of the block matrix representations for $Q$. Although the matrix representations are too large to reproduce here, we can nevertheless extract the equations satisfied by each entry of $G$, which we present in the subsequent sections.

\subsubsection{$E_{21}$}
The equations satisfied by the operator-valued Stieltjes transform $G$ of $\bar{Q}  _{21}$ induce the following structure on $G$,
\eq{
G = \begin{pmatrix} 0 & G_{12} \\ G_{12}^\top & 0 \end{pmatrix}\,,
}
where,
\eq{
G_{12} = \left(
\begin{array}{cccccccccccccc}
 g_8 & 0 & 0 & 0 & 0 & 0 & 0 & 0 & 0 & 0 & 0 & 0 & 0 & 0 \\
 0 & g_9 & 0 & g_6 & 0 & 0 & 0 & 0 & 0 & 0 & 0 & 0 & 0 & 0 \\
 0 & 0 & g_{11} & 0 & 0 & 0 & 0 & 0 & 0 & 0 & 0 & 0 & 0 & 0 \\
 0 & g_{12} & 0 & g_{10} & 0 & 0 & 0 & 0 & 0 & 0 & 0 & 0 & 0 & 0 \\
 0 & 0 & 0 & 0 & g_8 & 0 & 0 & 0 & 0 & 0 & 0 & 0 & 0 & 0 \\
 0 & 0 & 0 & 0 & 0 & g_1 & 0 & g_5 & 0 & g_4 & 0 & g_7 & 0 & 0 \\
 0 & 0 & 0 & 0 & 0 & 0 & g_9 & 0 & g_6 & 0 & 0 & 0 & 0 & 0 \\
 0 & 0 & 0 & 0 & 0 & 0 & 0 & g_{11} & 0 & g_3 & 0 & 0 & 0 & 0 \\
 0 & 0 & 0 & 0 & 0 & 0 & g_{12} & 0 & g_{10} & 0 & 0 & 0 & 0 & 0 \\
 0 & 0 & 0 & 0 & 0 & 0 & 0 & 0 & 0 & g_1 & 0 & 0 & 0 & 0 \\
 0 & 0 & 0 & 0 & 0 & 0 & 0 & 0 & 0 & 0 & g_8 & 0 & 0 & 0 \\
 0 & 0 & 0 & 0 & 0 & 0 & 0 & 0 & 0 & g_2 & 0 & g_{11} & 0 & 0 \\
 0 & 0 & 0 & 0 & 0 & 0 & 0 & 0 & 0 & 0 & 0 & 0 & g_9 & g_6 \\
 0 & 0 & 0 & 0 & 0 & 0 & 0 & 0 & 0 & 0 & 0 & 0 & g_{12} & g_{10} \\
\end{array}
\right)\,,
}
and the independent entry-wise component functions $g_i$ combine to produce the error $E_{21}$ through the relation,
\eq{
E_{21} = \frac{g_4 (\eta -\zeta )}{n_0}+\frac{2\sqrt{\zeta } g_6 \sqrt{n_0} \sigma _{W_2}^2}{\psi }-\frac{g_2}{n_0}\,,
}
and themselves satisfy the following system of polynomial equations,
{\small
\begin{subequations}
\begin{align}
0 &= 1-g_1\\
0 &= \sqrt{\zeta } g_9 g_{11} \sqrt{n_0}-g_{12} \psi\\
0 &= \sqrt{\zeta } g_6 g_{11} \sqrt{n_0}-g_{10} \psi +\psi\\
0 &= g_7 (\eta -\zeta )+\sqrt{\zeta } g_6 g_{11} \sqrt{n_0}\\
0 &= g_8 g_{11} n_0 \sqrt{\eta -\zeta }-g_3 \phi  \big(\gamma +\sigma _{W_2}^2 \big(\eta '-\zeta \big)\big)\\
0 &= -\sqrt{\zeta } g_8 g_9 \psi -g_6 \sqrt{n_0} \phi  \big(\gamma +\sigma _{W_2}^2 \big(\eta '-\zeta \big)\big)\\
0 &= -\sqrt{\zeta } g_8 g_{12} \psi -\big(g_{10}-1\big) \sqrt{n_0} \phi  \big(\gamma +\sigma _{W_2}^2 \big(\eta '-\zeta \big)\big)\\
0 &= g_6 \sqrt{n_0} \phi  \big(\gamma +\sigma _{W_2}^2 \big(\eta '-\zeta \big)\big)+g_8 \big(\sqrt{\zeta } g_{10} \psi +\zeta  g_6 \sqrt{n_0} \sigma _{W_2}^2\big)\\
0 &= g_8 g_{11} \psi  (\eta -\zeta )-\phi  \big(g_5 \sqrt{\eta -\zeta }-\sqrt{\zeta } g_6 g_{11} \sqrt{n_0}\big) \big(\sigma _{W_2}^2 \big(\zeta -\eta '\big)-\gamma \big)\\
0 &= \big(g_9-1\big) \sqrt{n_0} \phi  \big(\gamma +\sigma _{W_2}^2 \big(\eta '-\zeta \big)\big)+g_8 \big(\sqrt{\zeta } g_{12} \psi +\zeta  g_9 \sqrt{n_0} \sigma _{W_2}^2\big)\\
0 &= g_1 g_8 n_0 \sqrt{\eta -\zeta }+g_3 \big(g_8 \psi  (\zeta -\eta )+\phi  \big(\sqrt{\zeta } g_6 \sqrt{n_0}-1\big) \big(\gamma +\sigma _{W_2}^2 \big(\eta '-\zeta \big)\big)\big)\\
0 &= \sqrt{\zeta } g_{10} g_{11} \sqrt{n_0} \phi  \big(\sigma _{W_2}^2 \big(\zeta -\eta '\big)-\gamma \big)+g_{12} \psi  \big(\gamma  \phi +\sigma _{W_2}^2 \big(-\zeta  \phi +\phi  \eta '+\zeta  g_8\big)\big)\\
0 &= g_{11} \big(g_8 \psi  (\zeta -\eta )+\phi  \big(\sqrt{\zeta } g_6 \sqrt{n_0}-1\big) \big(\gamma +\sigma _{W_2}^2 \big(\eta '-\zeta \big)\big)\big)+\phi  \big(\gamma +\sigma _{W_2}^2 \big(\eta '-\zeta \big)\big)\\
0 &= g_{11} n_0 \big(g_8 \psi  (\eta -\zeta )+\sqrt{\zeta } g_6 \sqrt{n_0} \phi  \big(\sigma _{W_2}^2 \big(\zeta -\eta '\big)-\gamma \big)\big)-g_2 \psi  \phi  \big(\gamma +\sigma _{W_2}^2 \big(\eta '-\zeta \big)\big)\\
0 &= g_9 \psi  \big(\gamma  \phi +\sigma _{W_2}^2 \big(\phi  \big(\eta '-\zeta \big)+\zeta  g_8\big)\big)-\phi  \big(\sqrt{\zeta } g_6 g_{11} \sqrt{n_0}+\psi \big) \big(\gamma +\sigma _{W_2}^2 \big(\eta '-\zeta \big)\big)\\
0 &= g_8 \big(-\sqrt{\zeta } g_{12} \psi -\sqrt{n_0} \big(\gamma +g_{11} (\eta -\zeta )+\sigma _{W_2}^2 \big(\eta '+\zeta  \big(g_9-1\big)\big)\big)\big)+\sqrt{n_0} \big(\gamma +\sigma _{W_2}^2 \big(\eta '-\zeta \big)\big)\\
0 &= \sqrt{\zeta } g_1 g_6 \sqrt{n_0} \phi  \big(\sigma _{W_2}^2 \big(\zeta -\eta '\big)-\gamma \big)-g_7 (\zeta -\eta ) \big(g_8 \psi  (\zeta -\eta )+\phi  \big(\sqrt{\zeta } g_6 \sqrt{n_0}-1\big) \big(\gamma +\sigma _{W_2}^2 \big(\eta '-\zeta \big)\big)\big)\\
0 &= g_1 n_0 \big(g_8 \psi  (\eta -\zeta )+\sqrt{\zeta } g_6 \sqrt{n_0} \phi  \big(\sigma _{W_2}^2 \big(\zeta -\eta '\big)-\gamma \big)\big)+g_2 \psi  \big(g_8 \psi  (\zeta -\eta )\nonumber\\
&\quad\;+\phi  \big(\sqrt{\zeta } g_6 \sqrt{n_0}-1\big) \big(\gamma +\sigma _{W_2}^2 \big(\eta '-\zeta \big)\big)\big)\\
0 &= g_1 \big(g_8 \psi  (\eta -\zeta )+\sqrt{\zeta } g_6 \sqrt{n_0} \phi  \big(\sigma _{W_2}^2 \big(\zeta -\eta '\big)-\gamma \big)\big)+g_5 \sqrt{\eta -\zeta } \big(g_8 \psi  (\eta -\zeta )\nonumber\\
&\quad\;-\phi  \big(\sqrt{\zeta } g_6 \sqrt{n_0}-1\big) \big(\gamma +\sigma _{W_2}^2 \big(\eta '-\zeta \big)\big)\big)\\
0 &= n_0 \big(-\zeta  g_5 g_8 \psi  \sqrt{\eta -\zeta }+\eta  g_5 g_8 \psi  \sqrt{\eta -\zeta }+g_8 \psi  (\zeta -\eta ) \big(g_7 (\zeta -\eta )-g_1\big)\nonumber\\
&\quad\; +\sqrt{\zeta } g_6 \sqrt{n_0} \phi  \big(g_7 (\zeta -\eta )+g_1\big) \big(\gamma +\sigma _{W_2}^2 \big(\eta '-\zeta \big)\big)\big)+g_4 \psi  \phi  (\zeta -\eta ) \big(\gamma +\sigma _{W_2}^2 \big(\eta '-\zeta \big)\big)\\
0 &= \sqrt{n_0} \sqrt{\eta -\zeta } \big(g_1 g_8 \sqrt{n_0} \psi  (\eta -\zeta )+\sqrt{\zeta } g_1 g_6 n_0 \phi  \big(\gamma +\sigma _{W_2}^2 \big(\eta '-\zeta \big)\big)-\sqrt{\zeta } g_2 g_6 \psi  \phi  \big(\gamma +\sigma _{W_2}^2 \big(\eta '-\zeta \big)\big)\big)\nonumber\\
&\quad\; +g_3 \psi  (\zeta -\eta ) \big(g_8 \psi  (\eta -\zeta )+\sqrt{\zeta } g_6 \sqrt{n_0} \phi  \big(\sigma _{W_2}^2 \big(\zeta -\eta '\big)-\gamma \big)\big)+g_4 \psi  (-\phi ) (\eta -\zeta )^{3/2} \big(\gamma +\sigma _{W_2}^2 \big(\eta '-\zeta \big)\big)\,.
\end{align}
\end{subequations}
}
After some straightforward algebra, one can eliminate all $g_i$ except for $g_6$ and $g_{8}$, which satisfy coupled polynomial equations. Those equations can be shown to be identical to eqn.~(\ref{eqn:tau}) by invoking the change of variables,
\begin{equation}
    g_6 = - \frac{\sqrt{\zeta} \psi}{\sqrt{n_0} \phi} \tau_2\,,\quad\text{and}\quad
    g_8 = \big(\gamma +\sigma _{W_2}^2 \big(\eta '-\zeta \big)\big)\tau_1\,.
\end{equation}
In terms of these variables, the error $E_{21}$ is given by,
\begin{equation}
    E_{21} = 2(\tau_2/\tau_1 - 1)\,.
\end{equation}
\subsubsection{$E_{22}$}
The equations satisfied by the operator-valued Stieltjes transform $G$ of $\bar{Q}_{22}$ induce the following structure on $G$,
\eq{
G = \begin{pmatrix} 0 & G_{12} \\ G_{12}^\top & 0 \end{pmatrix}\,,
}
where,
\eq{
G_{12} = \left(
\begin{array}{ccccccc}
 g_{11} & 0 & 0 & 0 & 0 & 0 & g_7 \\
 0 & g_5 & 0 & g_2 & 0 & g_9 & 0 \\
 0 & 0 & g_{10} & 0 & 0 & 0 & 0 \\
 0 & g_3 & 0 & g_4 & 0 & g_8 & 0 \\
 g_{14} & 0 & 0 & 0 & g_1 & 0 & g_6 \\
 0 & 0 & 0 & 0 & 0 & g_{13} & 0 \\
 g_{14} & 0 & 0 & 0 & 0 & 0 & g_{12} \\
\end{array}
\right)\,,
}
and the independent entry-wise component functions $g_i$ combine to produce the error $E_{22}$ through the relation,
\eq{
E_{22} = \frac{2\sqrt{\zeta } g_9 \sigma _{W_2}^2 \big(\psi  (\eta -\zeta )+\zeta  \sigma _{W_2}^2\big)}{\sqrt{n_0} \psi }+2g_8 (\eta -\zeta ) \sigma _{W_2}^2\,,
}
and themselves satisfy the following system of polynomial equations,
{\small
\begin{subequations}
\begin{align}
0 &= \sqrt{\zeta } g_{11} g_{13} \sqrt{n_0}-g_{14} \psi\\
0 &= \sqrt{\zeta } g_7 g_{13} \sqrt{n_0}-g_{12} \psi +\psi\\
0 &= g_1 \psi  \big(g_3 \sqrt{n_0}-\sqrt{\zeta } g_4\big)-g_3 \sqrt{n_0} \sigma _{W_2}^2\\ 
0 &= -g_1 \psi  \big(\sqrt{\zeta } g_5+g_3 \sqrt{n_0}\big)-g_3 \sqrt{n_0} \sigma _{W_2}^2\\
0 &= g_1 \psi  \big(g_5 \sqrt{n_0}-\sqrt{\zeta } g_2\big)-\sqrt{\zeta } g_2 \sigma _{W_2}^2\\
0 &= g_1 \psi  \big(\sqrt{\zeta } g_2+g_4 \sqrt{n_0}\big)-\sqrt{\zeta } g_2 \sigma _{W_2}^2\\
0 &= g_1 \psi  \big(g_5 \sqrt{n_0}-\sqrt{\zeta } g_2\big)-\big(g_5-1\big) \sqrt{n_0} \sigma _{W_2}^2\\
0 &= -g_1 \psi  \big(\sqrt{\zeta } g_2+g_4 \sqrt{n_0}\big)-\big(g_4-1\big) \sqrt{n_0} \sigma _{W_2}^2\\
0 &= g_1 \psi  \big(g_3 \sqrt{n_0}-\sqrt{\zeta } g_4\big)-\sqrt{\zeta } \big(g_4-1\big) \sigma _{W_2}^2\\
0 &= g_1 \psi  \big(\sqrt{\zeta } g_5+g_3 \sqrt{n_0}\big)-\sqrt{\zeta } \big(g_5-1\big) \sigma _{W_2}^2\\
0 &= -\sqrt{\zeta } g_{10} g_{11} \psi -g_7 \sqrt{n_0} \phi  \big(\gamma +\sigma _{W_2}^2 \big(\eta '-\zeta \big)\big)\\
0 &= -\sqrt{\zeta } g_{10} g_{14} \psi -g_6 \sqrt{n_0} \phi  \big(\gamma +\sigma _{W_2}^2 \big(\eta '-\zeta \big)\big)\\
0 &= -\sqrt{\zeta } g_{10} g_{14} \psi -\big(g_{12}-1\big) \sqrt{n_0} \phi  \big(\gamma +\sigma _{W_2}^2 \big(\eta '-\zeta \big)\big)\\
0 &= g_1 \big(-\zeta  g_2+\sqrt{\zeta } \big(g_5-g_4\big) \sqrt{n_0}+g_3 n_0\big)-\sqrt{\zeta } \big(g_1-1\big) \sqrt{n_0} \sigma _{W_2}^2\\
0 &= g_1 \psi  \big(\sqrt{\zeta } g_9+g_8 \sqrt{n_0}\big)+\sqrt{\zeta } \big(g_7 g_{13} n_0-g_9\big) \sigma _{W_2}^2+g_6 g_{13} \sqrt{n_0} \psi\\
0 &= g_7 \sqrt{n_0} \phi  \big(\gamma +\sigma _{W_2}^2 \big(\eta '-\zeta \big)\big)+g_{10} \big(\sqrt{\zeta } g_{12} \psi +\zeta  g_7 \sqrt{n_0} \sigma _{W_2}^2\big)\\
0 &= \big(g_{11}-1\big) \sqrt{n_0} \phi  \big(\gamma +\sigma _{W_2}^2 \big(\eta '-\zeta \big)\big)+g_{10} \big(\sqrt{\zeta } g_{14} \psi +\zeta  g_{11} \sqrt{n_0} \sigma _{W_2}^2\big)\\
0 &= \sqrt{\zeta } g_{12} g_{13} \sqrt{n_0} \phi  \big(\sigma _{W_2}^2 \big(\zeta -\eta '\big)-\gamma \big)+g_{14} \psi  \big(\gamma  \phi +\sigma _{W_2}^2 \big(-\zeta  \phi +\phi  \eta '+\zeta  g_{10}\big)\big)\\
0 &= g_{13} \big(g_{10} \psi  (\zeta -\eta )+\phi  \big(\sqrt{\zeta } g_7 \sqrt{n_0}-1\big) \big(\gamma +\sigma _{W_2}^2 \big(\eta '-\zeta \big)\big)\big)+\phi  \big(\gamma +\sigma _{W_2}^2 \big(\eta '-\zeta \big)\big)\\
0 &= g_6 \psi  \big(-\zeta  g_2+\sqrt{\zeta } \big(g_5-g_4\big) \sqrt{n_0}+g_3 n_0\big)+\sqrt{\zeta } \sqrt{n_0} \sigma _{W_2}^2 \big(g_7 \big(\zeta  g_9+\sqrt{\zeta } \big(g_5+g_8\big) \sqrt{n_0}+g_3 n_0\big)-g_6 \psi \big)\\
0 &= g_{11} \psi  \big(\gamma  \phi +\sigma _{W_2}^2 \big(\phi  \big(\eta '-\zeta \big)+\zeta  g_{10}\big)\big)-\phi  \big(\sqrt{\zeta } g_7 g_{13} \sqrt{n_0}+\psi \big) \big(\gamma +\sigma _{W_2}^2 \big(\eta '-\zeta \big)\big)\\
0 &= g_{10} \big(-\sqrt{\zeta } g_{14} \psi -\sqrt{n_0} \big(\gamma +g_{13} (\eta -\zeta )+\sigma _{W_2}^2 \big(\eta '+\zeta  \big(g_{11}-1\big)\big)\big)\big)+\sqrt{n_0} \big(\gamma +\sigma _{W_2}^2 \big(\eta '-\zeta \big)\big)\\
0 &= g_{14} \psi  \big(-\zeta  g_2+\sqrt{\zeta } \big(g_5-g_4\big) \sqrt{n_0}+g_3 n_0\big)+\sqrt{\zeta } \sqrt{n_0} \sigma _{W_2}^2 \big(g_{11} \big(\zeta  g_9\nonumber\\
&\quad\quad+\sqrt{\zeta } \big(g_5+g_8\big) \sqrt{n_0}+g_3 n_0\big)-g_{14} \psi \big)\\
0 &= \sqrt{\zeta } g_6 g_{13} \sqrt{n_0} \phi  \big(\sigma _{W_2}^2 \big(\zeta -\eta '\big)-\gamma \big)-g_1 \phi  \big(\zeta  g_9+\sqrt{\zeta } \big(g_5+g_8\big) \sqrt{n_0}+g_3 n_0\big) \big(\gamma +\sigma _{W_2}^2 \big(\eta '-\zeta \big)\big)\nonumber\\
&\quad\quad+g_{14} \psi  \big(\gamma  \phi +\sigma _{W_2}^2 \big(-\zeta  \phi +\phi  \eta '+\zeta  g_{10}\big)\big)\\
0 &= g_1 \psi  \phi  \big(\sqrt{\zeta } g_9+g_8 \sqrt{n_0}\big) \big(\gamma +\sigma _{W_2}^2 \big(\eta '-\zeta \big)\big)+\sqrt{n_0} \big(\sigma _{W_2}^2 \big(g_{10} g_{13} \psi  (\eta -\zeta )+g_8 \phi  \big(\gamma +\sigma _{W_2}^2 \big(\eta '-\zeta \big)\big)\big)\nonumber\\
&\quad\quad+g_6 g_{13} \psi  \phi  \big(\gamma +\sigma _{W_2}^2 \big(\eta '-\zeta \big)\big)\big)\\
0 &= \sqrt{\zeta } g_8 \sigma _{W_2}^2 \big(g_{10} \psi  (\eta -\zeta )-\phi  \big(\sqrt{\zeta } g_7 \sqrt{n_0}-1\big) \big(\gamma +\sigma _{W_2}^2 \big(\eta '-\zeta \big)\big)\big)-g_3 \sqrt{n_0} \phi  \big(\sqrt{\zeta } g_7 \sqrt{n_0} \sigma _{W_2}^2+g_6 \psi \big)\nonumber\\
&\quad\quad \big(\gamma +\sigma _{W_2}^2 \big(\eta '-\zeta \big)\big)+\sqrt{\zeta } g_4 \psi  \big(g_6 \phi  \big(\gamma +\sigma _{W_2}^2 \big(\eta '-\zeta \big)\big)+g_{10} (\eta -\zeta ) \sigma _{W_2}^2\big)\\
0 &= \sqrt{\zeta } g_9 \sigma _{W_2}^2 \big(g_{10} \psi  (\eta -\zeta )-\phi  \big(\sqrt{\zeta } g_7 \sqrt{n_0}-1\big)\big(\gamma +\sigma _{W_2}^2 \big(\eta '-\zeta \big)\big)\big)-g_5 \sqrt{n_0} \phi  \big(\sqrt{\zeta } g_7 \sqrt{n_0} \sigma _{W_2}^2 + g_6 \psi \big)\nonumber\\
&\quad\quad \big(\gamma +\sigma _{W_2}^2 \big(\eta '-\zeta \big)\big)+\sqrt{\zeta } g_2 \psi  \big(g_6 \phi  \big(\gamma +\sigma _{W_2}^2 \big(\eta '-\zeta \big)\big)+g_{10} (\eta -\zeta ) \sigma _{W_2}^2\big)
\end{align}
\end{subequations}
}

After some straightforward algebra, one can eliminate all $g_i$ except for $g_7$ and $g_{10}$, which satisfy coupled polynomial equations. Those equations can be shown to be identical to eqn.~(\ref{eqn:tau}) by invoking the change of variables,
\begin{equation}
    g_7 = - \frac{\sqrt{\zeta} \psi}{\sqrt{n_0} \phi} \tau_2\,,\quad\text{and}\quad
    g_{10} = \big(\gamma +\sigma _{W_2}^2 \big(\eta '-\zeta \big)\big)\tau_1\,.
\end{equation}
The error $E_{22}$ is then given by,
\begin{equation}
    E_{22} = 2 \zeta  \big(\frac{\tau _2}{\tau _1}-1\big)+\frac{2\psi  \big(\zeta  \big(\tau _2-\tau _1\big)+\eta  \tau _1\big){}^2 \big(\big(\tau _2-\tau _1\big) \phi +\zeta  \tau _1 \tau _2 \sigma _{W_2}^2\big)}{\zeta  \tau _1^2 \tau _2 \phi }\,.
\end{equation}
\subsubsection{$E_{31}$}
The equations satisfied by the operator-valued Stieltjes transform $G$ of $\bar{Q}_{31}$ induce the following structure on $G$,
\eq{
G = \begin{pmatrix} 0 & G_{12} \\ G_{12}^\top & 0 \end{pmatrix}\,,
}
where,
\eq{
G_{12} =\left(
\begin{array}{cccccccc}
 g_5 & 0 & 0 & 0 & 0 & g_2 & 0 & 0 \\
 0 & g_6 & 0 & g_1 & g_3 & 0 & 0 & g_4 \\
 0 & 0 & g_8 & 0 & 0 & 0 & g_{12} & 0 \\
 0 & g_{11} & 0 & g_7 & g_{10} & 0 & 0 & g_9 \\
 0 & 0 & 0 & 0 & g_6 & 0 & 0 & g_1 \\
 0 & 0 & 0 & 0 & 0 & g_5 & 0 & 0 \\
 0 & 0 & 0 & 0 & 0 & 0 & g_8 & 0 \\
 0 & 0 & 0 & 0 & g_{11} & 0 & 0 & g_7 \\
\end{array}
\right)\,,
}
and the independent entry-wise component functions $g_i$ give the error $E_{31}$ through the relation,
\eq{
E_{31} = \frac{g_2 n_0 \sigma _{\varepsilon }^2}{\phi  \big(\gamma +\sigma _{W_2}^2 \big(\eta '-\zeta \big)\big)}\,,
}
and themselves satisfy the following system of polynomial equations,
{\small
\begin{subequations}
\begin{align}
0 &= \sqrt{\zeta } g_6 g_8 \sqrt{n_0}-g_{11} \psi\\
0 &= \sqrt{\zeta } g_1 g_8 \sqrt{n_0}-g_7 \psi +\psi\\
0 &= -\sqrt{\zeta } g_5 g_6 \psi -g_1 \sqrt{n_0} \phi  \big(\gamma +\sigma _{W_2}^2 \big(\eta '-\zeta \big)\big)\\
0 &= -\sqrt{\zeta } g_5 g_{11} \psi -\big(g_7-1\big) \sqrt{n_0} \phi  \big(\gamma +\sigma _{W_2}^2 \big(\eta '-\zeta \big)\big)\\
0 &= -\zeta  g_7 g_8 \psi +\sqrt{\zeta } \sqrt{n_0} \big(\big(g_4 g_8+g_1 g_{12}\big) n_0-\zeta  g_1 g_8 \sigma _{W_2}^2\big)-g_9 n_0 \psi\\
0 &= g_1 \sqrt{n_0} \phi  \big(\gamma +\sigma _{W_2}^2 \big(\eta '-\zeta \big)\big)+g_5 \big(\sqrt{\zeta } g_7 \psi +\zeta  g_1 \sqrt{n_0} \sigma _{W_2}^2\big)\\
0 &= \sqrt{\zeta } g_6 g_{12} n_0^{3/2}-g_8 \big(\zeta  g_{11} \psi +\sqrt{\zeta } \sqrt{n_0} \big(\zeta  g_6 \sigma _{W_2}^2-g_3 n_0\big)\big)-g_{10} n_0 \psi\\
0 &= \big(g_6-1\big) \sqrt{n_0} \phi  \big(\gamma +\sigma _{W_2}^2 \big(\eta '-\zeta \big)\big)+g_5 \big(\sqrt{\zeta } g_{11} \psi +\zeta  g_6 \sqrt{n_0} \sigma _{W_2}^2\big)\\
0 &= \sqrt{\zeta } g_7 g_8 \sqrt{n_0} \phi  \big(\sigma _{W_2}^2 \big(\zeta -\eta '\big)-\gamma \big)+g_{11} \psi  \big(\gamma  \phi +\sigma _{W_2}^2 \big(-\zeta  \phi +\phi  \eta '+\zeta  g_5\big)\big)\\
0 &= g_8 \big(g_5 \psi  (\zeta -\eta )+\phi  \big(\sqrt{\zeta } g_1 \sqrt{n_0}-1\big) \big(\gamma +\sigma _{W_2}^2 \big(\eta '-\zeta \big)\big)\big)+\phi  \big(\gamma +\sigma _{W_2}^2 \big(\eta '-\zeta \big)\big)\\
0 &= g_6 \psi  \big(\gamma  \phi +\sigma _{W_2}^2 \big(-\zeta  \phi +\phi  \eta '+\zeta  g_5\big)\big)-\phi  \big(\sqrt{\zeta } g_1 g_8 \sqrt{n_0}+\psi \big) \big(\gamma +\sigma _{W_2}^2 \big(\eta '-\zeta \big)\big)\\
0 &= g_5 \big(\sqrt{\zeta } g_{11} \psi +\sqrt{n_0} \big(\gamma +g_8 (\eta -\zeta )+\sigma _{W_2}^2 \big(\eta '+\zeta  \big(g_6-1\big)\big)\big)\big)-\sqrt{n_0} \big(\gamma +\sigma _{W_2}^2 \big(\eta '-\zeta \big)\big)\\
0 &= \sqrt{\zeta } g_5 \psi  \big(g_6 \big(\psi  (\eta -\zeta )+\zeta  \sigma _{W_2}^2\big)-g_3 n_0\big)-\sqrt{n_0} \big(\sqrt{\zeta } g_2 g_6 \sqrt{n_0} \psi +g_4 n_0 \phi  \big(\gamma +\sigma _{W_2}^2 \big(\eta '-\zeta \big)\big)\nonumber\\
&\quad\quad+\zeta  g_1 g_8 \phi  \big(\gamma +\sigma _{W_2}^2 \big(\eta '-\zeta \big)\big)\big)\\
0 &= \sqrt{\zeta } g_5 \psi  \big(g_{11} \big(\psi  (\eta -\zeta )+\zeta  \sigma _{W_2}^2\big)-g_{10} n_0\big)-\sqrt{n_0} \big(\sqrt{\zeta } g_2 g_{11} \sqrt{n_0} \psi +g_9 n_0 \phi  \big(\gamma +\sigma _{W_2}^2 \big(\eta '-\zeta \big)\big)\nonumber\\
&\quad\quad+\zeta  g_7 g_8 \phi  \big(\gamma +\sigma _{W_2}^2 \big(\eta '-\zeta \big)\big)\big)\\
0 &= g_5 \big(-\sqrt{\zeta } g_9 n_0 \psi +\zeta  \sqrt{n_0} \sigma _{W_2}^2 \big(\zeta  g_1 \sigma _{W_2}^2-g_4 n_0\big)+\sqrt{\zeta } g_7 \psi  \big(\psi  (\eta -\zeta )+\zeta  \sigma _{W_2}^2\big)\big)\nonumber\\
&\quad\quad-n_0 \big(g_4 \sqrt{n_0} \phi  \big(\gamma +\sigma _{W_2}^2 \big(\eta '-\zeta \big)\big)+g_2 \big(\sqrt{\zeta } g_7 \psi +\zeta  g_1 \sqrt{n_0} \sigma _{W_2}^2\big)\big)\\
0 &= g_5 \big(-\sqrt{\zeta } g_{10} n_0 \psi +\zeta  \sqrt{n_0} \sigma _{W_2}^2 \big(\zeta  g_6 \sigma _{W_2}^2-g_3 n_0\big)+\sqrt{\zeta } g_{11} \psi  \big(\psi  (\eta -\zeta )+\zeta  \sigma _{W_2}^2\big)\big)\nonumber\\
&\quad\quad-n_0 \big(g_3 \sqrt{n_0} \phi  \big(\gamma +\sigma _{W_2}^2 \big(\eta '-\zeta \big)\big)+g_2 \big(\sqrt{\zeta } g_{11} \psi +\zeta  g_6 \sqrt{n_0} \sigma _{W_2}^2\big)\big)\\
0 &= g_2 g_8 n_0 \psi  (\eta -\zeta )-g_5 \psi  (\zeta -\eta ) \big(g_8 \psi  (\zeta -\eta )+g_{12} n_0\big)-\sqrt{n_0} \phi  \big(g_{12} \big(\sqrt{\zeta } g_1 n_0-\sqrt{n_0}\big)\nonumber\\
&\quad\quad+\sqrt{\zeta } g_8 \big(g_4 n_0-\zeta  g_1 \sigma _{W_2}^2\big)\big) \big(\gamma +\sigma _{W_2}^2 \big(\eta '-\zeta \big)\big)+\zeta  g_7 g_8 \psi  \phi  \big(\gamma +\sigma _{W_2}^2 \big(\eta '-\zeta \big)\big)\\
0 &= g_2 n_0 \big(-\sqrt{\zeta } g_{11} \psi -\sqrt{n_0} \big(\gamma +g_8 (\eta -\zeta )+\sigma _{W_2}^2 \big(\eta '+\zeta  \big(g_6-1\big)\big)\big)\big)+g_5 \big(\sqrt{n_0} \big(g_8 \psi  (\zeta -\eta )^2\nonumber\\
&\quad\quad+g_{12} n_0 (\zeta -\eta )-\sqrt{\zeta } g_{10} \sqrt{n_0} \psi -\zeta  g_3 n_0 \sigma _{W_2}^2+\zeta ^2 g_6 \sigma _{W_2}^4\big)+\sqrt{\zeta } g_{11} \psi  \big(\psi  (\eta -\zeta )+\zeta  \sigma _{W_2}^2\big)\big)\\
0 &= g_3 n_0 \psi  \big(\gamma  \phi +\sigma _{W_2}^2 \big(\phi  \big(\eta '-\zeta \big)+\zeta  g_5\big)\big)-\sqrt{\zeta } \big(g_4 g_8 n_0^{3/2} \phi  \big(\gamma +\sigma _{W_2}^2 \big(\eta '-\zeta \big)\big)\nonumber\\
&\quad\quad+g_1 \sqrt{n_0} \phi  \big(g_{12} n_0-\zeta  g_8 \sigma _{W_2}^2\big) \big(\gamma +\sigma _{W_2}^2 \big(\eta '-\zeta \big)\big)+\sqrt{\zeta } g_6 \psi  \sigma _{W_2}^2 \big(\zeta  g_5 \sigma _{W_2}^2-g_2 n_0\big)\big)\\
0 &= g_{10} n_0 \psi  \big(\gamma  \phi +\sigma _{W_2}^2 \big(\phi  \big(\eta '-\zeta \big)+\zeta  g_5\big)\big)-\sqrt{\zeta } \big(g_7 g_{12} n_0^{3/2} \phi  \big(\gamma +\sigma _{W_2}^2 \big(\eta '-\zeta \big)\big)\nonumber\\
&\quad\quad+g_8 \sqrt{n_0} \phi  \big(g_9 n_0-\zeta  g_7 \sigma _{W_2}^2\big) \big(\gamma +\sigma _{W_2}^2 \big(\eta '-\zeta \big)\big)+\sqrt{\zeta } g_{11} \psi  \sigma _{W_2}^2 \big(\zeta  g_5 \sigma _{W_2}^2-g_2 n_0\big)\big)
\end{align}
\end{subequations}
}
After some straightforward algebra, one can eliminate all $g_i$ except for $g_1$ and $g_{5}$, which satisfy coupled polynomial equations. Those equations can be shown to be identical to eqn.~(\ref{eqn:tau}) by invoking the change of variables,
\begin{equation}
    g_1 = - \frac{\sqrt{\zeta} \psi}{\sqrt{n_0} \phi} \tau_2\,,\quad\text{and}\quad
    g_5 = \big(\gamma +\sigma _{W_2}^2 \big(\eta '-\zeta \big)\big)\tau_1\,.
\end{equation}
The error $E_{31}$ can then be written in terms of $\tau_1$ and its derivative $\tau_1'$~(\ref{eq:dtau1}),
\begin{equation}
    E_{31} = \sigma_{\varepsilon}^2\big(-\tau_1'/\tau_1^2-1\big)\,.
\end{equation}

\subsubsection{$E_{32}$}
The equations satisfied by the operator-valued Stieltjes transform $G$ of $\bar{Q}_{32}$ induce the following structure on $G$,
\eq{
G = \begin{pmatrix} 0 & G_{12} \\ G_{12}^\top & 0 \end{pmatrix}\,,
}
where,
\eq{
G_{12} = \left(
\begin{array}{ccccccccc}
 g_9 & 0 & 0 & 0 & 0 & 0 & g_6 & 0 & 0 \\
 0 & g_1 & g_3 & 0 & g_4 & g_7 & 0 & 0 & g_2 \\
 0 & 0 & g_{10} & 0 & g_4 & g_{13} & 0 & 0 & g_5 \\
 0 & 0 & 0 & g_{12} & 0 & 0 & 0 & g_{16} & 0 \\
 0 & 0 & g_{15} & 0 & g_{11} & g_{14} & 0 & 0 & g_8 \\
 0 & 0 & 0 & 0 & 0 & g_{10} & 0 & 0 & g_4 \\
 0 & 0 & 0 & 0 & 0 & 0 & g_9 & 0 & 0 \\
 0 & 0 & 0 & 0 & 0 & 0 & 0 & g_{12} & 0 \\
 0 & 0 & 0 & 0 & 0 & g_{15} & 0 & 0 & g_{11} \\
\end{array}
\right)\,,
}
and the independent entry-wise component functions $g_i$ give the error $E_{32}$ through the relation,
\eq{
E_{32} = -g_2 n_0^{3/2}/(\sqrt{\zeta } \psi )\,,
}
and themselves satisfy the following system of polynomial equations,
{\small
\begin{subequations}
\begin{align}
0 &= \sqrt{\zeta } g_{10} g_{12} \sqrt{n_0}-g_{15} \psi\\
0 &= \sqrt{\zeta } g_4 g_{12} \sqrt{n_0}-g_{11} \psi +\psi\\
0 &= -\sqrt{\zeta } g_9 g_{10} \psi -g_4 \sqrt{n_0} \phi  \big(\gamma +\sigma _{W_2}^2 \big(\eta '-\zeta \big)\big)\\
0 &= -\sqrt{\zeta } g_9 g_{15} \psi -\big(g_{11}-1\big) \sqrt{n_0} \phi  \big(\gamma +\sigma _{W_2}^2 \big(\eta '-\zeta \big)\big)\\
0 &= -\sqrt{\zeta } g_9 \psi -\sqrt{\zeta } g_3 g_9 \psi -g_4 \sqrt{n_0} \phi  \big(\gamma +\sigma _{W_2}^2 \big(\eta '-\zeta \big)\big)\\
0 &= -\sqrt{\zeta } g_6 g_{10} \psi -\sqrt{\zeta } g_9 g_{13} \psi -g_5 \sqrt{n_0} \phi  \big(\gamma +\sigma _{W_2}^2 \big(\eta '-\zeta \big)\big)\\
0 &= -\sqrt{\zeta } g_9 g_{14} \psi -\sqrt{\zeta } g_6 g_{15} \psi -g_8 \sqrt{n_0} \phi  \big(\gamma +\sigma _{W_2}^2 \big(\eta '-\zeta \big)\big)\\
0 &= \sqrt{\zeta } g_5 g_{12} n_0+\sqrt{\zeta } g_4 \big(g_{16} n_0+g_{12} \big(\zeta  \psi -\eta  \psi -\zeta  \sigma _{W_2}^2\big)\big)+g_8 \sqrt{n_0} (-\psi )\\
0 &= g_4 \sqrt{n_0} \phi  \big(\gamma +\sigma _{W_2}^2 \big(\eta '-\zeta \big)\big)+g_9 \big(\sqrt{\zeta } g_{11} \psi +\zeta  g_4 \sqrt{n_0} \sigma _{W_2}^2\big)\\
0 &= g_3 \sqrt{n_0} \phi  \big(\gamma +\sigma _{W_2}^2 \big(\eta '-\zeta \big)\big)+g_9 \big(\sqrt{\zeta } g_{15} \psi +\zeta  g_{10} \sqrt{n_0} \sigma _{W_2}^2\big)\\
0 &= \sqrt{\zeta } g_{12} g_{13} n_0+\sqrt{\zeta } g_{10} \big(g_{16} n_0+g_{12} \big(\zeta  \psi -\eta  \psi -\zeta  \sigma _{W_2}^2\big)\big)+g_{14} \sqrt{n_0} (-\psi )\\
0 &= \big(g_{10}-1\big) \sqrt{n_0} \phi  \big(\gamma +\sigma _{W_2}^2 \big(\eta '-\zeta \big)\big)+g_9 \big(\sqrt{\zeta } g_{15} \psi +\zeta  g_{10} \sqrt{n_0} \sigma _{W_2}^2\big)\\
0 &= -\sqrt{\zeta } \big(\big(g_1+g_3\big) g_6+g_7 g_9\big) \psi -\gamma  g_2 \sqrt{n_0} \phi +\zeta  g_2 \sqrt{n_0} \phi  \sigma _{W_2}^2+g_2 \sqrt{n_0} (-\phi ) \eta ' \sigma _{W_2}^2\\
0 &= \sqrt{\zeta } g_{11} g_{12} \sqrt{n_0} \phi  \big(\sigma _{W_2}^2 \big(\zeta -\eta '\big)-\gamma \big)+g_{15} \psi  \big(\gamma  \phi +\sigma _{W_2}^2 \big(-\zeta  \phi +\phi  \eta '+\zeta  g_9\big)\big)\\
0 &= g_{12} \big(g_9 \psi  (\zeta -\eta )+\phi  \big(\sqrt{\zeta } g_4 \sqrt{n_0}-1\big) \big(\gamma +\sigma _{W_2}^2 \big(\eta '-\zeta \big)\big)\big)+\phi  \big(\gamma +\sigma _{W_2}^2 \big(\eta '-\zeta \big)\big)\\
0 &= g_{10} \psi  \big(\gamma  \phi +\sigma _{W_2}^2 \big(-\zeta  \phi +\phi  \eta '+\zeta  g_9\big)\big)-\phi  \big(\sqrt{\zeta } g_4 g_{12} \sqrt{n_0}+\psi \big) \big(\gamma +\sigma _{W_2}^2 \big(\eta '-\zeta \big)\big)\\
0 &= g_9 \big(\sqrt{\zeta } g_{15} \psi +\sqrt{n_0} \big(\gamma +g_{12} (\eta -\zeta )+\sigma _{W_2}^2 \big(\eta '+\zeta  \big(g_{10}-1\big)\big)\big)\big)-\sqrt{n_0} \big(\gamma +\sigma _{W_2}^2 \big(\eta '-\zeta \big)\big)\\
0 &= -\sqrt{\zeta } g_4 g_{12} \sqrt{n_0} \phi  \big(\gamma +\sigma _{W_2}^2 \big(\eta '-\zeta \big)\big)+g_3 \psi  \big(\gamma  \phi +\sigma _{W_2}^2 \big(-\zeta  \phi +\phi  \eta '+\zeta  g_9\big)\big)+\zeta  g_9 \psi  \sigma _{W_2}^2\\
0 &= g_7 n_0 \phi  \big(\gamma +\sigma _{W_2}^2 \big(\eta '-\zeta \big)\big)+g_6 \big(\sqrt{\zeta } g_{15} \sqrt{n_0} \psi +\zeta  g_{10} n_0 \sigma _{W_2}^2\big)+g_9 \big(\sqrt{\zeta } g_{14} \sqrt{n_0} \psi +\zeta  \sigma _{W_2}^2 \big(g_{13} n_0-\zeta  g_{10} \sigma _{W_2}^2\big)\big)\\
0 &= \gamma  g_2 n_0 \phi +\sqrt{\zeta } g_8 g_9 \sqrt{n_0} \psi +g_6 \big(\sqrt{\zeta } g_{11} \sqrt{n_0} \psi +\zeta  g_4 n_0 \sigma _{W_2}^2\big)-\zeta  g_2 n_0 \phi  \sigma _{W_2}^2+\zeta  g_5 g_9 n_0 \sigma _{W_2}^2\nonumber\\
&\quad\quad+g_2 n_0 \phi  \eta ' \sigma _{W_2}^2-\zeta ^2 g_4 g_9 \sigma _{W_2}^4\\
0 &= g_6 \big(-\sqrt{\zeta } g_{15} \sqrt{n_0} \psi -n_0 \big(\gamma +g_{12} (\eta -\zeta )+\sigma _{W_2}^2 \big(\eta '+\zeta  \big(g_{10}-1\big)\big)\big)\big)+g_9 \big(g_{12} \psi  (\zeta -\eta )^2\nonumber\\
&\quad\quad+g_{16} n_0 (\zeta -\eta )-\sqrt{\zeta } g_{14} \sqrt{n_0} \psi -\zeta  g_{13} n_0 \sigma _{W_2}^2+\zeta ^2 g_{10} \sigma _{W_2}^4\big)\\
0 &= \gamma  g_5 n_0 \phi +\sqrt{\zeta } g_8 g_9 \sqrt{n_0} \psi +\sqrt{\zeta } g_6 g_{11} \sqrt{n_0} \psi +\zeta  g_4 \big(g_6 n_0 \sigma _{W_2}^2+g_{12} \phi  \big(\gamma +\sigma _{W_2}^2 \big(\eta '-\zeta \big)\big)-\zeta  g_9 \sigma _{W_2}^4\big)\nonumber\\
&\quad\quad-\zeta  g_5 n_0 \phi  \sigma _{W_2}^2+\zeta  g_5 g_9 n_0 \sigma _{W_2}^2+g_5 n_0 \phi  \eta ' \sigma _{W_2}^2\\
0 &= \gamma  g_{13} n_0 \phi +\sqrt{\zeta } g_6 g_{15} \sqrt{n_0} \psi +\zeta  g_{10} \big(g_6 n_0 \sigma _{W_2}^2+g_{12} \phi  \big(\gamma +\sigma _{W_2}^2 \big(\eta '-\zeta \big)\big)-\zeta  g_9 \sigma _{W_2}^4\big)\nonumber\\
&\quad\quad+g_9 \big(\sqrt{\zeta } g_{14} \sqrt{n_0} \psi +\zeta  g_{13} n_0 \sigma _{W_2}^2\big)-\zeta  g_{13} n_0 \phi  \sigma _{W_2}^2+g_{13} n_0 \phi  \eta ' \sigma _{W_2}^2\\
0 &= -\sqrt{\zeta } g_{12} \phi  \big(\gamma +\sigma _{W_2}^2 \big(\eta '-\zeta \big)\big) \big(\sqrt{n_0} \big(g_8 n_0+g_{11} \big(\zeta  \psi -\eta  \psi -\zeta  \sigma _{W_2}^2\big)\big)-\sqrt{\zeta } g_{15} \psi \big)\nonumber\\
&\quad\quad+g_{14} n_0 \psi  \big(\gamma  \phi +\sigma _{W_2}^2 \big(-\zeta  \phi +\phi  \eta '+\zeta  g_9\big)\big)-\sqrt{\zeta } g_{11} g_{16} n_0^{3/2} \phi  \big(\gamma +\sigma _{W_2}^2 \big(\eta '-\zeta \big)\big)+\zeta  g_{15} \psi  \sigma _{W_2}^2 \big(g_6 n_0-\zeta  g_9 \sigma _{W_2}^2\big)\\
0 &= g_9 \psi  (-(\zeta -\eta )) \big(g_{12} \psi  (\zeta -\eta )+g_{16} n_0\big)-\sqrt{\zeta } g_4 \sqrt{n_0} \phi  \big(\gamma +\sigma _{W_2}^2 \big(\eta '-\zeta \big)\big) \big(g_{16} n_0+g_{12} \big(\zeta  \psi -\eta  \psi -\zeta  \sigma _{W_2}^2\big)\big)\nonumber\\
&\quad\quad+n_0 \big(g_6 g_{12} \psi  (\eta -\zeta )+\phi  \big(g_{16}-\sqrt{\zeta } g_5 g_{12} \sqrt{n_0}\big) \big(\gamma +\sigma _{W_2}^2 \big(\eta '-\zeta \big)\big)\big)+\zeta  g_{10} g_{12} \psi  \phi  \big(\gamma +\sigma_{W_2}^2 \big(\eta '-\zeta \big)\big)\\
0 &= g_{13} n_0 \psi  \big(\gamma  \phi +\sigma _{W_2}^2 \big(-\zeta  \phi +\phi  \eta '+\zeta  g_9\big)\big)-\sqrt{\zeta } g_4 \sqrt{n_0} \phi  \big(\gamma +\sigma _{W_2}^2 \big(\eta '-\zeta \big)\big)+\sqrt{\zeta } g_5 g_{12} n_0^{3/2} \phi  \big(\sigma _{W_2}^2 \big(\zeta -\eta '\big)-\gamma \big) \nonumber\\
&\quad\quad\big(g_{16} n_0+g_{12} \big(\zeta  \psi -\eta  \psi -\zeta  \sigma _{W_2}^2\big)\big)+\zeta  g_{10} \psi  \big(g_6 n_0 \sigma _{W_2}^2+g_{12} \phi  \big(\gamma +\sigma _{W_2}^2 \big(\eta '-\zeta \big)\big)-\zeta  g_9 \sigma _{W_2}^4\big)\\
0 &= -\gamma  \sqrt{\zeta } g_2 g_{12} n_0^{3/2} \phi +\gamma  g_7 n_0 \psi  \phi -\sqrt{\zeta } g_4 \sqrt{n_0} \phi  \big(\gamma +\sigma _{W_2}^2 \big(\eta '-\zeta \big)\big) \big(g_{16} n_0+g_{12} \big(\zeta  \psi -\eta  \psi -\zeta  \sigma _{W_2}^2\big)\big)\nonumber\\
&\quad\quad+\zeta  g_3 \psi  \big(g_6 n_0 \sigma _{W_2}^2+g_{12} \phi  \big(\gamma +\sigma _{W_2}^2 \big(\eta '-\zeta \big)\big)-\zeta  g_9 \sigma _{W_2}^4\big)+\zeta ^{3/2} g_2 g_{12} n_0^{3/2} \phi  \sigma _{W_2}^2-\zeta ^2 g_9 \psi  \sigma _{W_2}^4\nonumber\\
&\quad\quad+n_0 \phi  \eta ' \sigma _{W_2}^2 \big(g_7 \psi -\sqrt{\zeta } g_2 g_{12} \sqrt{n_0}\big)+\zeta  g_6 n_0 \psi  \sigma _{W_2}^2+\zeta  g_7 g_9 n_0 \psi  \sigma _{W_2}^2-\zeta  g_7 n_0 \psi  \phi  \sigma _{W_2}^2
\end{align}
\end{subequations}
}
After some straightforward algebra, one can eliminate all $g_i$ except for $g_4$ and $g_9$, which satisfy coupled polynomial equations. Those equations can be shown to be identical to eqn.~(\ref{eqn:tau}) by invoking the change of variables,
\begin{equation}
    g_4 = - \frac{\sqrt{\zeta} \psi}{\sqrt{n_0} \phi} \tau_2\,,\quad\text{and}\quad
    g_9 = \big(\gamma +\sigma _{W_2}^2 \big(\eta '-\zeta \big)\big)\tau_1\,.
\end{equation}
In terms of $\tau_1$, $\tau_2$, and $\tau_2'$~(\ref{eq:dtau2}), the error $E_{32}$ is given by, 
\begin{equation}
    E_{32} = 1-2\tau_2/\tau_1-\tau_2'/\tau_1^2\,.
\end{equation}

\subsubsection{$E_{33}$}
The equations satisfied by the operator-valued Stieltjes transform $G$ of $\bar{Q}_{32}$ induce the following structure on $G$,
\eq{
G = \begin{pmatrix} 0 & G_{12} \\ G_{12}^\top & 0 \end{pmatrix}\,,
}
where,
\eq{
G_{12} = \left(
\begin{array}{ccccccccccc}
 g_{13} & 0 & 0 & 0 & 0 & 0 & 0 & g_8 & 0 & 0 & 0 \\
 0 & g_1 & 0 & 0 & g_5 & 0 & 0 & 0 & g_{11} & g_3 & 0 \\
 0 & 0 & g_1 & g_4 & 0 & g_6 & g_9 & 0 & 0 & 0 & g_2 \\
 0 & 0 & 0 & g_{14} & 0 & g_6 & g_{17} & 0 & 0 & 0 & g_7 \\
 0 & 0 & 0 & 0 & g_{16} & 0 & 0 & 0 & g_{20} & g_{12} & 0 \\
 0 & 0 & 0 & g_{19} & 0 & g_{15} & g_{18} & 0 & 0 & 0 & g_{10} \\
 0 & 0 & 0 & 0 & 0 & 0 & g_{14} & 0 & 0 & 0 & g_6 \\
 0 & 0 & 0 & 0 & 0 & 0 & 0 & g_{13} & 0 & 0 & 0 \\
 0 & 0 & 0 & 0 & 0 & 0 & 0 & 0 & g_1 & g_5 & 0 \\
 0 & 0 & 0 & 0 & 0 & 0 & 0 & 0 & 0 & g_{16} & 0 \\
 0 & 0 & 0 & 0 & 0 & 0 & g_{19} & 0 & 0 & 0 & g_{15} \\
\end{array}
\right)\,,
}
and the independent entry-wise component functions $g_i$ give the error $E_{32}$ through the relation,
\eq{
E_{33} =-g_3 n_0 \sigma _{W_2}^2/\psi\,,
}
and themselves satisfy the following system of polynomial equations,
{\small
\begin{subequations}
\begin{align}
0 &= \sqrt{\zeta } g_{14} g_{16} \sqrt{n_0}-g_{19} \psi\\
0 &= \sqrt{\zeta } g_6 g_{16} \sqrt{n_0}-g_{15} \psi +\psi\\
0 &= -\sqrt{\zeta } g_{13} g_{14} \psi -g_6 \sqrt{n_0} \phi  \big(\gamma +\sigma _{W_2}^2 \big(\eta '-\zeta \big)\big)\\
0 &= -\sqrt{\zeta } g_{13} g_{19} \psi -\big(g_{15}-1\big) \sqrt{n_0} \phi  \big(\gamma +\sigma _{W_2}^2 \big(\eta '-\zeta \big)\big)\\
0 &= -\sqrt{\zeta } g_{13} \psi -\sqrt{\zeta } g_4 g_{13} \psi -g_6 \sqrt{n_0} \phi  \big(\gamma +\sigma _{W_2}^2 \big(\eta '-\zeta \big)\big)\\
0 &= -\sqrt{\zeta } g_8 g_{14} \psi -\sqrt{\zeta } g_{13} g_{17} \psi -g_7 \sqrt{n_0} \phi  \big(\gamma +\sigma _{W_2}^2 \big(\eta '-\zeta \big)\big)\\
0 &= -\sqrt{\zeta } g_{13} g_{18} \psi -\sqrt{\zeta } g_8 g_{19} \psi -g_{10} \sqrt{n_0} \phi  \big(\gamma +\sigma _{W_2}^2 \big(\eta '-\zeta \big)\big)\\
0 &= g_{13} g_{16} \psi  (\zeta -\eta )-\phi  \big(g_5-\sqrt{\zeta } g_6 g_{16} \sqrt{n_0}\big) \big(\gamma +\sigma _{W_2}^2 \big(\eta '-\zeta \big)\big)\\
0 &= g_6 \sqrt{n_0} \phi  \big(\gamma +\sigma _{W_2}^2 \big(\eta '-\zeta \big)\big)+g_{13} \big(\sqrt{\zeta } g_{15} \psi +\zeta  g_6 \sqrt{n_0} \sigma _{W_2}^2\big)\\
0 &= g_4 \sqrt{n_0} \phi  \big(\gamma +\sigma _{W_2}^2 \big(\eta '-\zeta \big)\big)+g_{13} \big(\sqrt{\zeta } g_{19} \psi +\zeta  g_{14} \sqrt{n_0} \sigma _{W_2}^2\big)\\
0 &= \big(g_{14}-1\big) \sqrt{n_0} \phi  \big(\gamma +\sigma _{W_2}^2 \big(\eta '-\zeta \big)\big)+g_{13} \big(\sqrt{\zeta } g_{19} \psi +\zeta  g_{14} \sqrt{n_0} \sigma _{W_2}^2\big)\\
0 &= -\sqrt{\zeta } \big(\big(g_4+1\big) g_8+g_9 g_{13}\big) \psi -\gamma  g_2 \sqrt{n_0} \phi +\zeta  g_2 \sqrt{n_0} \phi  \sigma _{W_2}^2+g_2 \sqrt{n_0} (-\phi ) \eta ' \sigma _{W_2}^2\\
0 &= \sqrt{\zeta } g_{15} g_{16} \sqrt{n_0} \phi  \big(\sigma _{W_2}^2 \big(\zeta -\eta '\big)-\gamma \big)+g_{19} \psi  \big(\gamma  \phi +\sigma _{W_2}^2 \big(-\zeta  \phi +\phi  \eta '+\zeta  g_{13}\big)\big)\\
0 &= g_{16} \big(g_{13} \psi  (\zeta -\eta )+\phi  \big(\sqrt{\zeta } g_6 \sqrt{n_0}-1\big) \big(\gamma +\sigma _{W_2}^2 \big(\eta '-\zeta \big)\big)\big)+\phi  \big(\gamma +\sigma _{W_2}^2 \big(\eta '-\zeta \big)\big)\\
0 &= g_{13} \big(\sqrt{\zeta } g_{19} \psi +\sqrt{n_0} \big(\gamma +g_{16} (\eta -\zeta )+\sigma _{W_2}^2 \big(\eta '+\zeta  \big(g_{14}-1\big)\big)\big)\big)-\sqrt{n_0} \big(\gamma +\sigma _{W_2}^2 \big(\eta '-\zeta \big)\big)\\
0 &= g_{14} \psi  \big(\gamma  \phi +\sigma _{W_2}^2 \big(\phi  \big(\eta '-\zeta \big)+\zeta  g_{13}\big)\big)-\phi  \big(\sqrt{\zeta } g_6 g_{16} \sqrt{n_0}+\psi \big) \big(\gamma +\sigma _{W_2}^2 \big(\eta '-\zeta \big)\big)\\
0 &= -\sqrt{\zeta } g_6 g_{16} \sqrt{n_0} \phi  \big(\gamma +\sigma _{W_2}^2 \big(\eta '-\zeta \big)\big)+g_4 \psi  \big(\gamma  \phi +\sigma _{W_2}^2 \big(-\zeta  \phi +\phi  \eta '+\zeta  g_{13}\big)\big)+\zeta  g_{13} \psi  \sigma _{W_2}^2\\
0 &= \sqrt{\zeta } \big(g_7 g_{16}+g_6 \big(g_{12}+g_{20}\big)\big) n_0+g_{10} \sqrt{n_0} (-\psi )+\sqrt{\zeta } g_6 \big(\psi  (\zeta -\eta )-\zeta  \sigma _{W_2}^2\big)+\sqrt{\zeta } g_5 g_6 \big(\zeta  \psi -\eta  \psi -\zeta  \sigma _{W_2}^2\big)\\
0 &= \sqrt{\zeta } \big(g_{16} g_{17}+g_{14} \big(g_{12}+g_{20}\big)\big) n_0+g_{18} \sqrt{n_0} (-\psi )+\sqrt{\zeta } g_{14} \big(\psi  (\zeta -\eta )-\zeta  \sigma _{W_2}^2\big)+\sqrt{\zeta } g_5 g_{14} \big(\zeta  \psi -\eta  \psi -\zeta  \sigma _{W_2}^2\big)\\
0 &= g_{13} \psi  (\zeta -\eta )+g_5 \big(g_{13} \psi  (\zeta -\eta )+\phi  \big(\sqrt{\zeta } g_6 \sqrt{n_0}-1\big) \big(\gamma +\sigma _{W_2}^2 \big(\eta '-\zeta \big)\big)\big)+\sqrt{\zeta } g_6 \sqrt{n_0} \phi  \big(\gamma +\sigma _{W_2}^2 \big(\eta '-\zeta \big)\big)\\
0 &= g_9 n_0 \phi  \big(\gamma +\sigma _{W_2}^2 \big(\eta '-\zeta \big)\big)+g_8 \big(\sqrt{\zeta } g_{19} \sqrt{n_0} \psi +\zeta  g_{14} n_0 \sigma _{W_2}^2\big)+g_{13} \big(\sqrt{\zeta } g_{18} \sqrt{n_0} \psi +\zeta  \sigma _{W_2}^2 \big(g_{17} n_0-\zeta  g_{14} \sigma _{W_2}^2\big)\big)\\
0 &= \gamma  g_2 n_0 \phi +\sqrt{\zeta } g_{10} g_{13} \sqrt{n_0} \psi +g_8 \big(\sqrt{\zeta } g_{15} \sqrt{n_0} \psi +\zeta  g_6 n_0 \sigma _{W_2}^2\big)-\zeta  g_2 n_0 \phi  \sigma _{W_2}^2\nonumber\\
&\quad\quad+\zeta  g_7 g_{13} n_0 \sigma _{W_2}^2+g_2 n_0 \phi  \eta ' \sigma _{W_2}^2-\zeta ^2 g_6 g_{13} \sigma _{W_2}^4\\
0 &= g_{13} g_{16} \psi  (-(\zeta -\eta )) \big(\psi  (\zeta -\eta )-\zeta  \sigma _{W_2}^2\big)\nonumber\\
&\quad\quad-\phi  \big(\gamma +\sigma _{W_2}^2 \big(\eta '-\zeta \big)\big) \big(-\zeta  g_{14} g_{16} \psi +\sqrt{\zeta } g_6 g_{16} \sqrt{n_0} \big(\zeta  \psi -\eta  \psi -\zeta  \sigma _{W_2}^2\big)-g_{20} n_0\big)\\
0 &= -\sqrt{\zeta } \phi  \big(\gamma +\sigma _{W_2}^2 \big(\eta '-\zeta \big)\big) \big(g_6 \sqrt{n_0} \big(\zeta  \psi -\eta  \psi -\zeta  \sigma _{W_2}^2\big)-\sqrt{\zeta } g_{14} \psi \big)+g_{20} n_0 \big(g_{13} \psi  (\eta -\zeta )\nonumber\\
&\quad\quad-\phi  \big(\sqrt{\zeta } g_6 \sqrt{n_0}-1\big) \big(\gamma +\sigma _{W_2}^2 \big(\eta '-\zeta \big)\big)\big)+g_{13} \psi  (-(\zeta -\eta )) \big(\psi  (\zeta -\eta )-\zeta  \sigma _{W_2}^2\big)\\
0 &= \big(\psi  (\zeta -\eta )-\zeta  \sigma _{W_2}^2\big) \big(g_{13} \psi  (\eta -\zeta )+\sqrt{\zeta } g_6 \sqrt{n_0} \phi  \big(\sigma _{W_2}^2 \big(\zeta -\eta '\big)-\gamma \big)\big)+n_0 \big(g_{13} g_{20} \psi  (\eta -\zeta )\nonumber\\
&\quad\quad+\phi  \big(g_{11}-\sqrt{\zeta } g_6 g_{20} \sqrt{n_0}\big) \big(\gamma +\sigma _{W_2}^2 \big(\eta '-\zeta \big)\big)\big)+\zeta  g_4 \psi  \phi  \big(\gamma +\sigma _{W_2}^2 \big(\eta '-\zeta \big)\big)\\
0 &= \gamma  g_7 n_0 \phi +\sqrt{\zeta } g_{10} g_{13} \sqrt{n_0} \psi +\sqrt{\zeta } g_8 g_{15} \sqrt{n_0} \psi -\zeta  g_7 n_0 \phi  \sigma _{W_2}^2+\zeta  g_6 g_8 n_0 \sigma _{W_2}^2+\zeta  g_7 g_{13} n_0 \sigma _{W_2}^2+g_7 n_0 \phi  \eta ' \sigma _{W_2}^2\nonumber\\
&\quad\quad+\zeta  g_6 \phi  \big(\gamma +\sigma _{W_2}^2 \big(\eta '-\zeta \big)\big)+\zeta  g_5 g_6 \phi  \big(\gamma +\sigma _{W_2}^2 \big(\eta '-\zeta \big)\big)-\zeta ^2 g_6 g_{13} \sigma _{W_2}^4\\
0 &= \gamma  g_{17} n_0 \phi +\sqrt{\zeta } g_{13} g_{18} \sqrt{n_0} \psi +\sqrt{\zeta } g_8 g_{19} \sqrt{n_0} \psi -\zeta  g_{17} n_0 \phi  \sigma _{W_2}^2+\zeta  g_8 g_{14} n_0 \sigma _{W_2}^2+\zeta  g_{13} g_{17} n_0 \sigma _{W_2}^2+g_{17} n_0 \phi  \eta ' \sigma _{W_2}^2\nonumber\\
&\quad\quad+\zeta  g_{14} \phi  \big(\gamma +\sigma _{W_2}^2 \big(\eta '-\zeta \big)\big)+\zeta  g_5 g_{14} \phi  \big(\gamma +\sigma _{W_2}^2 \big(\eta '-\zeta \big)\big)-\zeta ^2 g_{13} g_{14} \sigma _{W_2}^4\\
0 &= g_5 \big(\psi  (\zeta -\eta )-\zeta  \sigma _{W_2}^2\big) \big(g_{13} \psi  (\eta -\zeta )+\sqrt{\zeta } g_6 \sqrt{n_0} \phi  \big(\sigma _{W_2}^2 \big(\zeta -\eta '\big)-\gamma \big)\big)+n_0 \big(\phi  \big(g_3-\sqrt{\zeta } \big(g_6 g_{12}+g_2 g_{16}\big) \sqrt{n_0}\big) \nonumber\\
&\quad\quad\big(\gamma +\sigma _{W_2}^2 \big(\eta '-\zeta \big)\big)-\big(g_{12} g_{13}+g_8 g_{16}\big) \psi  (\zeta -\eta )\big)+\zeta  g_4 g_5 \psi  \phi  \big(\gamma +\sigma _{W_2}^2 \big(\eta '-\zeta \big)\big)\\
0 &= \big(\psi  (\zeta -\eta )-\zeta  \sigma _{W_2}^2\big) \big(g_{13} \psi  (\eta -\zeta )+\sqrt{\zeta } g_6 \sqrt{n_0} \phi  \big(\sigma _{W_2}^2 \big(\zeta -\eta '\big)-\gamma \big)\big)+g_5 \big(g_{13} \psi  (-(\zeta -\eta )) \big(\psi  (\zeta -\eta )-\zeta  \sigma _{W_2}^2\big)\nonumber\\
&\quad\quad-\sqrt{\zeta } \phi  \big(\gamma +\sigma _{W_2}^2 \big(\eta '-\zeta \big)\big) \big(g_6 \sqrt{n_0} \big(\zeta  \psi -\eta  \psi -\zeta  \sigma _{W_2}^2\big)-\sqrt{\zeta } g_{14} \psi \big)\big)+g_{11} n_0 \phi  \big(\gamma +\sigma _{W_2}^2 \big(\eta '-\zeta \big)\big)\nonumber\\
&\quad\quad+\zeta  g_4 \psi  \phi  \big(\gamma +\sigma _{W_2}^2 \big(\eta '-\zeta \big)\big)\\
0 &= g_{12} n_0 \big(g_{13} \psi  (\eta -\zeta )-\phi  \big(\sqrt{\zeta } g_6 \sqrt{n_0}-1\big) \big(\gamma +\sigma _{W_2}^2 \big(\eta '-\zeta \big)\big)\big)-g_{16} \big(g_8 n_0 \psi  (\zeta -\eta )+\sqrt{\zeta } g_7 n_0^{3/2} \phi  \big(\gamma +\sigma _{W_2}^2 \big(\eta '-\zeta \big)\big)\nonumber\\
&\quad\quad+\zeta  g_{13} \psi  (\zeta -\eta ) \sigma _{W_2}^2\big)+g_5 \big(g_{13} \psi  (-(\zeta -\eta )) \big(\psi  (\zeta -\eta )-\zeta  \sigma _{W_2}^2\big)\nonumber\\
&\quad\quad-\sqrt{\zeta } \phi  \big(\gamma +\sigma _{W_2}^2 \big(\eta '-\zeta \big)\big) \big(g_6 \sqrt{n_0} \big(\zeta  \psi -\eta  \psi -\zeta  \sigma _{W_2}^2\big)-\sqrt{\zeta } g_{14} \psi \big)\big)\\
0 &= \gamma  \sqrt{\zeta } g_7 g_{16} n_0^{3/2} \phi +\gamma  \sqrt{\zeta } g_6 g_{20} n_0^{3/2} \phi +g_8 g_{16} n_0 \psi  (\zeta -\eta )+\zeta  g_{13} g_{20} n_0 \psi -\eta  g_{13} g_{20} n_0 \psi \nonumber\\
&\quad\quad+g_{12} n_0 \big(g_{13} \psi  (\zeta -\eta )+\phi  \big(\sqrt{\zeta } g_6 \sqrt{n_0}-1\big) \big(\gamma +\sigma _{W_2}^2 \big(\eta '-\zeta \big)\big)\big)-\zeta ^{3/2} g_7 g_{16} n_0^{3/2} \phi  \sigma _{W_2}^2-\zeta ^{3/2} g_6 g_{20} n_0^{3/2} \phi  \sigma _{W_2}^2\nonumber\\
&\quad\quad+\sqrt{\zeta } \big(g_7 g_{16}+g_6 g_{20}\big) n_0^{3/2} \phi  \eta ' \sigma _{W_2}^2+\zeta ^2 g_{13} g_{16} \psi  \sigma _{W_2}^2-\zeta  \eta  g_{13} g_{16} \psi  \sigma _{W_2}^2\\
0 &= -\gamma  g_8 n_0-\sqrt{\zeta } g_{13} g_{18} \sqrt{n_0} \psi -\sqrt{\zeta } g_8 g_{19} \sqrt{n_0} \psi +\zeta  g_{12} g_{13} n_0+\zeta  g_8 g_{16} n_0+\zeta  g_{13} g_{20} n_0-\eta  g_{12} g_{13} n_0\nonumber\\
&\quad\quad-\eta  g_8 g_{16} n_0-\eta  g_{13} g_{20} n_0+\zeta  g_8 n_0 \sigma _{W_2}^2-\zeta  g_8 g_{14} n_0 \sigma _{W_2}^2-\zeta  g_{13} g_{17} n_0 \sigma _{W_2}^2-g_8 n_0 \eta ' \sigma _{W_2}^2+\zeta ^2 g_{13} g_{14} \sigma _{W_2}^4\nonumber\\
&\quad\quad+\zeta ^2 g_{13} g_{16} \sigma _{W_2}^2+g_{13} (\zeta -\eta ) \big(\psi  (\zeta -\eta )-\zeta  \sigma _{W_2}^2\big)+g_5 g_{13} (\zeta -\eta ) \big(\zeta  \psi -\eta  \psi -\zeta  \sigma _{W_2}^2\big)-\zeta  \eta  g_{13} g_{16} \sigma _{W_2}^2\\
0 &= \gamma  \sqrt{\zeta } g_5 g_7 n_0^{3/2} \phi +\gamma  \sqrt{\zeta } g_6 g_{11} n_0^{3/2} \phi +\zeta  g_5 g_8 n_0 \psi +\zeta  g_{11} g_{13} n_0 \psi -\eta  g_5 g_8 n_0 \psi -\eta  g_{11} g_{13} n_0 \psi\nonumber\\
&\quad\quad +g_3 n_0 \big(g_{13} \psi  (\zeta -\eta )+\phi  \big(\sqrt{\zeta } g_6 \sqrt{n_0}-1\big) \big(\gamma +\sigma _{W_2}^2 \big(\eta '-\zeta \big)\big)\big)+n_0 \big(g_8 \psi  (\zeta -\eta )\nonumber\\
&\quad\quad+\sqrt{\zeta } g_2 \sqrt{n_0} \phi  \big(\gamma +\sigma _{W_2}^2 \big(\eta '-\zeta \big)\big)\big)-\zeta ^{3/2} g_5 g_7 n_0^{3/2} \phi  \sigma _{W_2}^2-\zeta ^{3/2} g_6 g_{11} n_0^{3/2} \phi  \sigma _{W_2}^2+\sqrt{\zeta } g_5 g_7 n_0^{3/2} \phi  \eta ' \sigma _{W_2}^2\nonumber\\
&\quad\quad+\sqrt{\zeta } g_6 g_{11} n_0^{3/2} \phi  \eta ' \sigma _{W_2}^2+\zeta ^2 g_5 g_{13} \psi  \sigma _{W_2}^2-\zeta  \eta  g_5 g_{13} \psi  \sigma _{W_2}^2\\
0 &= -\sqrt{\zeta } g_6 \sqrt{n_0} \phi  \big(\psi  (\zeta -\eta )-\zeta  \sigma _{W_2}^2\big) \big(\gamma +\sigma _{W_2}^2 \big(\eta '-\zeta \big)\big)-\sqrt{\zeta } g_5 g_6 \sqrt{n_0} \phi  \big(\psi  (\zeta -\eta )-\zeta  \sigma _{W_2}^2\big) \big(\gamma +\sigma _{W_2}^2 \big(\eta '-\zeta \big)\big)\nonumber\\
&\quad\quad+g_9 n_0 \psi  \big(\gamma  \phi +\sigma _{W_2}^2 \big(\phi  \big(\eta '-\zeta \big)+\zeta  g_{13}\big)\big)+\zeta  g_4 \psi  \big(g_8 n_0 \sigma _{W_2}^2+g_5 \phi  \big(\gamma +\sigma _{W_2}^2 \big(\eta '-\zeta \big)\big)-\zeta  g_{13} \sigma _{W_2}^4\big)\nonumber\\
&\quad\quad-\sqrt{\zeta } \big(g_2 g_{16}+g_6 \big(g_{12}+g_{20}\big)\big) n_0^{3/2} \phi  \big(\gamma +\sigma _{W_2}^2 \big(\eta '-\zeta \big)\big)+\zeta  \psi  \sigma _{W_2}^2 \big(g_8 n_0-\zeta  g_{13} \sigma _{W_2}^2\big)\nonumber\\
&\quad\quad+\zeta  g_4 \psi  \phi  \big(\gamma +\sigma _{W_2}^2 \big(\eta '-\zeta \big)\big)\\
0 &= -\gamma  \sqrt{\zeta } g_6 g_{12} n_0^{3/2} \phi -\gamma  \sqrt{\zeta } g_7 g_{16} n_0^{3/2} \phi -\gamma  \sqrt{\zeta } g_6 g_{20} n_0^{3/2} \phi +\gamma  g_{17} n_0 \psi  \phi -\sqrt{\zeta } \phi  \big(\gamma +\sigma _{W_2}^2 \big(\eta '-\zeta \big)\big) \nonumber\\
&\quad\quad\big(g_6 \sqrt{n_0} \big(\zeta  \psi -\eta  \psi -\zeta  \sigma _{W_2}^2\big)-\sqrt{\zeta } g_{14} \psi \big)-\sqrt{\zeta } g_5 \phi  \big(\gamma +\sigma _{W_2}^2 \big(\eta '-\zeta \big)\big) \big(g_6 \sqrt{n_0} \big(\zeta  \psi -\eta  \psi -\zeta  \sigma _{W_2}^2\big)-\sqrt{\zeta } g_{14} \psi \big)\nonumber\\
&\quad\quad+\zeta ^{3/2} g_6 g_{12} n_0^{3/2} \phi  \sigma _{W_2}^2+\zeta ^{3/2} g_7 g_{16} n_0^{3/2} \phi  \sigma _{W_2}^2+\zeta ^{3/2} g_6 g_{20} n_0^{3/2} \phi  \sigma _{W_2}^2-n_0 \phi  \eta ' \sigma _{W_2}^2 \big(\sqrt{\zeta } \big(g_7 g_{16}+g_6 \big(g_{12}+g_{20}\big)\big) \sqrt{n_0}\nonumber\\
&\quad\quad-g_{17} \psi \big)+\zeta  g_8 g_{14} n_0 \psi  \sigma _{W_2}^2+\zeta  g_{13} g_{17} n_0 \psi  \sigma _{W_2}^2-\zeta  g_{17} n_0 \psi  \phi  \sigma _{W_2}^2-\zeta ^2 g_{13} g_{14} \psi  \sigma _{W_2}^4\\
0 &= -\gamma  \sqrt{\zeta } g_{12} g_{15} n_0^{3/2} \phi -\gamma  \sqrt{\zeta } g_{10} g_{16} n_0^{3/2} \phi -\gamma  \sqrt{\zeta } g_{15} g_{20} n_0^{3/2} +\gamma  g_{18} n_0 \psi  \phi -\sqrt{\zeta } \phi  \big(\gamma +\sigma _{W_2}^2 \big(\eta '-\zeta \big)\big) \nonumber\\
&\quad\quad\big(g_{15} \sqrt{n_0} \big(\zeta  \psi -\eta  \psi -\zeta  \sigma _{W_2}^2\big)-\sqrt{\zeta } g_{19} \psi \big)-\sqrt{\zeta } g_5 \phi  \big(\gamma +\sigma _{W_2}^2 \big(\eta '-\zeta \big)\big) \big(g_{15} \sqrt{n_0} \big(\zeta  \psi -\eta  \psi -\zeta  \sigma _{W_2}^2\big)-\zeta  g_{18} n_0 \psi  \phi  \sigma _{W_2}^2\nonumber\\
&\quad\quad-\sqrt{\zeta } g_{19} \psi \big)+\zeta ^{3/2} g_{12} g_{15} n_0^{3/2} \phi  \sigma _{W_2}^2+\zeta ^{3/2} g_{10} g_{16} n_0^{3/2} \phi  \sigma _{W_2}^2+\zeta ^{3/2} g_{15} g_{20} n_0^{3/2} \phi  \sigma _{W_2}^2-\zeta ^2 g_{13} g_{19} \psi  \sigma _{W_2}^4\nonumber\\
&\quad\quad-n_0 \phi  \eta ' \sigma _{W_2}^2 \big(\sqrt{\zeta } \big(g_{10} g_{16}+g_{15} \big(g_{12}+g_{20}\big)\big) \sqrt{n_0}-g_{18} \psi \big)+\zeta  g_{13} g_{18} n_0 \psi  \sigma _{W_2}^2+\zeta  g_8 g_{19} n_0 \psi  \sigma _{W_2}^2
\end{align}
\end{subequations}
}
After some straightforward algebra, one can eliminate all $g_i$ except for $g_6$ and $g_{13}$, which satisfy coupled polynomial equations. Those equations can be shown to be identical to eqn.~(\ref{eqn:tau}) by invoking the change of variables,
\begin{equation}
    g_6 = - \frac{\sqrt{\zeta} \psi}{\sqrt{n_0} \phi} \tau_2\,,\quad\text{and}\quad
    g_{13} = \big(\gamma +\sigma _{W_2}^2 \big(\eta '-\zeta \big)\big)\tau_1\,.
\end{equation}
In terms of $\tau_1$, $\tau_2$, and their derivatives $\tau_1'$~(\ref{eq:dtau1}), $\tau_2'$~(\ref{eq:dtau2}), the error $E_{33}$ is given by,
\begin{equation}
    E_{33} = \sigma_{W_2}^2 \left[\left(\tau_1 + (\sigma_{W_2}^2(\eta'-\zeta) + \gamma)\tau_1'+\sigma_{W_2}^2\zeta \tau_2'\right)/\tau_1^2 -\eta\right] - E_{22}\,.
\end{equation}
\subsection{Total test error}
Recall from eqns.~(\ref{eq:Etest}, \ref{eqn:E1}-\ref{eqn:E3}) that the total test error can be written as
\eq{
\Etest = 1 + E_{21} + E_{31} + E_{32} + \nu  \big(\eta \sigma_{W_2}^2 + E_{22} + E_{33}\big)\,,
}
where $\nu = 0$ with centering and $\nu = 1$ without it. Combining the results from the previous subsections, we find
\begin{align}
    \Etest &= 1 + 2(\tau_2/\tau_1-1) + \sigma_{\varepsilon}^2\big(\tau_1'/\tau_1^2-1\big) + 1-2\tau_2/\tau_1+\tau_2'/\tau_1^2 \\
    &\quad + \nu \sigma_{W_2}^2 \left[\left(\tau_1 + (\sigma_{W_2}^2(\eta'-\zeta) + \gamma)\tau_1'+\sigma_{W_2}^2\zeta \tau_2'\right)/\tau_1^2\right]\\
    &= \tau_2'/\tau_1^2 + \sigma_{\varepsilon}^2 \tau_1'/\tau_1^2 + \nu \sigma_{W_2}^2 \left[\left(\tau_1 + (\sigma_{W_2}^2(\eta'-\zeta) + \gamma)\tau_1'+\sigma_{W_2}^2\zeta \tau_2'\right)/\tau_1^2\right] - \sigma_{\varepsilon}^2\\
    & = (\gamma \tau_1)^{-2} \Etrain - \sigma_{\varepsilon}^2\,,
\end{align}
thereby establishing the result of the main theorem~(\ref{eq:Etest_Etrain}).

\end{document}